\documentclass[10pt,twocolumn,letterpaper]{article}
\pdfoutput=1

\usepackage{cvpr}
\usepackage{times}
\usepackage{epsfig}
\usepackage{graphicx}
\usepackage{amsmath}
\usepackage{amssymb}
\usepackage{amsthm}
\usepackage{subfig}
\usepackage{tabularx}
\usepackage{booktabs}
\usepackage{multirow}
\usepackage{multicol}
\usepackage{enumitem}
\usepackage{array}
\usepackage{color, colortbl}
\usepackage{fixltx2e}
\usepackage{makecell}
\usepackage{threeparttable}
\usepackage{microtype}

\usepackage[pagebackref=true,breaklinks=true,letterpaper=true,colorlinks,bookmarks=false]{hyperref}

\newcommand{\bx}{\mathbf{x}}

\newcommand{\bz}{\mathbf{z}}

\newcommand{\bs}{\mathbf{s}}

\newcommand{\bbf}{\mathbf{f}}

\newcommand{\bw}{\mathbf{w}}

\newcommand{\cZ}{\mathcal{Z}}
\newcommand{\norm}[1]{\left\lVert#1\right\rVert}

\cvprfinalcopy % *** Uncomment this line for the final submission

\def\cvprPaperID{5260} % *** Enter the CVPR Paper ID here

\begin{document}

%%%%%%%%% TITLE
\title{Towards Universal Representation Learning for Deep Face Recognition }

\author{Yichun Shi$^{1}$\qquad Xiang Yu$^{2}$\qquad Kihyuk Sohn$^{2}$\qquad Manmohan Chandraker$^{2}$\qquad Anil K. Jain$^{1}$ \\
 $^{1}$Michigan State University \qquad
 $^{2}$NEC Labs America
}

%\and

\maketitle
%\thispagestyle{empty}

%%%%%%%%% ABSTRACT
\begin{abstract}
Recognizing wild faces is extremely hard as they appear with all kinds of variations.  Traditional methods either train with specifically annotated variation data from target domains, or by introducing unlabeled target variation data to adapt from the training data. Instead, we propose a universal representation learning framework that can deal with larger variation unseen in the given training data without leveraging target domain knowledge. We firstly synthesize training data alongside some semantically meaningful variations, such as low resolution, occlusion and head pose. However, directly feeding the augmented data for training will not converge well as the newly introduced samples are mostly hard examples. We propose to split the feature embedding into multiple sub-embeddings, and associate different confidence values for each sub-embedding to smooth the training procedure. The sub-embeddings are further decorrelated by regularizing variation classification loss and variation adversarial loss on different partitions of them. Experiments show that our method achieves top performance on general face recognition datasets such as LFW and MegaFace, while significantly better on extreme benchmarks such as TinyFace and IJB-S.
\end{abstract}
%%%%%%%%% BODY TEXT
\section{Introduction}
\begin{figure} [t]
    \captionsetup{font=small}
    \centering
    \includegraphics[width=\linewidth]{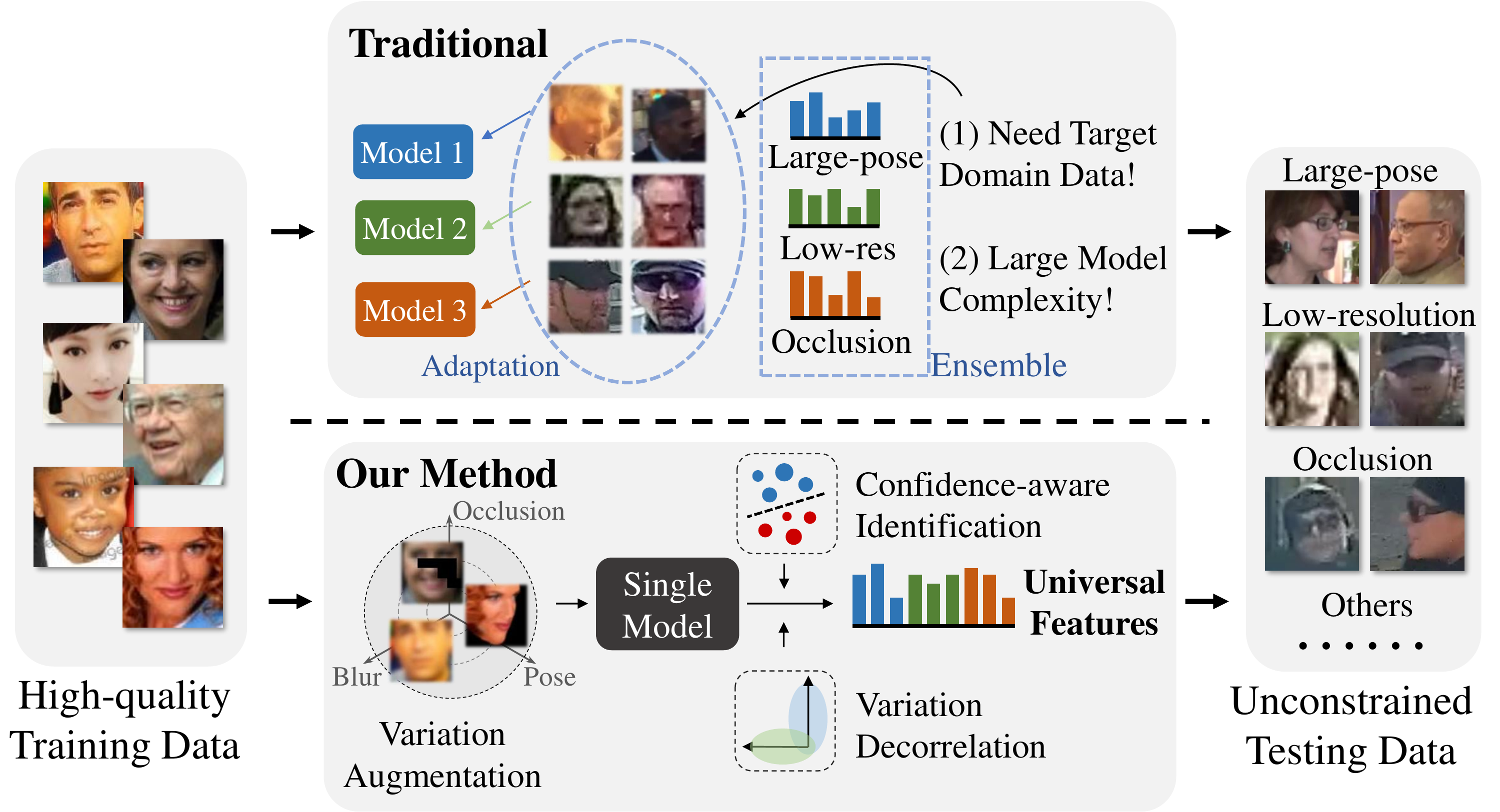}
    \caption{Traditional recognition models require target domain data to adapt from the high-quality training data to conduct unconstrained/low-quality face recognition. Model ensemble is further needed for a universal representation purpose which significantly increases model complexity. In contrast, our method works only on original training data without any target domain data information, and can deal with unconstrained testing scenarios.}
    \label{fig:frontal}
    \vspace{-4mm}
\end{figure}

Deep face recognition seeks to map input images to a feature space with small intra-identity distance and large inter-identity distance, which has been achieved by prior works through loss design and datasets with rich within-class variations \cite{schroff2015facenet,wen2016discriminative,liu2017sphereface,wang2018cosface,deng2018arcface}. However, even very large public datasets such as MS-Celeb-1M manifest strong biases, such as ethnicity \cite{sohn2019iclr} or head poses \cite{masi2016we,pengiccv2017}. This lack of variation leads to significant performance drops on challenging test datasets, for example, accuracies reported by prior state-of-the-art \cite{shi2019probabilistic} on IJB-S or TinyFace \cite{IJBS,TinyFace} are about $30\%$ lower than IJB-A \cite{IJBA} or LFW \cite{LFWTech}.

Recent works seek to mitigate this issue by identifying relevant factors of variation and augmenting datasets to incorporate them through domain adaptation method \cite{sohn2019iclr}. Sometimes, such variations are hard to identify, so domain adaptation methods are used to align features between training and test domains \cite{sankaranarayanan2016triplet}. Alternatively, individual models might be trained on various datasets and ensembled to obtain good performance on each \cite{masi2016pose}. All these approaches either only handle specific variations, or require access to test data distributions, or accrue additional runtime complexity to handle wider variations. In contrast, we propose learning a single ``universal'' deep feature representation that handles the variations in face recognition without requiring access to test data distribution and retains runtime efficiency, while achieving strong performance across diverse situations especially on the low-quality images (see Figure \ref{fig:frontal}).

This paper introduces several novel contributions in Section \ref{sec:methods} to learn such a universal representation. First, we note that inputs with non-frontal poses, low resolutions and heavy occlusions are key nameable factors that present challenges for ``in-the-wild'' applications, for which training data may be synthetically augmented. But directly adding hard augmented examples into training leads to a harder optimization problem. We mitigate this by proposing an identification loss that accounts for per-sample confidence to learn a probabilistic feature embedding. Second, we seek to maximize representation power of the embedding by decomposing it into sub-embeddings, each of which has an independent confidence value during training. 
Third, all the sub-embeddings are encouraged to be further decorrelated through two opposite regularizations over different partitions of the sub-embeddings, i.e., variation classification loss and variation adversarial loss. 
Fourth, we extend further decorrelation regularization by mining additional variations within the training data for which synthetic augmentation is not trivial. 
Finally, we account for the varying discrimination power of sub-embeddings for various factors through a probabilistic aggregation that accounts for their uncertainties.

In Section \ref{sec:experiments}, we extensively evaluate the proposed methods on public datasets. Compared to our baseline model, the proposed method maintains the high accuracy on general face recognition benchmarks, such as LFW and YTF, while significantly boosting the performance on challenging datasets such as IJB-C, IJB-S, where new state-of-the-art performance is achieved. Detailed ablation studies show the impact of each of the above contributions in achieving these strong performance.

In summary, the main contributions of this paper are:
\begin{itemize}[leftmargin=10pt]\vspace{-0.5em}
\item{A face representation learning framework that learns universal features by associating them with different variations, leading to improved generalization on diverse testing datasets.}\vspace{-0.5em} 
\item{A confidence-aware identification loss that utilizes sample confidence during training to learn features from hard samples.}\vspace{-0.5em}
\item{A feature decorrelation regularization that applies both variation classification loss and variation adversarial loss on different partitions of the sub-embeddings, leading to improved performance.}\vspace{-0.5em}
\item{A training strategy to effectively combine synthesized data to train a face representation applicable to images outside the original training distribution.}\vspace{-0.5em}
\item{State-of-the-art results on several challenging benchmarks, such as IJB-A, IJB-C, TinyFace and IJB-S.}
%  IJB-A~\cite{IJBA}, IJB-C~\cite{IJBC}, TinyFace~\cite{TinyFace} and IJB-S~\cite{IJBS}.
\end{itemize}

\section{Related Work}
\noindent\textbf{Deep Face Recognition}:
Deep neural networks is widely adopted in the ongoing research in face recognition ~\cite{taigman2014deepface,deepid2,schroff2015facenet,masi2016we,liu2017sphereface,hasnat2017deepvisage,ranjan2017l2,wang2018additive,deng2018arcface}. Taigman~\etal~\cite{taigman2014deepface} proposed the first deep convolutional neural network for face recognition. The following works have been exploring different loss functions to improve the discrimination power of the feature representation. Wen~\etal ~\cite{wen2016discriminative} proposed center loss to reduce the intra-class variation. A series of work have proposed to use metric learning methods for face recognition ~\cite{schroff2015facenet,sohn2016improved}. Recent work has been trying to achieve discriminative embeddings with a single identification loss function where proxy/prototype vectors are used to represent each class in the embedding space~\cite{liu2017sphereface,wang2018additive,wang2018cosface,ranjan2017l2,deng2018arcface}.
\begin{figure}[t]
    \centering
    \captionsetup{font=small}
    \captionsetup[subfloat]{captionskip=1pt}
    \subfloat[Blur]{\label{fig:augmentation_blur}\includegraphics[width=0.49\linewidth]{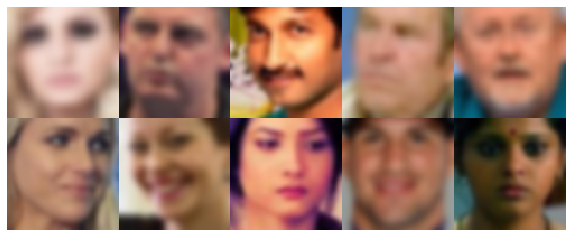}}\hfill
    \subfloat[Occlusion]{\label{fig:augmentation_occlusion}\includegraphics[width=0.49\linewidth]{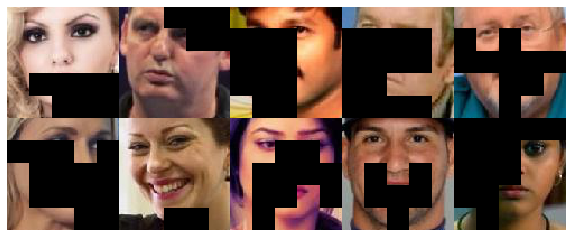}}\\[-1.0em]
    \subfloat[Pose]{\label{fig:augmentation_pose}\includegraphics[width=0.49\linewidth]{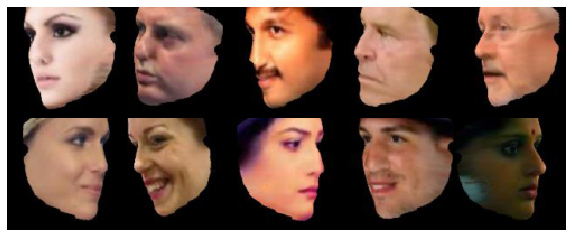}}\hfill
    \subfloat[Randomly Combined]{\label{fig:augmentation_combined}\includegraphics[width=0.49\linewidth]{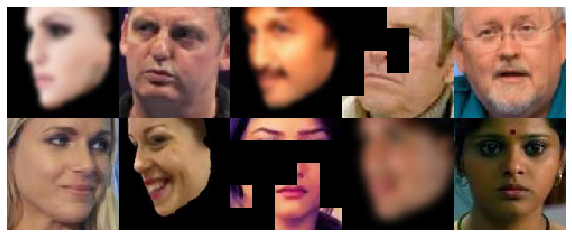}}
    \vspace{-0.5em}\caption{Samples with augmentation alongside different variations.}\vspace{-1.5em}
    \label{fig:augmentation_sample}
    \vspace{-1mm}
\end{figure}

\noindent\textbf{Universal Representation}:
Universal representation refers to a single model that can be applied to variant visual domains (usually different tasks), e.g. object, character, road signs, while maintaining the performance of using a set of domain-specific models~\cite{bilen2017universal,rebuffi2017learning,rebuffi2018efficient,wang2019towards}. The features learned by such a single model are believed to be more universal than domain-specific models. Different from domain generalization~\cite{khosla2012undoing,muandet2013domain,li2017deeper,li2018learning,tamaazousti2019learning}, which targets adaptability on unseen domains by learning from various seen domains, the universal representation does not involve re-training on unseen domains. Most of these methods focus on increasing the parameter efficiency by reducing the domain-shift with techniques such as conditioned BatchNorm~\cite{bilen2017universal} and residual adapters~\cite{rebuffi2017learning,rebuffi2018efficient}. Based on SE modules~\cite{hu2018squeeze}, Wang~\etal~\cite{wang2019towards} proposed a domain-attentive module for intermediate (hidden) features of a universal object detection network. Our work is different from these methods in two ways: (1) it is a method for similarity metric learning rather than detection or classification tasks and (2) it is model-agnostic. The features learned by our model can then be directly applied to different domains by computing the pairwise similarity between samples of unseen classes.

\begin{figure*} [t]
    \captionsetup{font=small}
    \centering
    \includegraphics[width=0.92\linewidth]{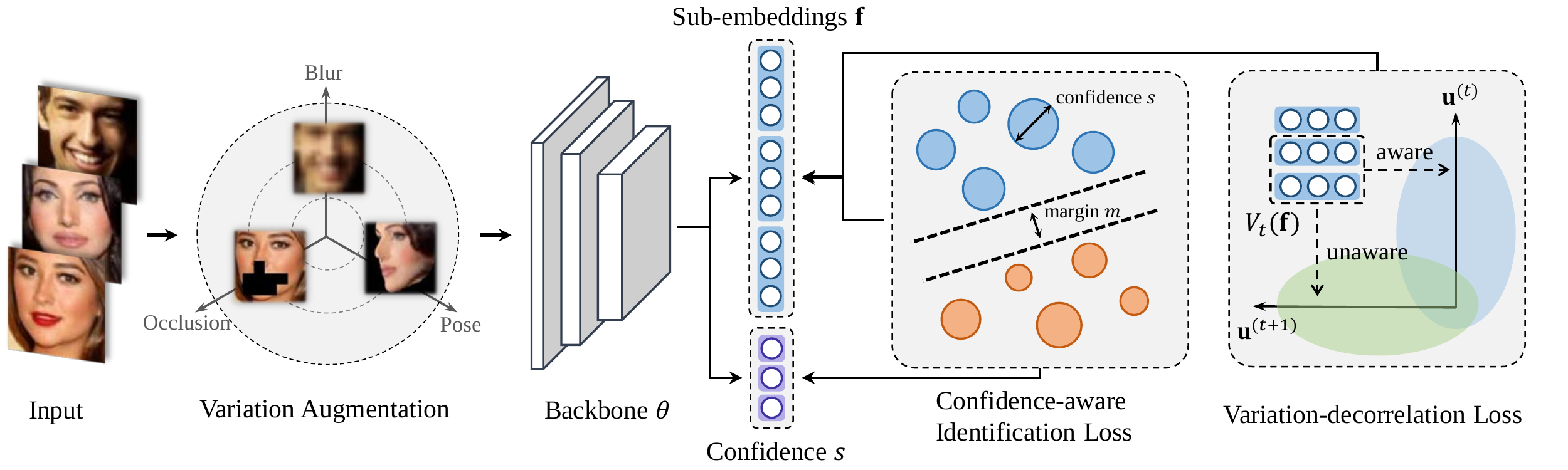}
    \caption{Overview of the proposed method. High-quality input images are firstly augmented alongside our defined augmentable variations, i.e., blur, occlusion and pose. The feature representation is then split into sub-embeddings associated with sample-specific confidence s. Confidence-aware identification loss and variation decorrelation loss are developed to learn the sub-embeddings.}
    \label{fig:overview}
    \vspace{-5mm}
\end{figure*}

\section{Proposed Approach}
\label{sec:methods}
In this section, we first introduce three augmentable variations, namely blur, occlusion and head pose, to augment the training data. Visual examples of augmented data are shown in Figure~\ref{fig:augmentation_sample} and the details can be found in Section~\ref{sec:details}. Then in Section~\ref{sec:confidence_loss}, we introduce a confidence-aware identification loss to learn from hard examples, which is further extended in Section~\ref{sec:subembedding} by splitting the feature vectors into sub-embeddings with independent confidence. In Section~\ref{sec:method_decorrelation}, we apply the introduced augmentable variations to further decorrelate the feature embeddings. A non-augmentable variation discovery is proposed to explore more variations for better decorrelation. Finally, an uncertainty guided pair-wise metric is proposed for inference.

\subsection{Confidence-aware Identification Loss}
\label{sec:confidence_loss}
\begin{figure} [t]
    \vspace{-4mm}
    \captionsetup{font=small}
    \centering
    \subfloat[w/o confidence]{\includegraphics[width=0.47\linewidth]{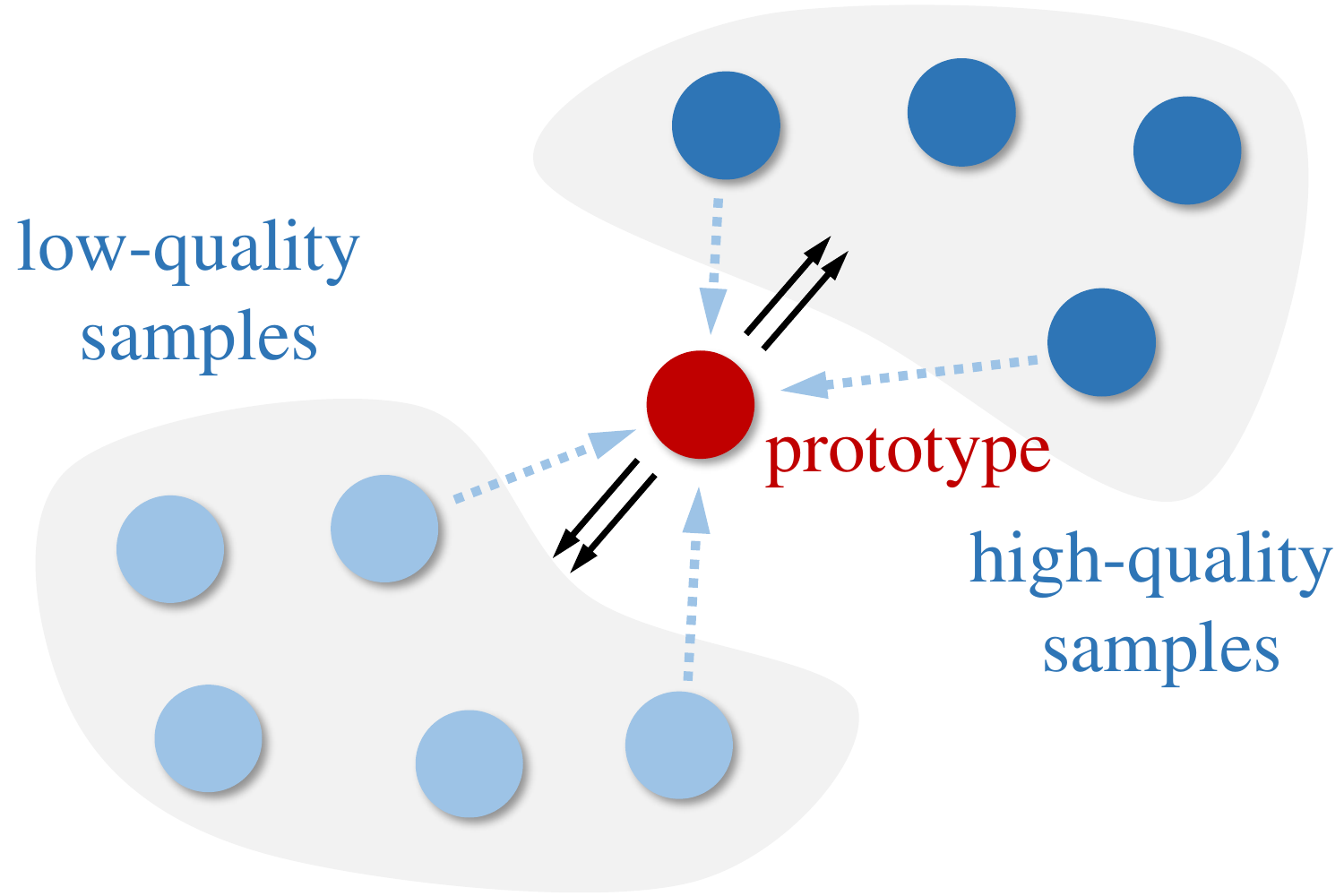}}\hfill
    \subfloat[w/ confidence]{\includegraphics[width=0.47\linewidth]{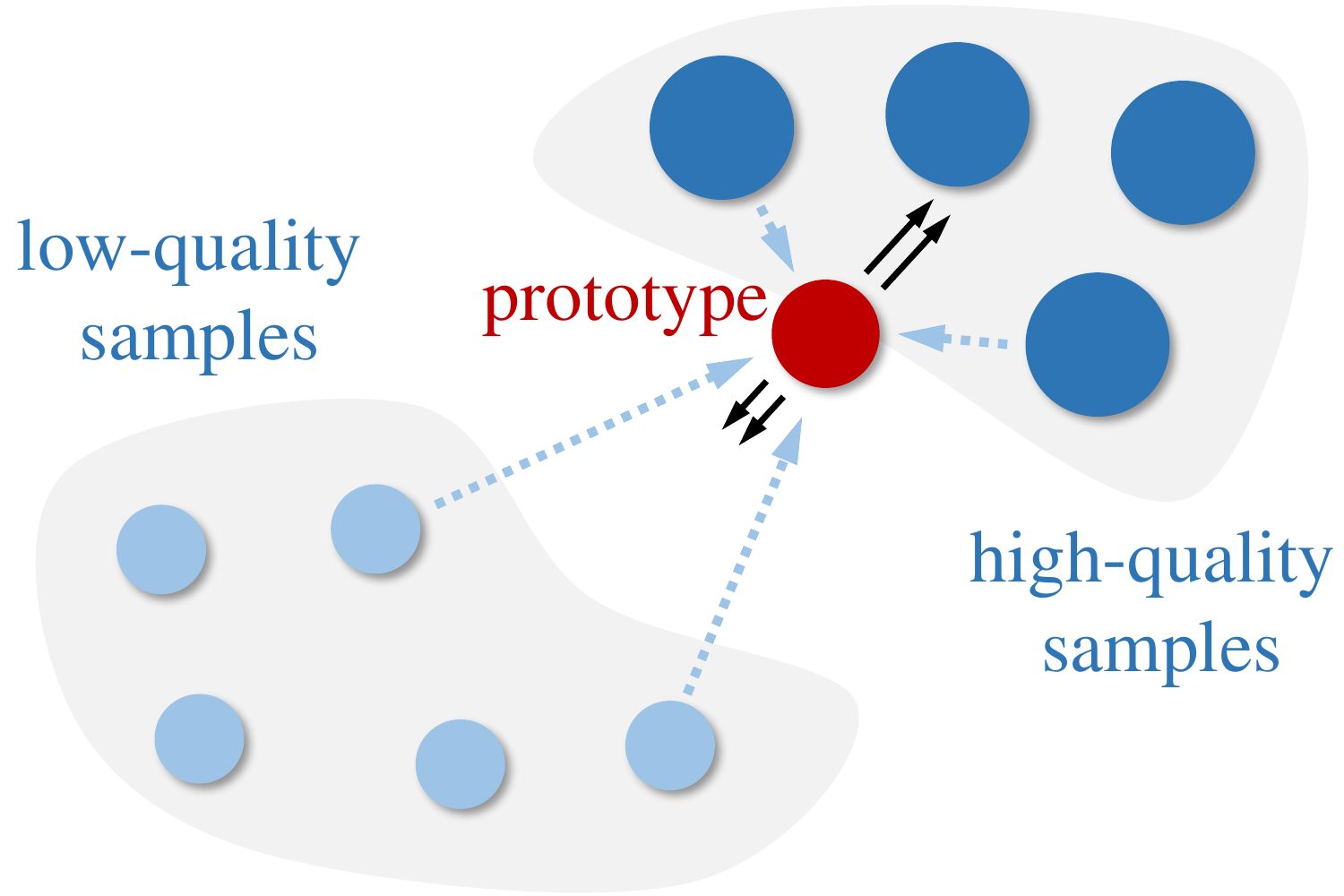}}
    \caption{Illustration of confidence-aware embedding learning on quality-various data. With confidence guiding, the learned prototype is closer to high-quality samples which represents the identity better.}
    \label{fig:csoftmax}
    \vspace{-5mm}
\end{figure}

We investigate the posterior of the probability being classified to identity $j\in\{1,2,\dots,N\}$, given the input sample $\bx_i$. Denote the feature embedding of sample $i$ as $\mathbf{f}_{i}$ and $j^{\text{th}}$ identity prototype vector as $\bw_j$, which is the identity template feature.
A probabilistic embedding network $\theta$ represents each sample $\bx_i$ as a Gaussian distribution $\mathcal{N}(\mathbf{f}_i, \sigma_i^2\mathbf{I})$ in the feature space.  
%Denote $y$ the identity label and $N$ the number of identities. For each identity $j\in\{1,2,\dots,N\}$, we maintain a prototype vector $\bw_j$, indicating the intrinsic feature template of the $j^{\text{th}}$ identity. For each input $\bx_i$, a probabilistic embedding $\theta$ represents in the latent space as a Gaussian distribution $\mathcal{N}(\mathbf{f}_i, \sigma_i^2\mathbf{I})$.
%In other words, $p(\bz|y=j)=\delta(\bz-\bw_j)$, where $\delta$ is the Dirac delta function. Assuming an non-informative prior $p(\bz)$, 
The likelihood of $\bx_i$ being a sample of $j^{\text{th}}$ class is given by:
\begin{align}
    p(\bx_i|y=j) \propto\; & p_\theta(\bw_j|\bx_i) \\
                =\; & \frac{1}{(2\pi\sigma^{2}_i)^{\frac{D}{2}}}\exp(-\frac{\norm{\mathbf{f}_i-\bw_j}^2}{2\sigma^{2}_i}),
\end{align}
where $D$ is feature dimension. Further assuming the prior of assigning a sample to any identity as equal, the posterior of $\bx_i$ belonging to the $j^{\text{th}}$ class is derived as:
\begin{align}
p(y=j|\bx_i) = & \frac{p(\bx_i|y=j)p(y=j)}{\sum_{c=1}^{N}{p(\bx_i|y=c)p(y=c)}} \\
= & \frac{\exp(-\frac{\norm{\mathbf{f}_i-\bw_j}^2}{2\sigma^{2}_i})}{\sum_{c=1}^{N}{\exp(-\frac{\norm{\mathbf{f}_i-\bw_c}^2}{2\sigma^{2}_i})}},
\label{eq:uncertainty}
\end{align}
For simplicity, let us define a confidence value $s_i=\frac{1}{\sigma^{2}_i}$. Constraining both $\mathbf{f}_i$ and $\bw_j$ on the $\ell_2$-normalized unit sphere, we have $\frac{\norm{\mathbf{f}_i-\bw_j}^2}{2\sigma^{2}_i} = s_i(1-\bw_j^T\mathbf{f}_i)$ and
\begin{equation}
p_(y=j|\bx_i) = \frac{\exp(s_i\bw_j^T\mathbf{f}_i)}{\sum_{c=1}^{N}{\exp(s_i\bw_c^T\mathbf{f}_i)}}.
\label{eq:posterior}
\end{equation}
The effect of confidence-aware posterior in Equation~\ref{eq:posterior} is illustrated in Figure~\ref{fig:csoftmax}. When training is conducted among samples of various qualities, if we assume the same confidence across all samples, the learned prototype will be in the center of all samples. This is not ideal, as low-quality samples convey more ambiguous identity information. In contrast, if we set up sample-specific confidence $s_i$, where high-quality samples show higher confidence, it pushes the prototype $\bw_j$ to be more similar to high-quality samples in order to maximize the posterior. Meanwhile, during update of the embedding $\mathbf{f}_i$, it provides a stronger push for low-quality $\mathbf{f}_i$ to be closer to the prototype.

Adding loss margin~\cite{wang2018cosface} over the exponential logit has been shown to be effective in narrowing the within-identity distribution. We also incorporate it into our loss:
\begin{align}
\label{eq:loss_scalar}
%\mathcal{L}_{idt}' =&-\log{p(y_i=\hat{y}_i|\bx_i)} \\
\mathcal{L}_{idt}' = -\log \frac{\exp(s_i\bw_{y_i}^T\mathbf{f}_i-m)}{\exp(s_i\bw_{y_i}^T\mathbf{f}_i-m)+\sum_{j\neq y_i}{\exp(s_i\bw_j^T\mathbf{f}_i)}},
\vspace{-4mm}
\end{align}
where $y_i$ is the ground-truth label of $\bx_i$. 
%Notice that Equation~\ref{eq:loss_scalar} is similar to the CosFace loss function~\cite{wang2018cosface} except 
Our confidence-aware identification loss (C-Softmax) is mainly different from Cosine Loss\cite{wang2018cosface} as follows:
(1) each image has an independent and dynamic $s_i$ rather than a constant shared scalar and (2) the margin parameter $m$ is not multiplied by $s_i$. 
The independence of $s_i$ allows it to gate the gradient signals of $\bw_j$ and $\mathbf{f}_i$ during network training in a sample-specific way, as the confidence (degree of variation) of training samples can have large differences. Though samples are specific, we aim to pursue an homogeneous feature space such that the metric across different identities should be consistent. Thus, allowing $s_i$ to compensate for the confidence difference of the samples, we expect $m$ to be consistently shared across all the identities.

%\begin{figure} [t]
%     \centering
%     \includegraphics[width=0.9\linewidth]{fig/embedding.pdf}
%     \caption{Illustration of the confidence-aware multi-embedding network.}
%     \label{fig:network}
%\end{figure}

\subsection{Confidence-aware Sub-Embeddings}
\label{sec:subembedding}
Though the embedding $\mathbf{f}_i$ learned through a sample-specific gating $s_i$ can deal with sample-level variations, we argue that the correlation among the entries of $\mathbf{f}_i$ itself is still high. To maximize the representation power and achieve a compact feature size, decorrelating the entries of the embedding is necessary. This encourages us to further break the entire embedding $\mathbf{f}_i$ into partitioned sub-embeddings, each of which is further assigned a scalar confidence value.

\begin{figure} [t]
    \centering
    \captionsetup{font=small}
    \subfloat[sub-embedding of size 8]{\includegraphics[width=0.48\linewidth,trim={0mm 10mm 0mm 10mm},clip]{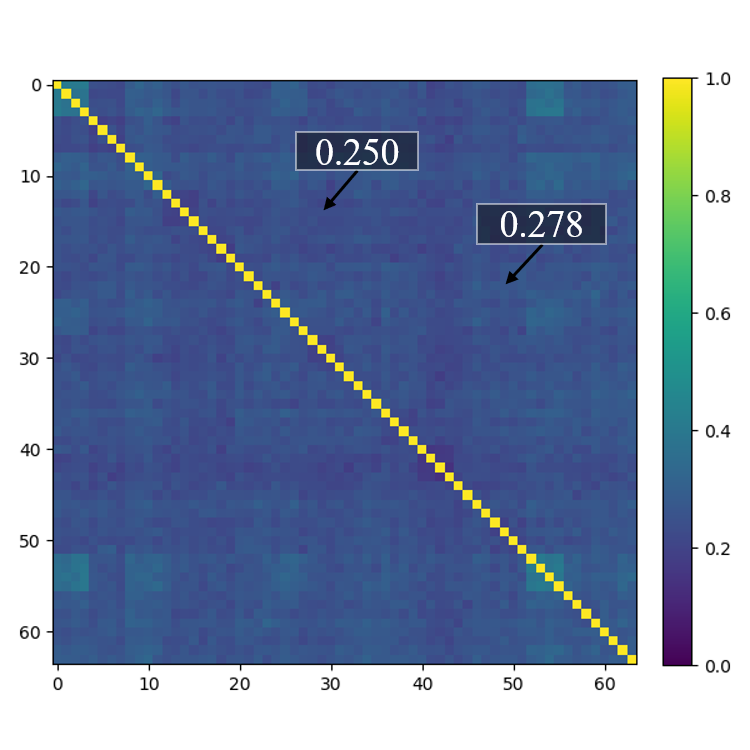}\vspace{-1.0em}}\hfill
    \subfloat[sub-embedding of size 32]{\includegraphics[width=0.48\linewidth,trim={0mm 10mm 0mm 10mm},clip]{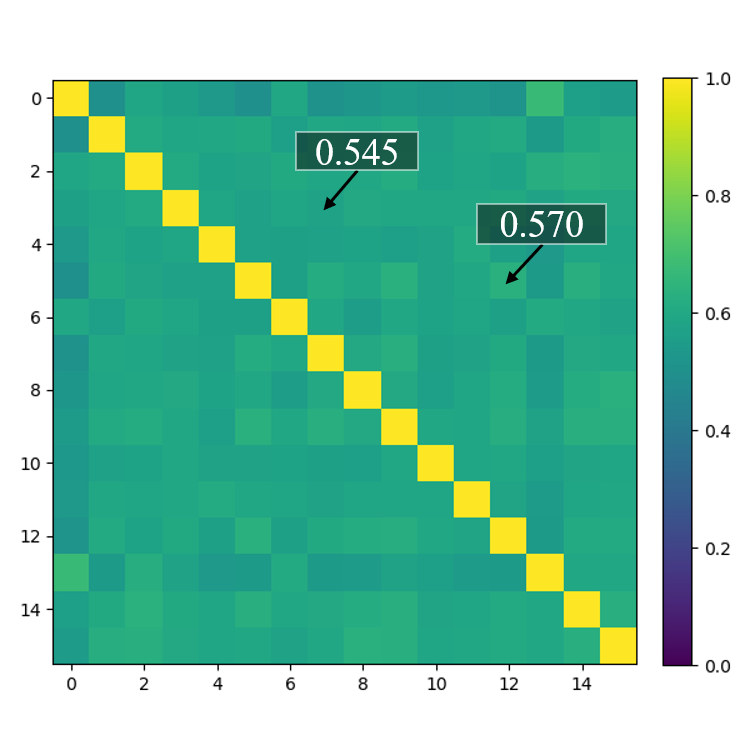}\vspace{-1.0em}}
    \caption{The correlation matrices of sub-embeddings by splitting the feature vector into different sizes. The correlation is computed in terms of distance to class center.}
    \label{fig:correlation}
    \vspace{-4mm}
\end{figure}
Illustrated in Figure~\ref{fig:overview}, we partition the entire feature embedding $\mathbf{f}_i$ into $K$ equal-length sub-embeddings as in Equation~\ref{eq:partition}. Accordingly, the prototype vector $\bw_j$ and the confidence scalar $s_i$ are also partitioned into the same size $K$ groups.
\begin{align}
\begin{split}
\bw_j &= [\bw_j^{(1)T},\bw_j^{(2)T},\dots,\bw_j^{(K)T}],\\
\mathbf{f}_i &= [\mathbf{f}_i^{(1)T},\mathbf{f}_i^{(2)T},\dots,\mathbf{f}_i^{(K)T}],\\
\bs_i &= [s_i^{(1)},s_i^{(2)},\dots,s_i^{(K)}],\\
\end{split}
\label{eq:partition}
\end{align}
Each group of sub-embeddings $\mathbf{f}_i^{(k)}$ is $\ell_2$ normalized onto unit sphere separately. The final identification loss thus is: 
\begin{equation}
\label{eq:loss_multi}
\mathcal{L}_{idt} = -\log \frac{\exp(\mathbf{a}_{i,y_i}-m)}{\exp(\mathbf{a}_{i,y_i}-m)+\sum_{j\neq y_i}{\exp(\mathbf{a}_{i,j})}},
\end{equation}
\begin{equation}
    \mathbf{a}_{i,j} = \frac{1}{K}\sum_{k=1}^{K}{ s_i^{(k)}\bw_j^{(k)T}\mathbf{f}_i^{(k)}}.
\end{equation}
A common issue for neural networks is that they tend to be ``over-confident'' on predictions~\cite{confidence_icml2017}. We add an additional $l_2$ regularization to constrain the confidence from growing arbitrarily large:
\begin{equation}
\mathcal{L}_{reg} = \frac{1}{K}\sum_{k=1}^{K} s_i^{(k)2}.\\
\label{eq:regularization}
\end{equation}

\subsection{Sub-Embeddings Decorrelation}
\label{sec:method_decorrelation}

% \begin{figure} [t]
%     \centering
%     \captionsetup{font=small}
%     \includegraphics[width=0.7\linewidth]{fig/splitting.pdf}
%     \caption{The variation decorrelation loss disentangles different sub-embeddings by associating them with different variations in the training dataset. In this example, the first two sub-embeddings are forced to be invariant to occlusion while the second two sub-embeddings are forced to be invariant to blur. By pushing stronger invariance for each variation, the correlation/overlap between two variations is reduced.}
%     \label{fig:decorrelation}
% \end{figure}
\begin{figure} [t]
    \captionsetup{font=small}
    \centering
    \subfloat[variation-correlated features]{\includegraphics[width=0.47\linewidth]{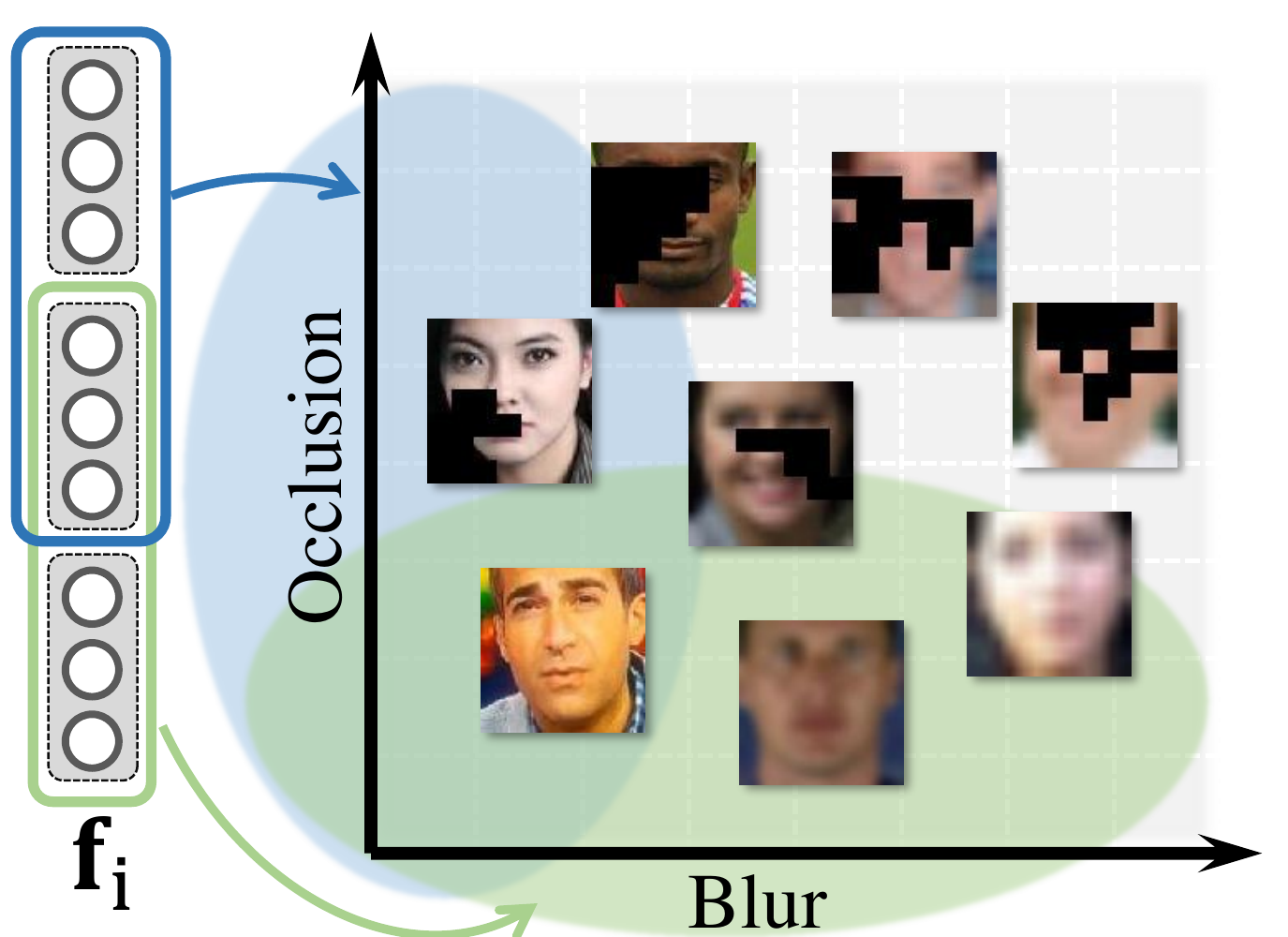}}\hfill
    \subfloat[variation-decorrelated features]{\includegraphics[width=0.47\linewidth]{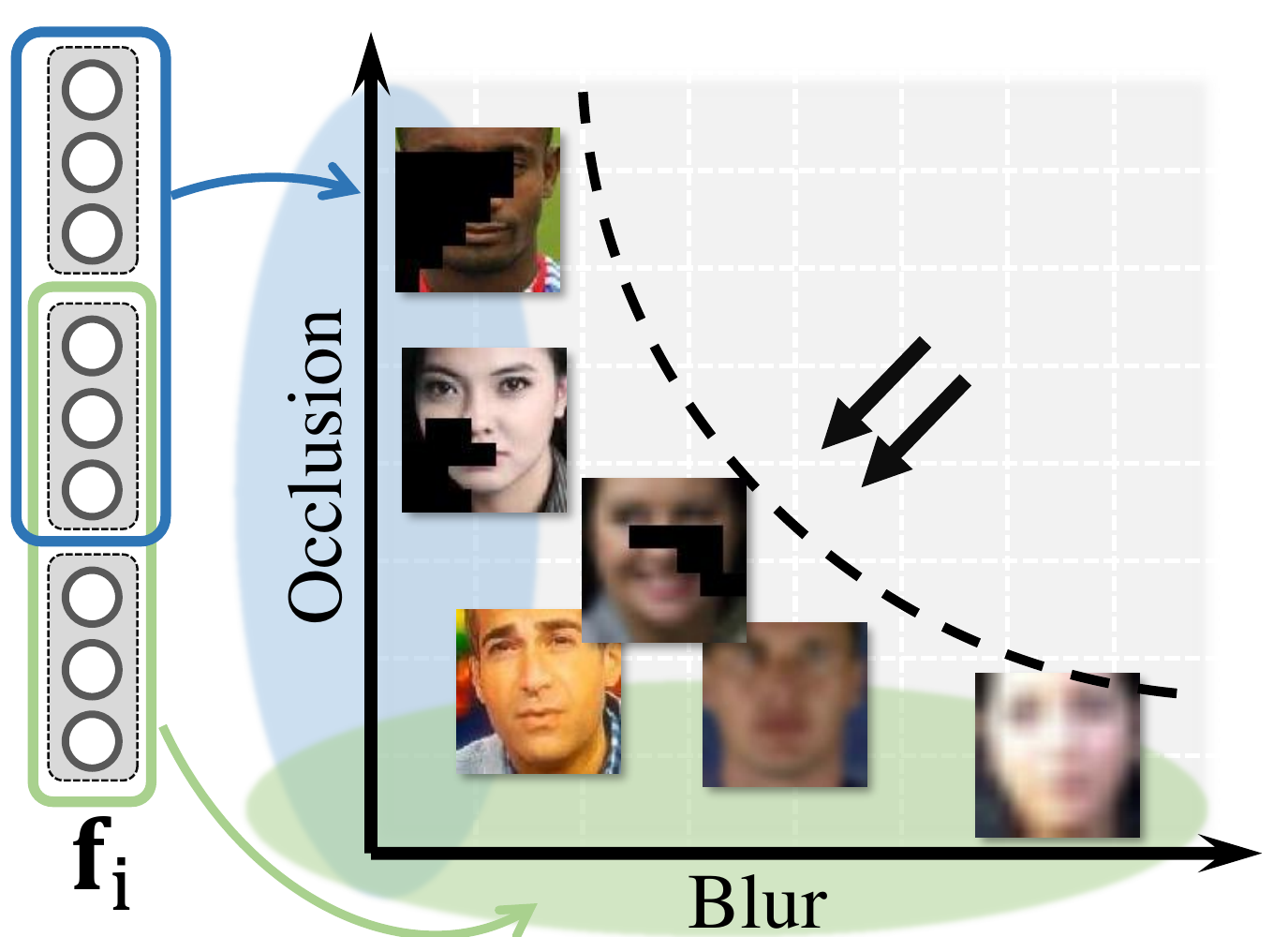}}
    \caption{The variation decorrelation loss disentangles different sub-embeddings by associating them with different variations. In this example, the first two sub-embeddings are forced to be invariant to occlusion while the second two sub-embeddings are forced to be invariant to blur. By pushing stronger invariance for each variation, the correlation/overlap between two variations is reduced.}
    \label{fig:confidence}
    \vspace{-5mm}
\end{figure}

Setting up multiple sub-embeddings alone does not guarantee the features in different groups are learning complementary information. Empirically shown in Figure~\ref{fig:correlation}, we find the sub-embeddings are still highly correlated, i.e., dividing $\mathbf{f}_i$ into equal $16$ groups, the average correlation among all the sub-embeddings is $0.57$. 
If we penalize the sub-embeddings with different regularization, the correlation among them can be reduced. By associating different sub-embeddings with different variations, we conduct variation classification loss on a subset of all the sub-embeddings while conducting variation adversarial loss in terms of other variation types. Given multiple variations, such two regularization terms are forced on different subsets, leading to better sub-embedding decorrelation. 

For each augmentable variation $t\in \{1,2,\dots,M\}$, we generate a binary mask $V_t$, which selects a random $\frac{K}{2}$ subset of all sub-embeddings while setting the other half to be zeros. The masks are generated at the beginning of the training and will remain fixed during training. We guarantee that for different variations, the masks are different. We expect $V_t(\mathbf{f}_i)$ to reflect $t^{\text{th}}$ variation while invariant to the others. Accordingly, we build a multi-label binary discriminator $C$ by learning to predict all variations from each masked subset:
\vspace{-0.5em}\begin{align}
\begin{split}
\min_{C} \mathcal{L}_{C} = & -\sum_{t=1}^{M}\log p_{C}(\mathbf{u}_i=\hat{\mathbf{u}}_i|V_t(\mathbf{f}_i)) \\
    = & -\sum_{t=1}^{M}\sum_{t'=1}^{M} \log p_{C}(u^{(t')}_i=\hat{u}^{(t')}_i|V_t(\mathbf{f}_i)) \\
\end{split}\vspace{-1.0em}
\label{eq:domain_cls}
\end{align}
where $\mathbf{u}_i=[u_i^{(1)}, u_i^{(2)}, \dots, u_i^{(M)}]$ are the binary labels (0/1) of the known variations and $\mathbf{\hat{\mathbf{u}}}_i$ is the ground-truth label. For example, if $t=1$ corresponds to resolution, $\hat{u}_i^{(1)}$ would be $1$ and $0$ for high/low-resolution images, respectively. Note that Equation~\ref{eq:domain_cls} is only used for training the discriminator $C$. The corresponding classification and adversarial loss of the embedding network is then given by:
\vspace{-0.5em}
\begin{align}
\begin{split}
\mathcal{L}_{cls} = -\sum_{t=1}^{M}\log p_{C}({u}^{(t)}=\hat{{u}}_i^{(t)}|V_t(\mathbf{f}_i)) \\
\end{split}\vspace{-2.0em}
\label{eq:decorrelation_cls}
\end{align}
\begin{align}
\begin{split}
\mathcal{L}_{adv} = -\sum_{t=1}^{M}\sum_{t'\neq t} (\frac{1}{2}\log p_{C}({u}^{(t')}=0|V_t(\mathbf{f}_i)) +\\
    \frac{1}{2} \log p_{C}({u}^{(t')}=1|V_t(\mathbf{f}_i))) 
\end{split}\vspace{-1.0em}
\label{eq:decorrelation_adv}
\end{align}
The classification loss $\mathcal{L}_{cls}$ to encourage $V_t$ to be variation-specific while $\mathcal{L}_{adv}$ is an adversarial loss to encourage invariance to the other variations. As long as no two masks are the same, it guarantees that the selected subsets $V_t$ is functionally different from other $V_{t'}$. We thus achieve decorrelation between $V_t$ and $V_{t'}$. The overall loss function for each sample is:
\vspace{-0.5em}\begin{equation}
    \min_{\theta}\mathcal{L} = \mathcal{L}_{idt} + \lambda_{reg}\mathcal{L}_{reg} + \lambda_{cls}\mathcal{L}_{cls} + \lambda_{adv}\mathcal{L}_{adv}.
\label{eq:loss_all}\vspace{-0.5em}
\end{equation}
During the optimization, Equation~(\ref{eq:loss_all}) is averaged across the samples in the mini-batch.

\subsection{Mining More Variations}
The limited number (three in our method) of augmentable variations leads to limited effect of decorrelation as the number of $V_t$ are too small. To further enhance the decorrelation, as well to introduce more variations for better generalization ability, we aim to explore more variations with semantic meaning.
%the unnameable domains from the existing training data. 
Notice that not all the variations are easy to conduct data augmentation, e.g. smiling or not is hard to augment. For such variations, we attempt to mine out the variation labels from the original training data. In particular, we leverage an off-the-shelf attribute dataset CelebA~\cite{celeba} to train a attribute classification model $\theta_{A}$ with identity adversarial loss:
\vspace{-0.5em}\begin{align}
\begin{split}
& \min_{\theta_{A}} \mathcal{L}_{\theta_A} = -\log p(l_A|\bx_A)
    - \frac{1}{N_A}\sum_{c}^{N_A} \log p(y_A=c|\bx_A)\\
& \min_{D_{A}} \mathcal{L}_{D_A} = -\log p(y_A=y_{\bx_A}|\bx_A),\\
\end{split}\raisetag{1.0\baselineskip}\vspace{-1.0em}
\end{align}
where $l_A$ is the attribute label and $y_A$ is the identity label. $\bx_A$ is the input face image and $N_A$ is the number of identities in the CelebA dataset. The first term penalizes the feature to classify facial attributes and the second term penalizes the feature to be invariant to identities. 

% From our recognition training data, we manually select few samples (e.g., 10s) to define the newly introduced variations, i.e. smiling or not, young or old, etc. With the learned attribute feature representation, we learn a linear classifier for the recognition dataset to set up the $T$ additional variation binary labels merging with the original augmentable variation binary labels as: $\mathbf{u}_i=[u_i^{(1)},\dots,u_i^{(M)},u_i^{(M+1)},\dots,u_i^{(M+T)}]$. The decorrelation loss Equation~\ref{eq:decorrelation} is then updated with newly introduced $\mathbf{u}_i$.

The attribute classifier is then applied to the recognition training set to generate $T$ new soft variation labels, e.g. smiling or not, young or old. These additional variation binary labels are merged with the original augmentable variation labels as: $\mathbf{u}_i=[u_i^{(1)},\dots,u_i^{(M)},u_i^{(M+1)},\dots,u_i^{(M+T)}]$ and are then incorporated into the decorrelation learning framework in Section~
\ref{sec:method_decorrelation}.

\subsection{Uncertainty-Guided Probabilistic Aggregation}
\label{sec:method_aggregation}
Considering the metric for inference, simply taking the average of the learned sub-embeddings is sub-optimal. This is because different sub-embeddings show different discriminative power for different variations. Their importance should vary according to the given image pairs. Inspired by~\cite{shi2019probabilistic}, we consider to apply the uncertainty associated with each embedding for a pairwise similarity score:
% \begin{align}
% \begin{split}
%     s(\bx_i,\bx_j)=  &-\frac{1}{2}\sum_{k=1}^{K}\frac{\norm{\bmu^{(k)}_i-\bmu^{(k)}_j}^2}{\sigma_i^{2(k)}+\sigma_j^{2(k)}}\\
%     & -\frac{D}{2K}\sum_{k=1}^{K}\log(\sigma_i^{2(k)}+\sigma_j^{2(k)})\\
% \end{split}
% \label{eq:likelihood}
% \end{align}
\vspace{-0.5em}\begin{align}
\begin{split}
    score(\bx_i,\bx_j)= & -\frac{1}{2}\sum_{k=1}^{K}\frac{\norm{\mathbf{f}^{(k)}_i-\mathbf{f}^{(k)}_j}^2}{\sigma_i^{(k)2}+\sigma_j^{(k)2}}\\
                        & -\frac{D}{2K}\sum_{k=1}^{K}\log (\sigma_i^{(k)2}+\sigma_j^{(k)2})\\
\end{split}\vspace{-3em}
\label{eq:likelihood}
\end{align}
Though with Equation~\ref{eq:regularization} for regularization, we empirically find that the confidence learned with the identification loss still tend to be overconfident and hence cannot be directly used for Equation~\ref{eq:likelihood}, so we fine-tune the original confidence branch to predict $\sigma$ while fixing the other parts. We refer the readers to ~\cite{shi2019probabilistic} for the training details of fine-tuning.

\section{Implementation Details}
\label{sec:details}

\noindent\textbf{Training Details and Baseline}
All the models are implemented with Pytorch v1.1. We use the clean list from ArcFace~\cite{deng2018arcface} for MS-Celeb-1M~\cite{guo2016msceleb} as training data. After cleaning the overlapped subjects with the testing sets, we have 4.8M images of 76.5K classes. We use the method in~\cite{yu2016deep} for face alignment and crop all images into a size of $110\times110$. Random and center cropping are applied during training and testing, respectively, to transform the images into $100\times100$. We use the modified 100-layer ResNet in~\cite{deng2018arcface} as our architecture. The embedding size is $512$ for all models, and the features are split into $16$ groups for multi-embedding methods. The model $C$ is a linear classifier. The baseline models in the experiments are trained with CosFace loss function~\cite{wang2018cosface,wang2018additive}, which achieves state-of-the-art performance on general face recognition tasks. The models without domain augmentation are trained for $18$ epochs and models with domain augmentation are trained for $27$ epochs to ensure convergence. We empirically set $\lambda_{reg}$, $\lambda_{cls}$ and $\lambda_{adv}$ as 0.001, 2.0 and 2.0, respectively. The margin $m$ is empirically set to $30$. For non-augmentable variations, we choose $T=3$ attributes, namely smiling, young and gender.

\noindent\textbf{Variation Augmentation}
For the low-resolution, we use Gaussian blur with a kernel size between $3$ and $11$. For the occlusion, we split the images into $7\times7$ blocks and randomly replace some blocks with black masks. (3) For pose augmentation, we use PRNet~\cite{feng2018joint} to fit the 3D model of near-frontal faces in the dataset and rotate them into a yaw degree between $40^\circ$ and $60^\circ$. All the augmentations are randomly combined with a probability of $30\%$ for each.

\section{Experiments}
\label{sec:experiments}
In this section, we firstly introduce different types of datasets reflecting different levels of variation. Different levels of variation indicate different image quality and thus lead to different performance. Then we conduct detailed ablation study over the proposed confidence-aware loss and all the proposed modules. Further, we show evaluation on those different types of testing datasets and compare to state-of-the-art methods. 
% Finally, we visualize the uncertainty heat map and provide the insights under different variations.
% \subsection{Implementation Details and Baseline}
% \label{sec:detail}
% All the models in the paper are implemented using Pytorch v1.1. Four and Eight GeForce GTX 1080 Ti GPUs are used for training models on CASIA-Webface~\cite{yi2014learning} and MS-Celeb-1M~\cite{guo2016msceleb}, respectively. We use the clean list provided in ArcFace~\cite{deng2018arcface} for the MS-celeb-1M dataset. After cleaning the overlapped subjects with LFW~\cite{LFWTech}, MegaFace~\cite{MegaFace} and IJB-A/C~\cite{IJBC}, we have 4.8M images of 76.5K classes. We use the method in~\cite{yu2016deep} for face alignment and crop all images into a size of $110\times110$. Random and center cropping are applied during training and testing, respectively, to transform the images into $100\times100$. We use the modified 100-layer ResNet in~\cite{deng2018arcface} as our architecture. The embedding size is $512$ for all models, and the features are split into $16$ groups for multi-embedding methods. The models without domain augmentation are trained for $18$ epochs and models with domain augmentation are trained for $27$ epochs to ensure convergence. The baseline models in the experiments are trained with CosFace loss function~\cite{wang2018cosface,wang2018additive}, which achieves state-of-the-art performance on general face recognition tasks.

\begin{figure}
    \centering
    \footnotesize
    \captionsetup{font=small}
    \begin{minipage}{0.32\linewidth}
    \includegraphics[width=0.33\linewidth]{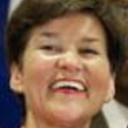}\hfill
    \includegraphics[width=0.33\linewidth]{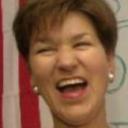}\hfill
    \includegraphics[width=0.33\linewidth]{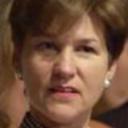}\hfill\\[-0.1em]
    \includegraphics[width=0.33\linewidth]{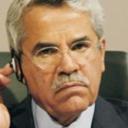}\hfill
    \includegraphics[width=0.33\linewidth]{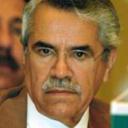}\hfill
    \includegraphics[width=0.33\linewidth]{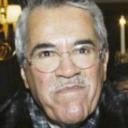}\hfill\\[-0.1em]
    \includegraphics[width=0.33\linewidth]{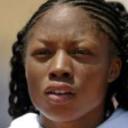}\hfill
    \includegraphics[width=0.33\linewidth]{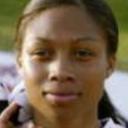}\hfill
    \includegraphics[width=0.33\linewidth]{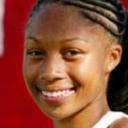}\hfill\\
    \vspace{-2.0em}\begin{center}(a) Type I\end{center}
    \end{minipage}\hfill
    \begin{minipage}{0.32\linewidth}
    \includegraphics[width=0.33\linewidth]{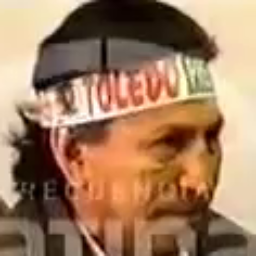}\hfill
    \includegraphics[width=0.33\linewidth]{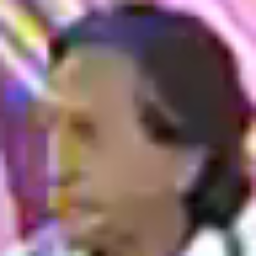}\hfill
    \includegraphics[width=0.33\linewidth]{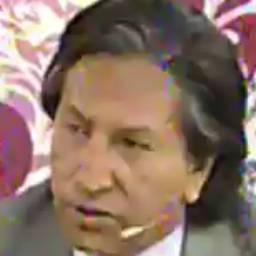}\hfill\\[-0.1em]
    \includegraphics[width=0.33\linewidth]{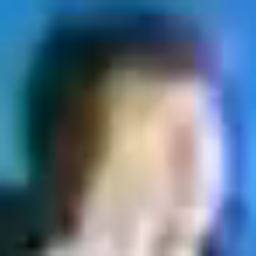}\hfill
    \includegraphics[width=0.33\linewidth]{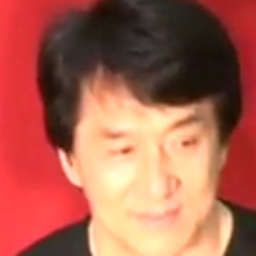}\hfill
    \includegraphics[width=0.33\linewidth]{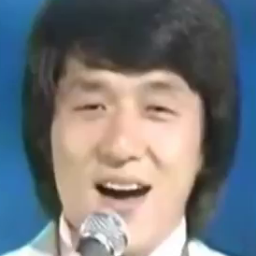}\hfill\\[-0.1em]
    \includegraphics[width=0.33\linewidth]{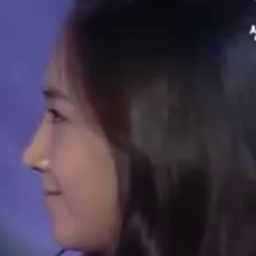}\hfill
    \includegraphics[width=0.33\linewidth]{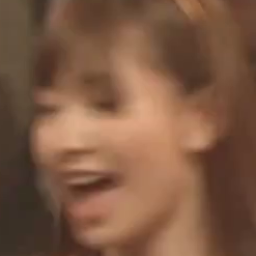}\hfill
    \includegraphics[width=0.33\linewidth]{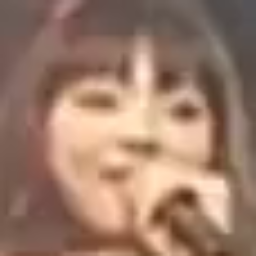}\hfill\\
    \vspace{-2.0em}\begin{center}(b) Type II\end{center}
    \end{minipage}\hfill
    \begin{minipage}{0.32\linewidth}
    \includegraphics[width=0.33\linewidth]{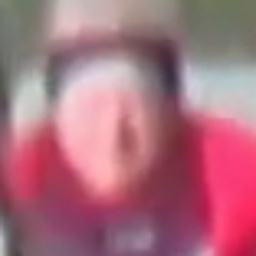}\hfill
    \includegraphics[width=0.33\linewidth]{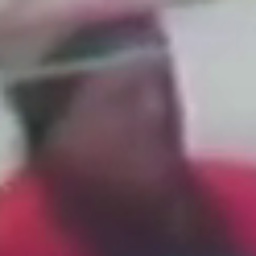}\hfill
    \includegraphics[width=0.33\linewidth]{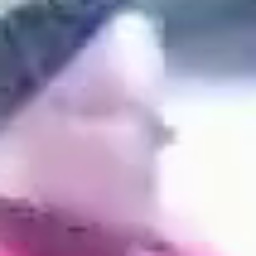}\hfill\\[-0.1em]
    \includegraphics[width=0.33\linewidth]{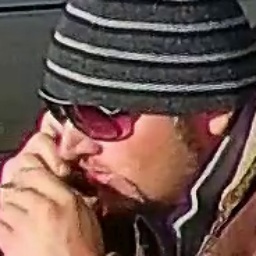}\hfill
    \includegraphics[width=0.33\linewidth]{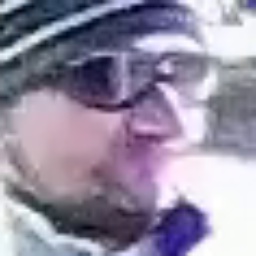}\hfill
    \includegraphics[width=0.33\linewidth]{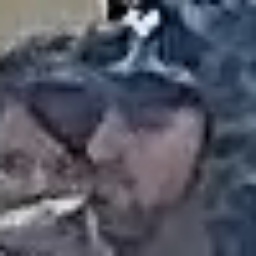}\hfill\\[-0.1em]
    \includegraphics[width=0.33\linewidth]{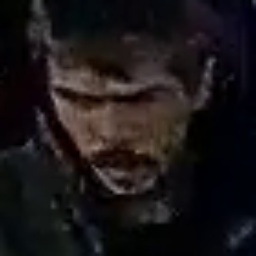}\hfill
    \includegraphics[width=0.33\linewidth]{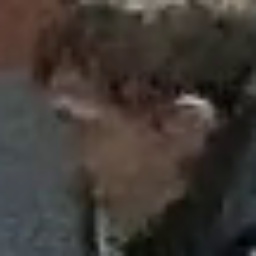}\hfill
    \includegraphics[width=0.33\linewidth]{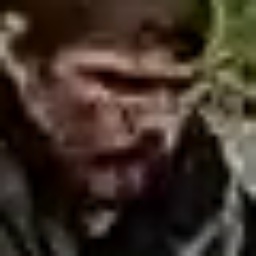}\hfill\\
    \vspace{-2.0em}\begin{center}(c) Type III\end{center}
    \end{minipage}\\
    \vspace{-0.9em}\caption{Examples of the three types of datasets. The images are sampled from LFW~\cite{LFWTech}, IJB-A~\cite{IJBA}, IJB-S~\cite{IJBS}, respectively.}\vspace{-1.6em}
    \label{fig:exp_dataset}
    \vspace{-2mm}
\end{figure}

\subsection{Datasets}
We evaluate our models on eight face recognition benchmarks, covering different real-world testing scenarios. The datasets are roughly categorized into three types based on the level of variations:

\begin{figure}[t]
    \captionsetup{font=footnotesize}
    \centering
    \subfloat[Baseline]{\includegraphics[width=0.49\linewidth,trim={1mm 6mm 2mm 0},clip]{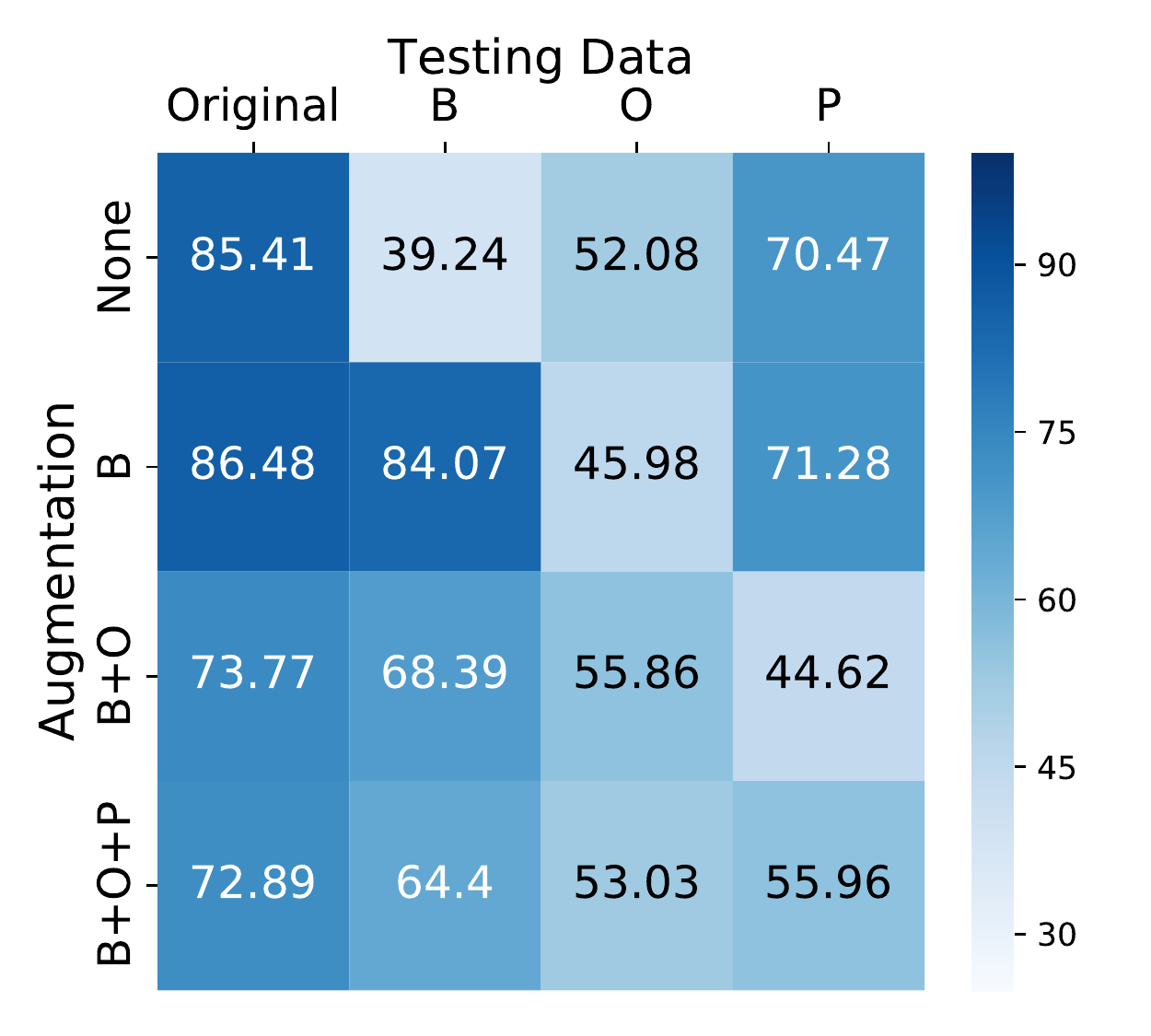}}\hfill
    \subfloat[Proposed]{\includegraphics[width=0.49\linewidth,trim={1mm 6mm 2mm 0},clip]{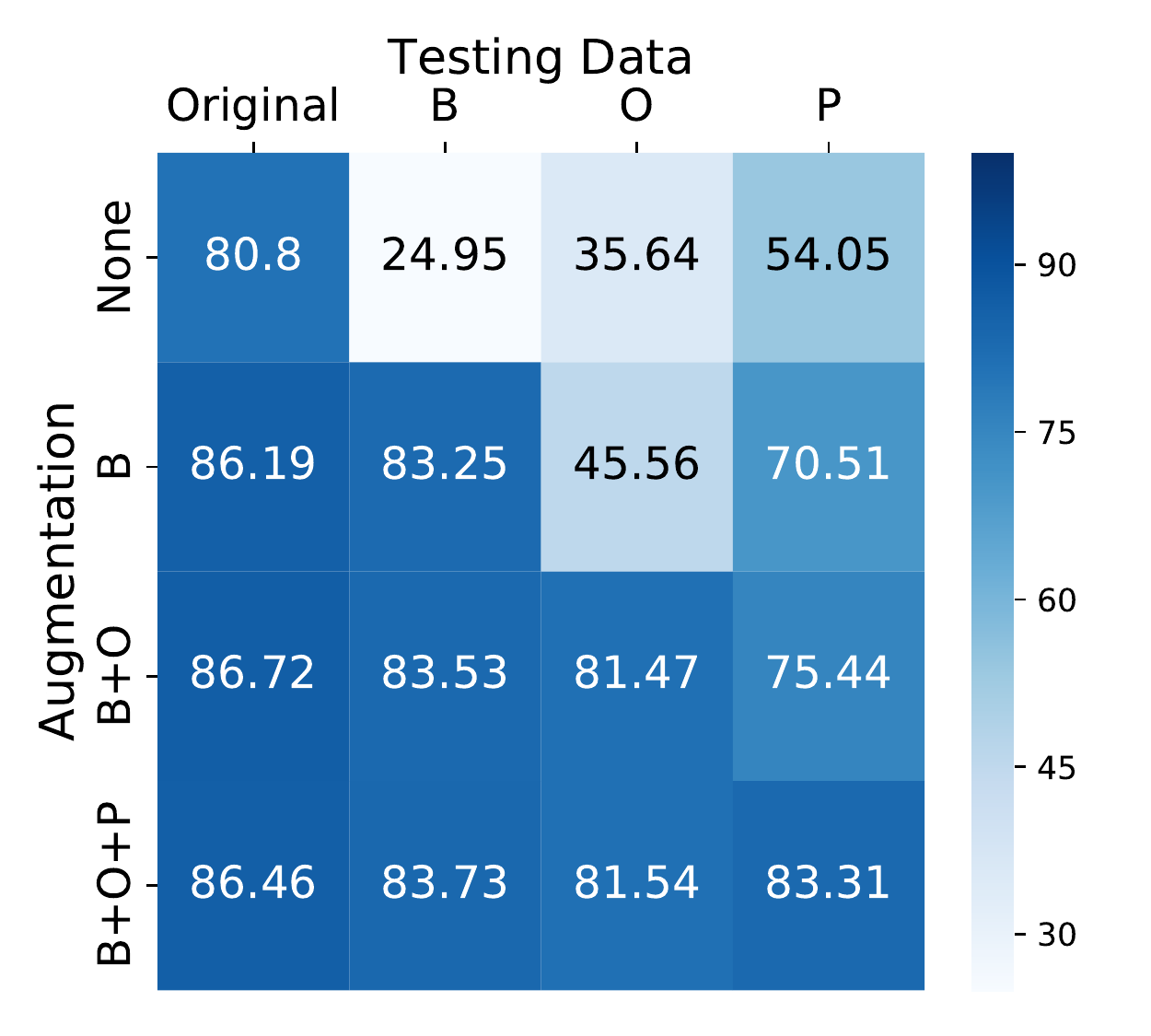}}
    \vspace{-0.5em}\caption{Testing results on synthetic data of different variations from IJB-A benchmark (TAR@FAR=0.01\%). Different rows correspond to different augmentation strategies during training. Columns are different synthetic testing data. ``B'', ``O'', ``P'' represents ``Blur'', ``Occlusion'' and ``Pose'', respectively. The performance of the proposed method is improved in a monotonous way with more augmentations being added.}
    \label{fig:augmentation}
    \vspace{-4mm}
\end{figure}

\vspace{-1.0em}\paragraph{Type I: Limited Variation} LFW~\cite{LFWTech}, CFP~\cite{CFP}, YTF~\cite{YTF} and MegaFace~\cite{MegaFace} are four widely applied benchmarks for general face recognition. We believe the variations in those datasets are limited, as only one or few of the variations being presented. In particular, YTF are video samples with relatively lower resolution; CFP~\cite{CFP} are face images with large pose variation but of high resolution; MegaFace includes 1 million distractors crawled from internet while its labeled images are all high-quality frontal faces from FaceScrub dataset~\cite{FaceScrub}. For both LFW and YTF, we use the unrestricted verification protocol. For CFP, we focus on the frontal-profile (FP) protocol. We test on both verification and identification protocols of MegaFace.

\vspace{-1.0em}\paragraph{Type II: Mixed Quality} IJB-A~\cite{IJBA} and IJB-C~\cite{IJBC} include both high quality celebrity photos taken from the wild and low quality video frames with large variations of illumination, occlusion, head pose, etc. We test on both verification and identification protocols of the two benchmarks.

\vspace{-1.0em}\paragraph{Type III: Low Quality} We test on TinyFace~\cite{TinyFace} and IJB-S~\cite{IJBS}, two extremely challenging benchmarks that are mainly composed of low-quality face images. In particular, TinyFace only consists of low-resolution face images captured in the wild, which also includes other variations such as occlusion and pose. IJB-S is a video face recognition dataset, where all images are video frames captured by surveillance cameras except a few high-quality registration photos for each person. Example images of the three types of datasets are shown in Figure~\ref{fig:exp_dataset}.

% \begin{table}[t]
% \captionsetup{font=footnotesize}
% \newcommand{\mr}[1]{\multirow{4}{*}{#1}}
% \setlength{\tabcolsep}{5pt}
% \footnotesize
% \begin{center}
% \resizebox{1.0\columnwidth}{!}{%
% \begin{tabularx}{1.0\linewidth}{X||c|c|c|c|c}
% \Xhline{2\arrayrulewidth}
% Method & Augmentation & Original & Blur & Occlusion & Pose \\
% \Xhline{2\arrayrulewidth}          
% \mr{Baseline}   & -     & 85.41 & 39.24 & 52.08 & 70.47 \\         
%                 & B     & 86.48 & 84.07 & 45.98 & 71.28\\         
%                 & B+O   & 73.77 for the IJB-A fine-tuning. It is not working well as expected.& 68.39 & 55.86 & 44.62\\         
%                 & B+O+P & 72.89 & 64.40 & 53.03 & 55.96\\\hline    
% \mr{Proposed}   & -     & 80.80 & 24.95 & 35.64 & 54.05\\         
%                 & B     & 86.19 & 83.25 & 45.56 & 70.51\\         
%                 & B+O   & 86.72 & 83.53 & 81.47 & 75.44\\         
%                 & B+O+P & 86.46 & 83.73 & 81.54 & 83.31\\
% \Xhline{2\arrayrulewidth}
% \end{tabularx}
% }
% \vspace{-0.9em}\caption{Testing results on synthetic data of different domains on IJB-A benchmark (TAR@FAR=0.01\%). Different rows correspond to different augmentation strategies during training and columns are different synthetic testing data. ``B'', ``O'', ``P'' represents ``Blur'', ``Occlusion'' and ``Pose'', respectively.}
% \label{tab:augmentation}
% \end{center}
% \end{table}

\begin{figure}[t]\vspace{-1.5em}
    \captionsetup{font=small}
    \centering
    \subfloat[Baseline]{\includegraphics[width=0.49\linewidth]{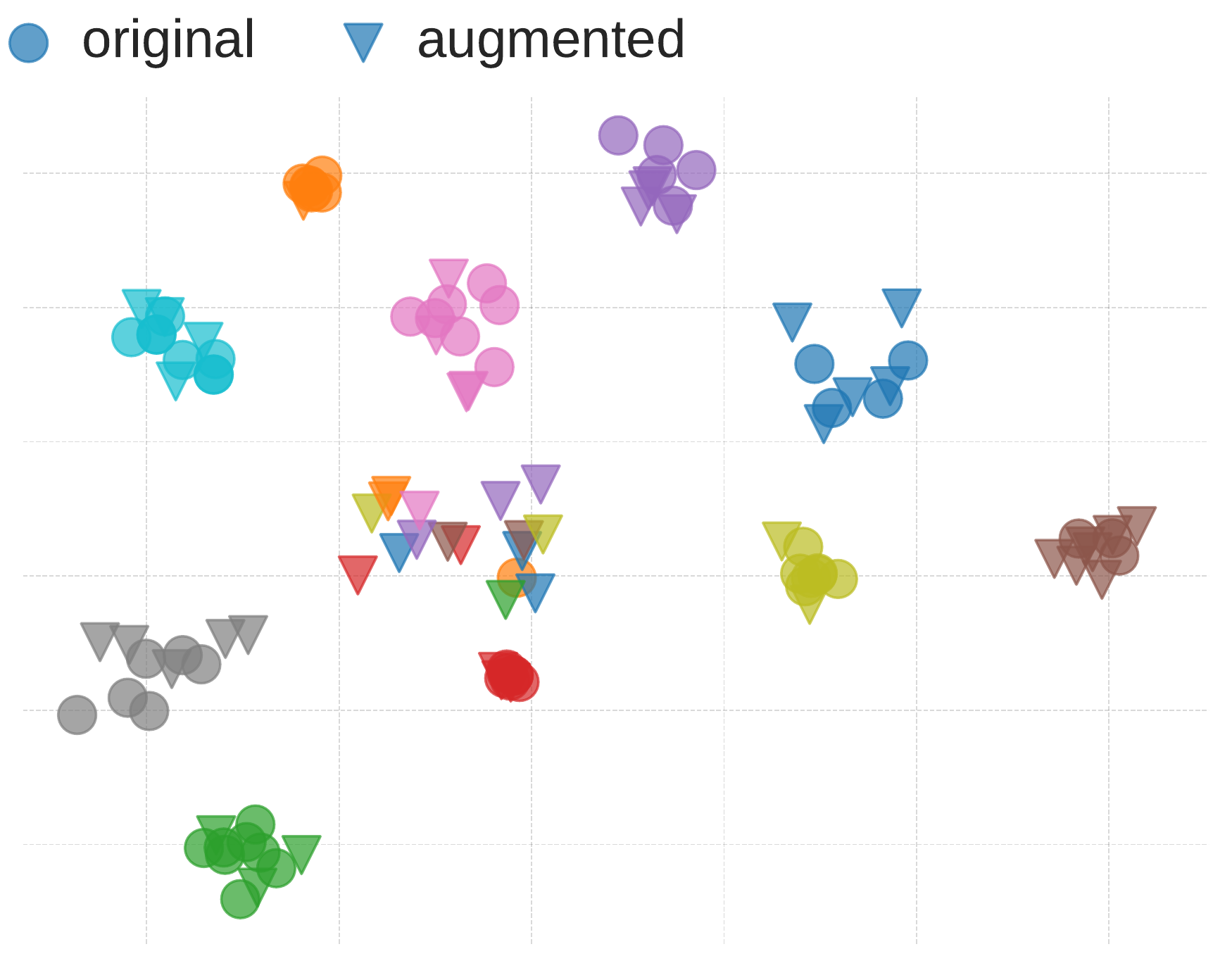}}\hfill
    \subfloat[Proposed]{\includegraphics[width=0.49\linewidth]{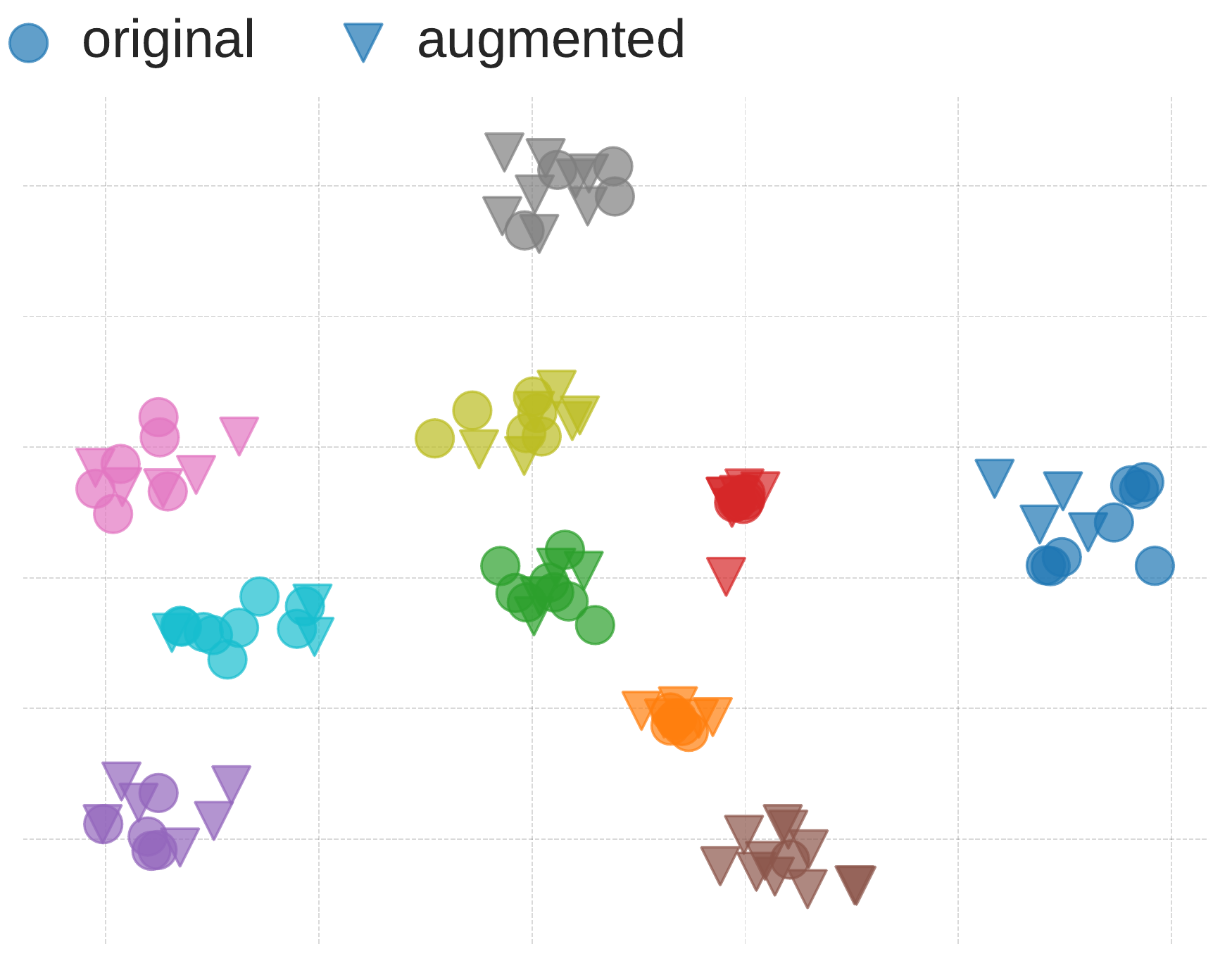}}
    \vspace{-0.5em}\caption{t-SNE visualization of the features in a 2D space. Colors indicate the identities. Original training samples and augmented training samples are shown in circle and triangle, respectively.}
    \label{fig:tsne_mixed_baseline}
    \vspace{-4mm}
\end{figure}

\begin{figure}[t]
    \captionsetup{font=small}
    \centering
    \includegraphics[width=0.85\linewidth]{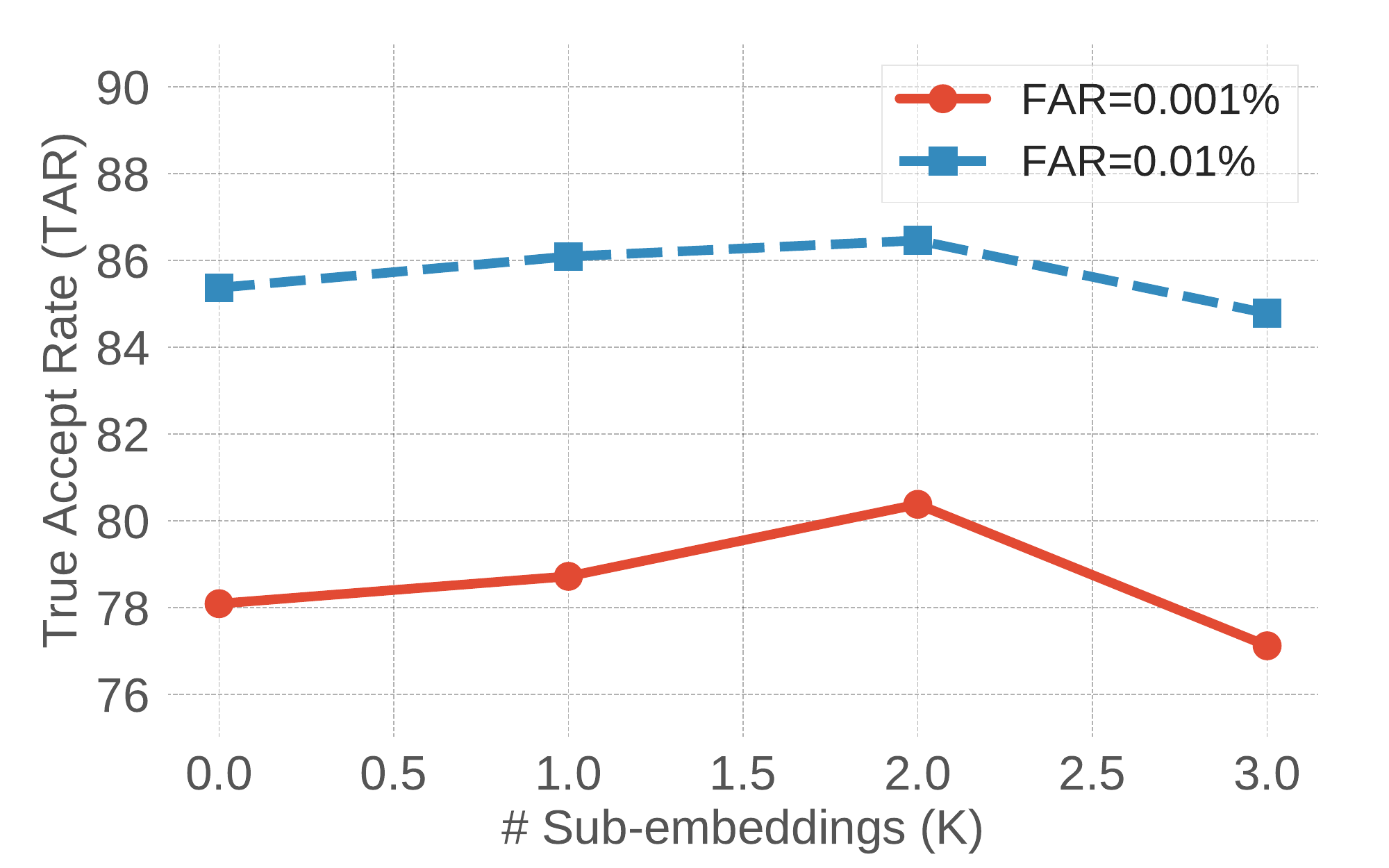}
    \vspace{-1.0em}\caption{Performance change with respect to difference choice of K.}
    \label{fig:plt_ijba_ngroup}
    \vspace{-6mm}
\end{figure}

\subsection{Ablation Study}

\begin{table*}[t]
\captionsetup{font=footnotesize}
\newcommand{\mr}[1]{\multirow{2}{*}{#1}}
\footnotesize
\setlength{\tabcolsep}{5.5pt}
\begin{center}
\begin{tabularx}{1.00\linewidth}{X || c|c|c|c|c || c|c|c|c|c|c|c|c}
\Xhline{2\arrayrulewidth}
Model               & \multicolumn{5}{c||}{Method} & LFW & CFP-FP & \multicolumn{2}{c|}{IJB-A (TAR@FAR)} & \multicolumn{2}{c|}{TinyFace} & \multicolumn{2}{c}{IJB-S} \\\hline
            & VA & CI & ME & DE & PA    & Accuracy & Accuracy       & FAR=0.001\% & FAR=0.01\% & Rank1 & Rank5 & Rank1 & Rank 5\\
\Xhline{2\arrayrulewidth}
Baseline    & & & & &                                                   & 99.75 & 98.16 & 82.20 & 93.05 & 46.75 & 51.79 & 37.14 & 46.75\\\hline
A & \checkmark & & & &                                                  & 99.70 & 98.35 & 82.42 & 93.86  & 55.26 & 59.04 & 51.27 & 58.94\\\hline
B & \checkmark & \checkmark & & &                                       & 99.78 & 98.30 & 94.70 & 96.02 & 57.11 & 63.09 & 59.87 & 66.90\\\hline
\mr{C} & \checkmark & \checkmark & \checkmark & &                       & 99.77 & 98.50 & 94.75 & 96.27 & 57.30 & 63.73 & 59.66 & 66.30\\
 & \checkmark & \checkmark & \checkmark & & \checkmark                  & 99.78 & \textbf{98.66} & \textbf{96.10} & 97.29 & 55.04 & 60.97 & 59.71 & 66.32\\\hline
\mr{D} & & \checkmark & \checkmark & \checkmark &                       & 99.65 & 97.77 & 80.06 & 92.14 & 34.76 & 39.86 & 29.87 & 40.69\\
 & & \checkmark & \checkmark & \checkmark   & \checkmark                & 99.68 & 98.00 & 94.37 & 96.42 & 35.05 & 40.13 & 50.00 & 56.27\\\hline
% \mr{E (partial)} & \checkmark & \checkmark & \checkmark & \checkmark &  & 99.68 & 98.51 & 94.72 & 96.38 & 57.22 & 62.37 & 60.13 & 66.14\\
%  & \checkmark & \checkmark & \checkmark & \checkmark & \checkmark       & 99.77 & \textbf{98.71} & \textbf{96.23} & \textbf{97.37} & 54.99 & 61.32 & \textbf{62.22} & 67.03\\\hline
 \mr{E (all)} & \checkmark & \checkmark & \checkmark & \checkmark &     & 99.75 & 98.30 & 95.00 & 96.27 & 61.32 & 66.34 & 60.74 & 66.59\\
 & \checkmark & \checkmark & \checkmark & \checkmark & \checkmark       & \textbf{99.78} & 98.64 & 96.00 & \textbf{97.33} & \textbf{63.89} & \textbf{68.67} & \textbf{61.98} & \textbf{67.12}\\
\Xhline{2\arrayrulewidth}
\end{tabularx}
\vspace{-0.7em}\caption{Ablation study over the whole framework. ``VA'' indicates ``Variation Augmentation'' (Section~\ref{sec:methods}), ``CI'' indicates ``Confidence-aware Identification loss'' (Section~\ref{sec:confidence_loss}), ``ME'' indicates ``Multiple Embeddings'' (Section~\ref{sec:method_decorrelation}), ``DE'' indicates ``Decorrelated Embeddings'' (Section~\ref{sec:method_decorrelation}) and ``PA'' indicates ``Probabilistic Aggregation''. (Section~\ref{sec:method_aggregation}). E(all) uses all the proposed modules. }
\label{tab:ablation}
\vspace{-6mm}
\end{center}
\end{table*}

\begin{table}[t]
\captionsetup{font=footnotesize}
\newcommand{\mr}[1]{\multirow{2}{*}{#1}}
\setlength{\tabcolsep}{5.5pt}
\footnotesize
\begin{center}
\resizebox{1.0\columnwidth}{!}{%
\begin{tabularx}{1.0\linewidth}{X||c|c|c|c|c}
\Xhline{2\arrayrulewidth}
\mr{Method}                             & \mr{LFW}  & \mr{YTF}  & \mr{CFP-FP}   & \multicolumn{2}{c}{MF1}   \\\cline{5-6}
                                        &  & & & Rank1  & Veri.            \\
\Xhline{2\arrayrulewidth}          
% DeepFace+~\cite{taigman2014deepface}    & 97.35     & 91.4      & -         & -  & -        \\
FaceNet~\cite{schroff2015facenet}       & 99.63     & 95.1      & -         & -  & -        \\
% DeepID2+~\cite{deepid2plus}             & 99.47     & 93.2      & -         & -  & -        \\
CenterFace~\cite{wen2016discriminative} & 99.28     & 94.9      & -         & 65.23 & 76.52    \\
SphereFace~\cite{liu2017sphereface}     & 99.42     & 95.0      &  -        & 75.77 & 89.14   \\
ArcFace~\cite{deng2018arcface}          & 99.83     & 98.02     & 98.37     & 81.03   & 96.98    \\
CosFace~\cite{wang2018cosface}          & 99.73     & 97.6      & -         & 77.11 & 89.88    \\\hline
% L2-Face~\cite{ranjan2017l2}             & 99.78     & 96.08     & -         & -  & -        \\
% PFE~\cite{shi2019probabilistic}         & 99.82     & 97.36     & 93.34     & 78.95   & 92.51 \\\hline
Ours (Baseline)                         & 99.75     & 97.16     & 98.16     & 80.03 & 95.54 \\
Ours (Baseline+VA)                      & 99.70     & 97.10     & 98.36     & 78.10 & 94.31 \\
% Ours (all)                              & 99.77     & 97.74   & 98.71     & 79.16 & 95.04 \\
% Ours (all) + PA                       & 99.77     & 97.64     & 98.71     & 77.34 & 94.06 \\
Ours (all)                              & 99.75     & 97.68     & 98.30     & 79.10 & 94.92 \\
Ours (all) + PA                         & 99.78     & 97.92     & 98.64     & 78.60 & 95.04 \\
\Xhline{2\arrayrulewidth}
\end{tabularx}
}
\vspace{-0.9em}\caption{Our method compared to state-of-the-art methods on Type I datasets. The MegaFace verification rates are computed at FAR=$0.0001\%$. ``-'' indicates that the author did not report the performance on the corresponding protocol.}
\label{tab:lfw}
\vspace{-8mm}
\end{center}
\end{table}

\begin{table*}[t]
\captionsetup{font=footnotesize}
\newcommand{\mr}[1]{\multirow{2}{*}{#1}}
\footnotesize
\setlength{\tabcolsep}{3.0pt}
\begin{threeparttable}
\begin{tabularx}{1.00\linewidth}{X ||c|c|c|c| |c|c|c|c| |c|c|c}
\Xhline{2\arrayrulewidth}
\mr{Method}                             & \multicolumn{2}{c|}{IJB-A (Vrf)} & \multicolumn{2}{c||}{IJB-A (Idt)} & \multicolumn{2}{c|}{IJB-C (Vrf)} & \multicolumn{2}{c||}{IJB-C (Idt)}  & \multicolumn{3}{c}{IJB-S (S2B)} \\\cline{2-12}
        & FAR=0.001\%   & FAR=0.01\%  & Rank1 & Rank5 & FAR=0.001\%   & FAR=0.01\%  & Rank1 & Rank5  & Rank1   & Rank5 & FPIR=1\% \\
\Xhline{2\arrayrulewidth}          
% TPE~\cite{sankaranarayanan2016triplet}\tnote{*}     & - & 81.3$\pm$2.0  & 93.2$\pm$1.0 & 97.7$\pm$0.5 &- &- &- &- &- &- &-\\
NAN~\cite{yang2017neural}\tnote{*}                  & - & 88.1$\pm$1.1  & 95.8$\pm$0.5 & 98.0$\pm$0.5 &- &- &- &- &- &- &-\\
% QAN~\cite{liu2017quality}\tnote{*}                  & - & 89.3$\pm$3.9  & - & -  &- &- &- &- &- &- &-\\
L2-Face~\cite{ranjan2017l2}\tnote{*}                &  90.9$\pm$0.7     & 94.3$\pm$0.5 & 97.3$\pm$0.5 & 98.8$\pm$0.3  &- &- &- &- &- &- &-\\
DA-GAN~\cite{zhao2018DAGAN}\tnote{*}                &  94.6$\pm$0.1     & 97.3$\pm$0.5 & \textbf{99.0$\pm$0.2} & \textbf{99.5$\pm$0.3}  &- &- &- &- &- &- &-\\
Cao~\etal~\cite{cao2018vggface2}                    & - & 92.1$\pm$1.4  & 98.2$\pm$0.4 & 99.3$\pm$0.2 & 76.8  & 86.2 & 91.4 & 95.1 & - & - & - \\
Multicolumn~\cite{xie2018multicolumn}               & - & 92.0$\pm$1.3  & - & - & 77.1  & 86.2  & - & - & - & - & - \\
% DCN~\cite{}                                         & - & - & - & - & -     & 88.5 & - & - & -  & - & - \\
PFE~\cite{shi2019probabilistic}                     & - & 95.3$\pm$0.3  & - & - & 89.6  & 93.3  & - & - & 50.16 & 58.33 & 31.88 \\
ArcFace~\cite{deng2018arcface}\tnote{+}             & 93.7$\pm$1.0      & 94.2$\pm$0.8  & 97.0$\pm$0.6 & 97.9$\pm$0.4  & 93.5 & 95.8   & 95.87 & \textbf{97.27} & 57.36 & 64.95 & 41.23 \\\hline
Ours (Baseline)                                     & 82.6$\pm$8.3      & 93.3$\pm$3.0  & 95.5$\pm$0.7 & 96.9$\pm$0.6 & 43.9  & 86.7  &  89.85 & 90.86 & 37.14 & 46.75 & 24.75 \\ 
Ours (Baseline + VA)                                & 82.4$\pm$8.1      & 93.9$\pm$3.5  & 95.8$\pm$0.6 & 97.2$\pm$0.5 & 47.6  & 90.6  &  90.16 & 91.20 & 51.27 & 58.94 & 31.19 \\ 
% Ours                                                & 94.2$\pm$1.9      & 96.5$\pm$0.5  & 97.4$\pm$0.4 & 98.2$\pm$0.4 & 91.6  & 94.2  &  94.18 & 95.84 & 60.82 & 67.67 & 35.90 \\ 
Ours (all)                                          & 95.0$\pm$0.9      & 96.3$\pm$0.6  & 97.5$\pm$0.4 & 98.4$\pm$0.4 & 91.6  & 93.7  &  94.39 & 96.08 & 60.74 & 66.59 & 37.11 \\ 
Ours (all) + PA                                     & \textbf{96.0$\pm$0.8}      & \textbf{97.3$\pm$0.4}  & 97.5$\pm$0.3 & 98.4$\pm$0.3 & \textbf{95.0}  & \textbf{96.6}  &  \textbf{96.00} & 97.06 & \textbf{61.98} & \textbf{67.12} & \textbf{42.73}\\ 
% Ours \tnote{*}                                      & / & /& /& / & - & - & - & - & - & - & -\\ 
\Xhline{2\arrayrulewidth}
\end{tabularx}
% \begin{tablenotes}
% \item[*] indicates fine-tuning on the target dataset.
% \end{tablenotes}
\vspace{-0.7em}\caption{Our model compared to state-of-the-art methods on IJB-A, IJB-C and IJB-S. ``-'' indicates that the author did not report the performance on the corresponding protocol. ``*'' indicates fine-tuning on the target dataset during evaluation on IJB-A benchmark and ``+'' indicates the testing performance by using the released models from corresponding authors.}
\label{tab:ijb_all}
\end{threeparttable}
\vspace{-3mm}
\end{table*}

\subsubsection{Effect of Confidence-aware Learning}
%We first analyze the effect of confidence-guided learning and show that it is a necessary module when training on a large-variation dataset. 
We train a set of models by gradually adding the nameable variations. The ``Baseline'' model is an 18-layer ResNet trained on a randomly selected subset of MS-Celeb-1M (0.6M images). The ``Proposed'' model is trained with the confidence-aware identification loss and $K=16$ embedding groups. As a controlled experiment, we apply the same type of augmentation on IJB-A dataset to synthesize testing data of the corresponding variations. In Figure~\ref{fig:augmentation}, ``Baseline'' model shows decreasing performance when gradually adding new variations as in the grid going down from top row to bottom row. In comparison, the proposed method shows improving performance when adding new variations from top to bottom, which highlights the effect of our confidence-aware representation learning and it further allows to add more variations into the framework training.

We also visualize the features with t-SNE projected onto 2D embedding space. Figure~\ref{fig:tsne_mixed_baseline} shows that for ``Baseline'' model, with different variation augmentations, the features actually are mixed and thus are erroneous for recognition. While for ``Proposed'' model, different variation augmentation generated samples are still clustered together to its original samples, which indicates that identity is well preserved. Under the same settings as above, we also show the effect of using different number of groups in Figure~\ref{fig:plt_ijba_ngroup}. At the beginning, splitting the embedding space into more groups increases performance for both TARs. When the size of each sub-embedding becomes too small, the performance starts to drop because of the limited capacity for each sub-embedding.

\subsubsection{Ablation on All Modules}
\label{sec:exp_ablation}
We investigate each module's effect by looking into the ablative models in Table~\ref{tab:ablation}. Starting from the baseline, model A is trained with variation augmentation. Based on model A, we add confidence-aware identification loss to obtain model B. Model C is further trained by setting up multiple sub-embeddings. In model E, we further added the decorrelation loss. 
We also compare with a Model D with all the modules except variation augmentation. Model C, D and E, which have multiple embeddings, are tested w/ and w/o probabilistic aggregation (PA). The methods are tested on two type I datasets (LFW and CFP-FP), one type-II dataset (IJB-A) and one type-III dataset (TinyFace).

Shown in Table~\ref{tab:ablation}, compared to baseline, adding variation augmentation improves performance on CFP-FP, TinyFace, and IJBA. These datasets present exactly the variations introduced by data augmentation, i.e., pose variation and low resolution. However, the performance on LFW fluctuates from baseline as LFW is mostly good quality images with few variations. In comparison, model B and C are able to reduce the negative impact of hard examples introduced by data augmentation and leads to consistent performance boost across all benchmarks. Meanwhile, we observe that splitting into multiple sub-embeddings alone does not improve (compare B to C first row) significantly, which can be explained by the strongly correlated confidence among the sub-embeddings (see Figure~\ref{fig:correlation}). Nevertheless, with the decorrelation loss and probabilistic aggregation, different sub-embeddings are able to learn and combine complementary features to further boost the performance, i.e., the performance in the second row of Model E is consistently better than its first row. 
% Further compare E (partial) to F (all), we observe that adding more variations for decorrelation loss boosts the performance, which is also verified from the controlled setting in Figure~\ref{fig:augmentation}. We draw a take-home that adding more variations to cope with decorrelation loss is beneficial.

\subsection{Evaluation on General Datasets}
%To comprehensively evaluate the proposed method, we test our models on more benchmarks and compare it with the state-of-the-art methods. Table~\ref{tab:lfw} shows the performance on four Type I datasets. The images in these datasets represent a similar domain as the original training dataset and are mostly of high quality. Therefore, the performance of state-of-the-art systems are known to be saturated on these benchmarks. It can be seen that even our baseline model achieves very good performance on all of the four datasets. The baseline with domain augmentation achieves a slightly worse performance. In comparison, our method does a better job in terms of maintaining the performance of the baseline model.
We compare our method with state-of-the-art methods on general face recognition datasets, i.e., those Type I datasets with limited variation and high quality. Since the testing images are mostly with good quality, there is limited advantage of our method which is designed to deal with larger variations. Even though, shown in Table~\ref{tab:lfw}, our method still stands on top being better than most of the methods while slightly worse than ArcFace. Notice that our baseline model already achieves good performance across all the testing sets. It actually verifies that the type I testing sets do not show significant domain gap from the training set, where even without variation augmentation or embedding decorrelation, the straight training can lead to good performance.

\begin{figure}[t]
    \centering
    \captionsetup{font=small}
    \newcolumntype{Y}{>{\centering\arraybackslash}X}
    \captionsetup[subfloat]{captionskip=1pt}
    \setlength{\tabcolsep}{0.0pt}
    \footnotesize
    \begin{tabularx}{1.00\linewidth}{p{2.0pt}Y|Y|Y|Yp{18.0pt}}
       & High-quality & Blur & Occlusion & Large-pose & \\[-0.2em]
    \end{tabularx}
    \includegraphics[height=93pt]{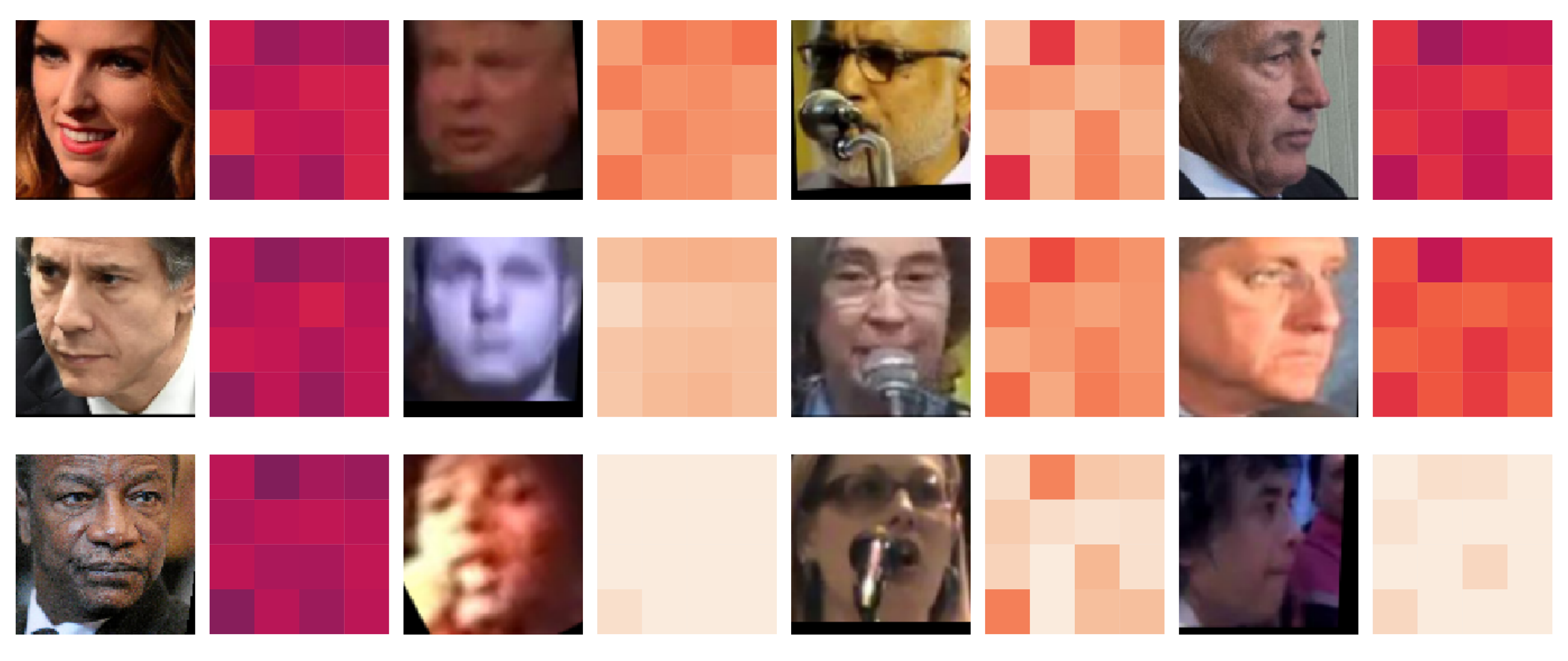}\hfill
    \includegraphics[height=93pt]{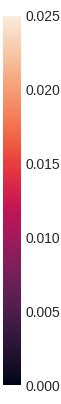}
    \vspace{-0.5em}\caption{Heatmap visualization of sub-embedding uncertainty on different types of images from IJB-C dataset, shown on the right of each face image. 16 values are arranged in 4$\times$4 grids (no spatial meaning). Brighter color indicates higher uncertainty.}\vspace{-1.5em}
    \label{fig:visual}
    \vspace{-2mm}
\end{figure}
\subsection{Evaluation on Mixed/Low Quality Datasets}
When evaluating on more challenging datasets, those state-of-the-art general methods encounter performance drop as the challenging datasets present large variations and thus large domain gap from the good quality training datasets.
Table~\ref{tab:ijb_all} shows the performance on three challenging benchmarks: IJB-A, IJB-C and IJB-S. The proposed model achieves consistently better results than the state-of-the-arts. 
In particular, simply adding variation augmentation (``Ours (Baseline + VA)'') actually leads to a worse performance on IJB-A and IJB-C. When variation augmentation is combined with our proposed modules (``Ours''), significant performance boost is achieved. Further adding PA with ``Ours'', we achieve even better performance across all datasets and protocols.
Notice that IJB-A is a cross-validation protocol. Many works fine-tune on training splits before evaluation (shown with ``*''). Even though, our method without fine-tuning still outperforms the state-of-the-art methods with significant margin on IJB-A verification protocol, which suggests that our method indeed learns the representation towards dealing with unseen variations.

Table~\ref{tab:ijb_all} last column shows the evaluation on IJB-S, which is so far the most challenging benchmark targeting real surveillance scenario with severe poor quality images. We show the Surveillance-to-Booking (S2B) protocol of IJB-S. Other protocol results can be found in supplementary. As IJB-S is recently released, there are few studies that have evaluated on this dataset. To comprehensively evaluate our model, we use the publicly released models from ArcFace~\cite{deng2018arcface} for comparison. Our method achieves consistently better performance across Rank-1 and Rank-5 identification protocol. For TinyFace, as in Table~\ref{tab:ablation}, we achieve $63.89\%$, $68.67\%$ rank-1 and rank-5 accuracy, where \cite{TinyFace} reports $44.80\%$, $60.40\%$, and ArcFace achieves $47.39\%$, $52.28\%$.
Combining Table~\ref{tab:lfw}, our method achieves top level accuracy on general recognition datasets and significantly better accuracy on challenging datasets, which demonstrates the advantage in dealing with extreme or unseen variations.

\noindent\textbf{Uncertainty Visualization} Figure~\ref{fig:visual} shows the $16$ sub-embeddings' uncertainty score reshaped into $4\times 4$ grids. High-quality and low-quality sub-embeddings are shown in dark, light colors respectively. For different variations, the uncertainty map shows different patterns.

% \subsection{Analysis}

% \begin{figure*}[t]
%     \captionsetup{font=small}
%     \centering
%     \includegraphics[width=0.7\linewidth]{fig/images_clustering_attention.png}
%     \caption{Example images from the training dataset by clustering on the confidence vectors.}
%     \label{fig:clustering_attention}
% \end{figure*}
% \paragraph{Clustering on confidence vectors} To understand the role played by the embedding confidence, we apply a k-means clustering the confidence vectors of the training samples and show example images from 9 clusters in Figure~\ref{fig:clustering_attention}. Images of different domains tend to be categorized into different clusters, which indicates similar confidence distributions inside the same domain while different confidence between different domains. For example, images that are of extremely low quality usually receive low confidence on all embedding groups and hence are assigned into the same cluster, similarly for high-quality photos. The results indirectly implies that different embedding groups are learning features associated with different domains.

\section{Conclusion}
In this work, we propose a universal face representation learning framework to recognize faces under all kinds of variations. We firstly introduce three nameable variations into MS-Celeb-1M training set via data augmentation. Traditional methods encounter convergence problem when directly feeding the augmented hard examples into training. We propose a confidence-aware representation learning by partitioning the embedding into multiple sub-embeddings and relaxing the confidence to be sample and sub-embedding specific. 
Further, the variation classification and variation adversarial loss are proposed to decorrelate the sub-embeddings. By formulating the inference with an uncertainty model, the sub-embeddings are aggregated properly.
%
%Besides the nameable variations, we further mine out the unnameable variations together to regularize the sub-embeddings with less correlation. 
%
Experimental results show that our method achieves top performance on general benchmarks such as LFW and MegaFace, and significantly better accuracy on challenging benchmarks such as IJB-A, IJB-C and IJB-S.

{\small
\bibliographystyle{ieee_fullname}
\bibliography{egbib}
}

\clearpage
\appendix

\addtocounter{equation}{14}

\section{Proofs}
\subsection{Confidence-aware Identification Loss}
\subsubsection{Single Embedding}
Let $\cZ$ denotes the latent embedding space and $\bz$ a variable from the $\cZ$. Different $\bz$ represents different facial appearance. Given a face image $\bx$, the network $\theta$ estimates the encoded appearance $p_{\theta}(\bz|\bx)=\mathcal{N}(\bz;\bbf_i,\sigma^2_i\mathbf{I})$ where $\bbf_i$ is the embedded feature vector while $\sigma_i^2$ is the uncertainty of the representation.
Let $y$ denotes the identity label and $C$ the number of identities. For each class $j\in\{1,2,\dots,C\}$, we maintain a prototype vector $\bw_j$, which represents the intrinsic appearance of the $j^{\text{th}}$ identity in the latent space. In other words, $p(\bz|y=j)=\delta(\bz-\bw_j)$, where $\delta$ is the Dirac delta function. Assuming an non-informative prior $p(\bz)$, the likelihood of $\bx_i$ being a sample of $j^{\text{th}}$ class is given by:
\begin{align}
\begin{split}
    p(\bx_i|y=j) & = \int{p(\bx_i|\bz)p(\bz|y=j)d\bz} \\
                & = \int{\frac{p_\theta(\bz|\bx_i)p(\bx_i)}{p(\bz)}\delta(\bz-\bw_j)d\bz} \\
                & = p_\theta(\bw_j|\bx_i)p(\bx_i) \\
                & \propto p_\theta(\bw_j|\bx_i)
\end{split}
\end{align}
where
\begin{equation}
p_\theta(\bw_j|\bx_i) = \frac{1}{(2\pi\sigma^{2}_i)^{\frac{D}{2}}}\exp(-\frac{\norm{\bbf_i-\bw_j}^2}{2\sigma^{2}_i}).
\end{equation}
Therefore, the posterior probability of $\bx_i$ belonging to the $j^{\text{th}}$ class is:
\begin{align}
%\begin{split}
%p(y_i=j|\bx_i) =& \frac{p_\theta(\bw_j|\bx_i)}{\sum_{c=1}^{C}{p_\theta(\bw_c|\bx_i)}} \\
p(y=j|\bx_i) = & \frac{p(\bx_i|y=j)p(y=j)}{\sum_{c=1}^{N}{p(\bx_i|y=c)p(y=c)}} \\
= & \frac{p_{\theta}(\bw_j|\bx_i)}{\sum_{c=1}^{N}{p_{\theta}(\bw_c|\bx_i)}} \\
= & \frac{\exp(-\frac{\norm{\mathbf{f}_i-\bw_j}^2}{2\sigma^{2}_i})}{\sum_{c=1}^{N}{\exp(-\frac{\norm{\mathbf{f}_i-\bw_c}^2}{2\sigma^{2}_i})}},
%\end{split}
\end{align}
which is the Equation~(4) in the main paper.

\subsubsection{Multiple Sub-embeddings}
For a sub-embedding network, the likelihood function becomes:
\begin{equation}
p_\theta(\bw_j|\bx_i) = \prod_{k=1}^{K}\frac{1}{(2\pi\sigma^{(k)2}_i)^{\frac{D}{2K}}}\exp(-\frac{\norm{\bbf^{(k)}_i-\bw^{(k)}_j}^2}{2\sigma^{(k)2}_i}).
\end{equation}
And therefore, the posterior classification probability is:
\begin{align}
%\begin{split}
%p(y_i=j|\bx_i) =& \frac{p_\theta(\bw_j|\bx_i)}{\sum_{c=1}^{C}{p_\theta(\bw_c|\bx_i)}} \\
p(y=j|\bx_i) = & \frac{p_{\theta}(\bw_j|\bx_i)}{\sum_{c=1}^{N}{p_{\theta}(\bw_c|\bx_i)}} \\
= & \frac{\exp(\mathbf{a'}_{i,j})}{\sum_{c=1}^{N}{\exp(\mathbf{a'}_{i,j})}},
\end{align}

where
\begin{equation}
\mathbf{a'}_{i,j} = -\sum_{k=1}^{K}\frac{\norm{\bbf^{(k)}_i-\bw^{(k)}_j}^2}{2\sigma^{(k)2}_i}.
\label{app:eq:logit}
\end{equation}
Given that $\norm{\bbf_i^{(k)}}^2=\norm{\bw_j^{(k)}}^2=1$ and $s_i^{(k)}=\frac{1}{\sigma^{(k)2}_i}$, Equation~(\ref{app:eq:logit}) becomes:
\begin{equation}
\mathbf{a'}_{i,j} = \sum_{k=1}^{K}{s_i^{(k)}\bbf^{(k)T}_i\bw^{(k)}_j}-\sum_{k=1}^{K}{s_i^{(k)}}.
\label{app:eq:logit_new}
\end{equation}
The second term is cancelled out when computing the probability. By further incorporating the margin $m$ in to Equation~(\ref{app:eq:logit_new}) and taking the average score instead of the sum, one could derive the Equation~(8) in the main paper.

\subsection{Gradient of the sub-embeddings}
Here, we try to understand how the confidence helps the training by looking at the gradient of the sub-embeddings in Equation~(8) in the main paper.
Notice that we have 
\begin{align}
\frac{\partial\mathcal{L}_{idt}}{\mathbf{a}_{i,j}} =  p_{i,j} - \delta_{y_i,j},
\label{app:eq:gradient_loss}
\end{align}
where $\delta_{y_i,j}$ is $1$ if $y_i=j$ and $0$ otherwise. $p_{i,j}=p(y=j|\bx_i)$ is the posterior classification probability. Since
$\frac{\partial\mathbf{a}_{i,j}}{\partial\bw_j^{(k)}} = {s_i^{(k)}\bbf_i^{(k)}}$ and $\frac{\partial\mathbf{a}_{i,j}}{\partial\bbf_j^{(k)}} = {s_i^{(k)}\bw_j^{(k)}}$, we have
\begin{align}
\begin{split}
\frac{\partial\mathcal{L}_{idt}}{\partial\mathbf{\bw}_{j}^{(k)}} = & s_i^{(k)}(p_{i,j} - \delta_{y_i,j})\bbf_i^{(k)}\\
\frac{\partial\mathcal{L}_{idt}}{\partial\mathbf{\bbf}_{j}^{(k)}} = & s_i^{(k)}((p_{i,y_i} - 1)\bw_{y_i}^{(k)} + \sum_{j\neq y_i}{p_{i,j}\bw_{j}^{(k)}})
\end{split}
\label{app:eq:gradient_weights}
\end{align}
From Equation~(\ref{app:eq:gradient_weights}), it can be seen that the gradient of the prototypes and sub-embeddings depend on both the confidence value and the classification probability. In particular, confidence value $s_i^{(k)}$ serves as a gating parameter during the back-propagation. In such a way, the prototypes would be affected more by the confident samples than the not confident ones. Similarly, the confident sub-embedding would also have a larger impact on the prototype.

\section{Additional Implementation Details}
The backbone of our embedding network $\theta$ is a modified 100-layer ResNet in~\cite{deng2018arcface}. The network is split into two different branches after the last convolution layer, each of which includes one fully connected layer. The first branch outputs a $512$-D vector, which is further split into $16$ sub-embeddings. The other branch outputs a $16$-D vector, which are confidence values for the sub-embeddings. The $\exp$ function is used to guarantee all the confidence values $s_i^{(k)}$ are positive. The model $\theta_A$ that we used for mining additional variations is a four layer CNN. The four layers have $64$, $128$, $256$ and $512$ kernels, respectively, all of which are $3\times3$.

\section{Ablation Study on Variation Decorrelation Loss}
In Table~\ref{tab:ablation_variation} we show the results of training with different number of variations for the variation decorrelation loss. The base model in the first line is a model trained with all the modules proposed in the paper except variation decorrelation loss. The second to fourth line show the results of using different number augmentable variations (blur, occlusion and pose) and additional variations (gender, age and smiling). It can be seen that with more variation added into the training, the decorrelation becomes more effective and leads to a better performance.

\section{Additional Results on IJB-S}
Table~\ref{tab:ijbs} shows more results of our models as well as state-of-the-art methods on the IJB-S dataset. The ``Surveillance-to-Single'' protocol uses one single image in the gallery templates while the ``surveillance-to-booking'' use a set of face images with different poses in the gallery templates. Compared with our own baseline, significant performance boost can be observed on all the metrics, which proves the efficacy of the proposed method. Compared with the state-of-the-art methods, our final model achieves better performance on most of the metrics.

\section{Visualization of Sub-embedding Confidence}
\begin{figure*} [t]
    \centering
    \includegraphics[width=0.24\linewidth]{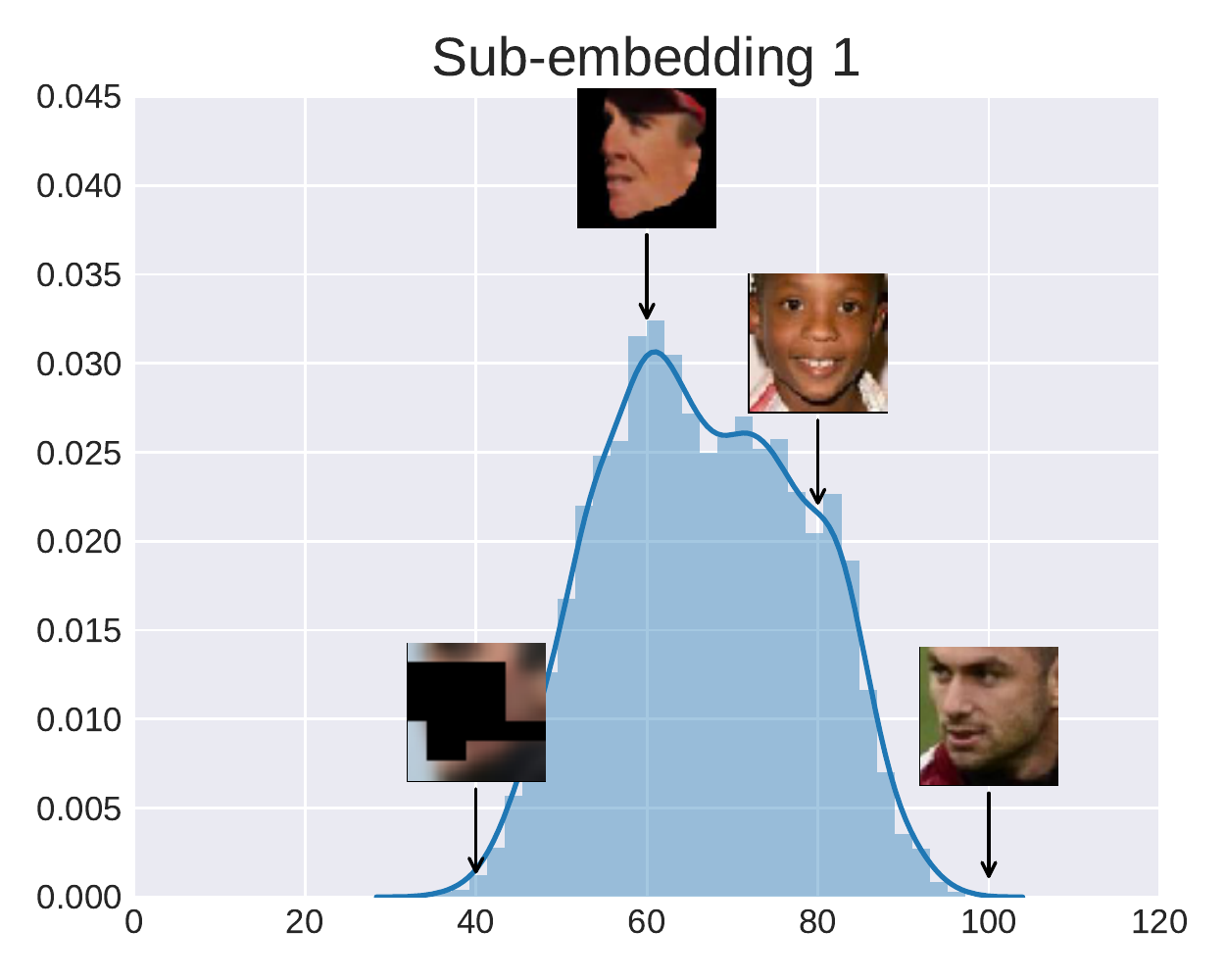}
    \includegraphics[width=0.24\linewidth]{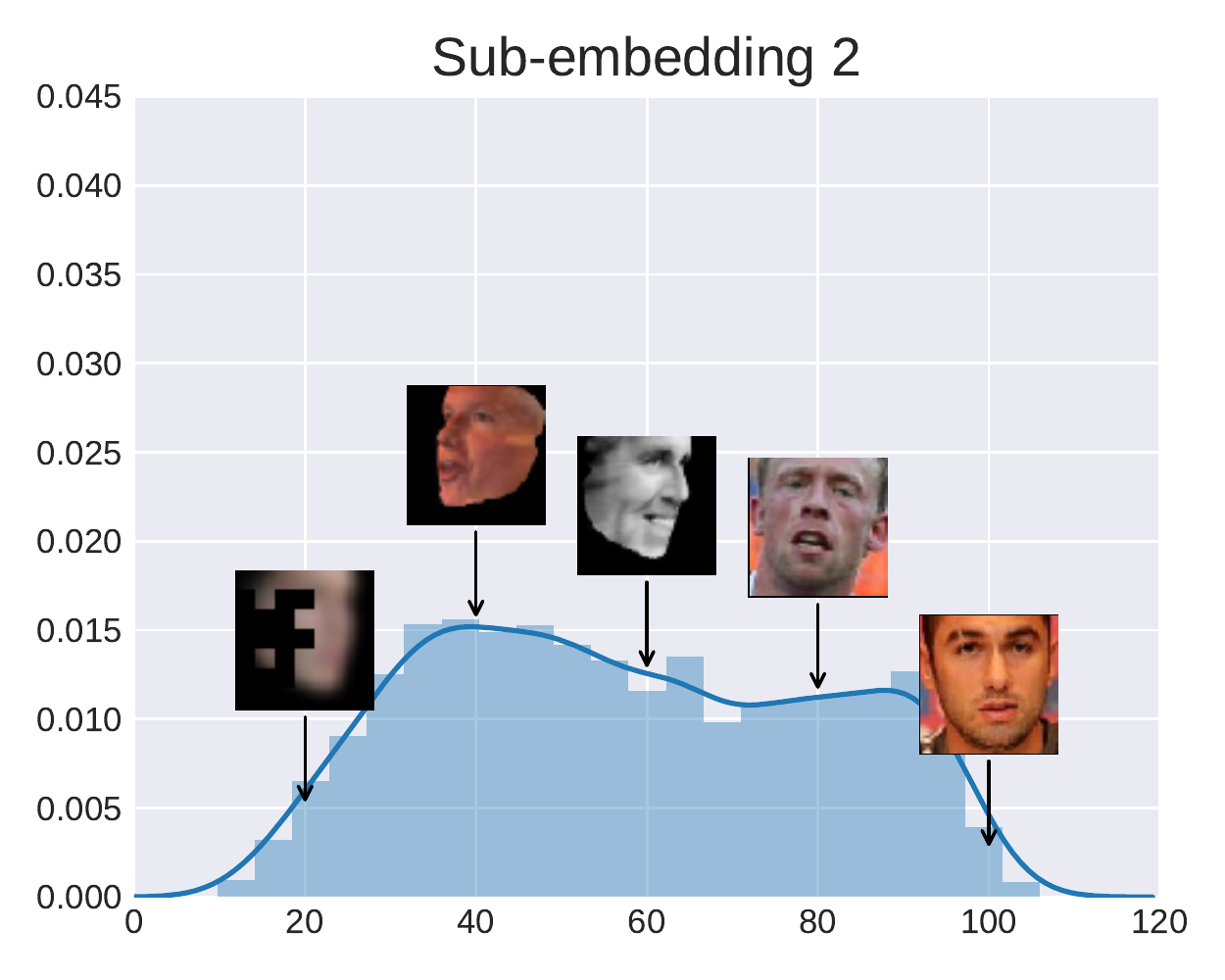}
    \includegraphics[width=0.24\linewidth]{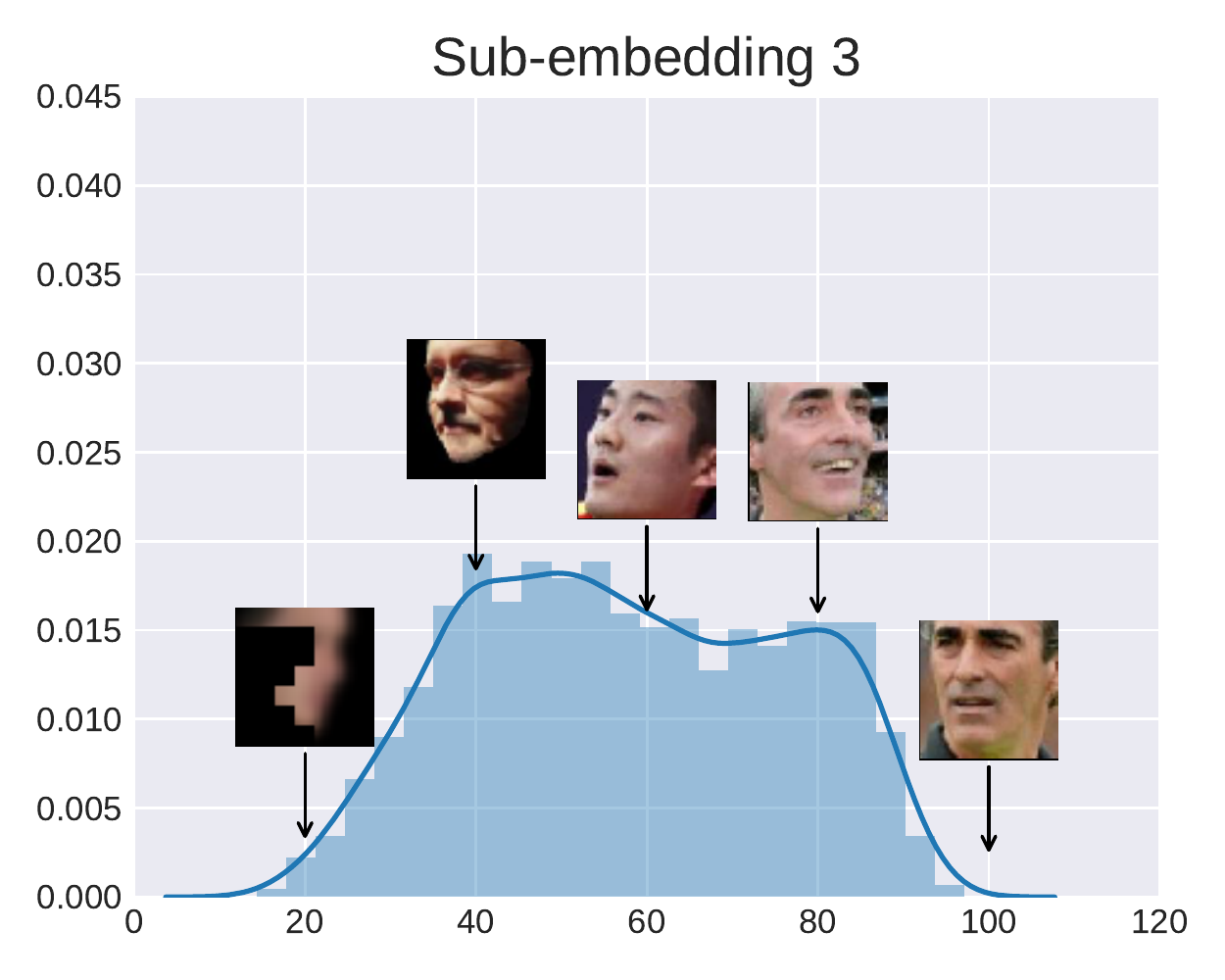}
    \includegraphics[width=0.24\linewidth]{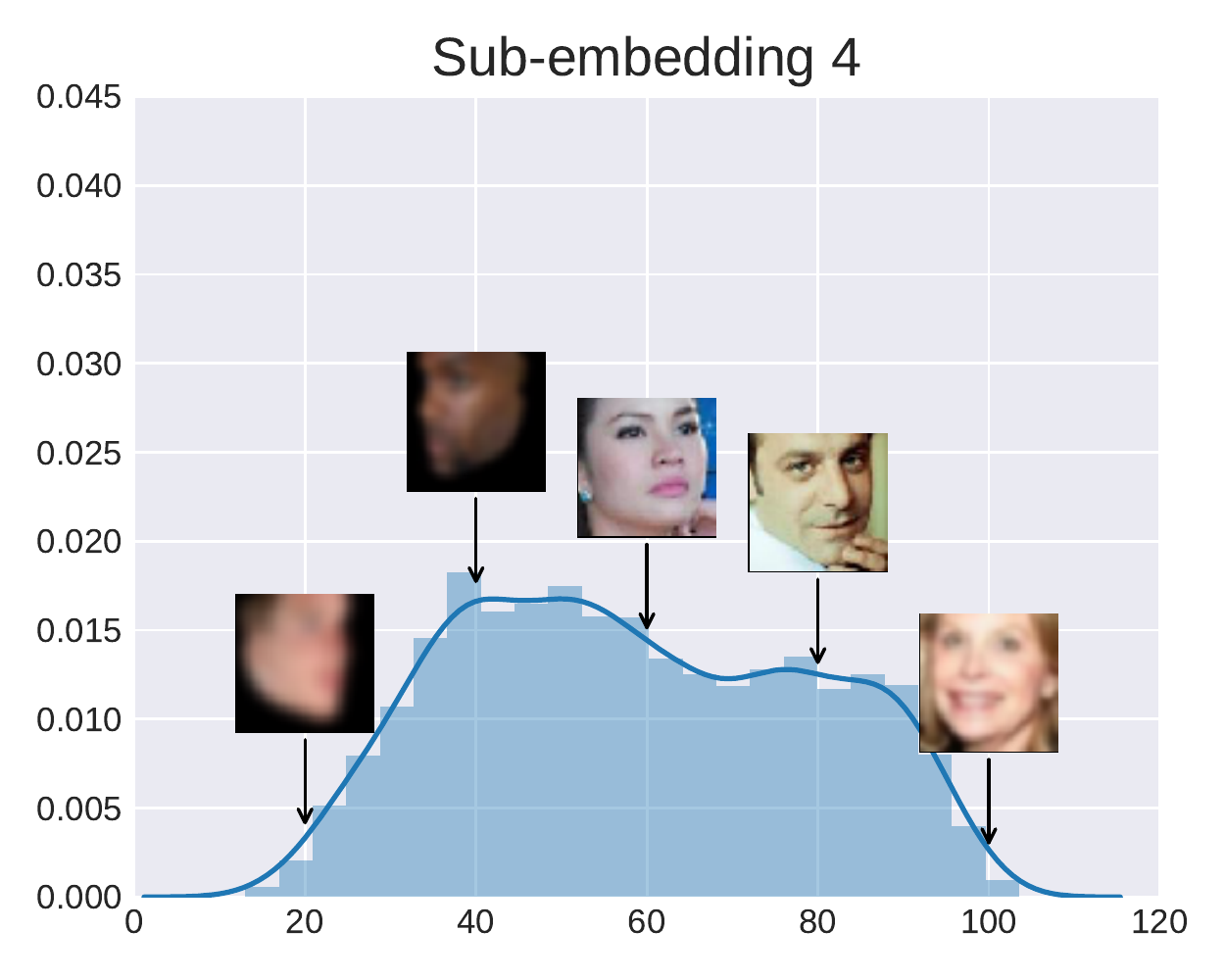}\\[1.0em]
    \includegraphics[width=0.24\linewidth]{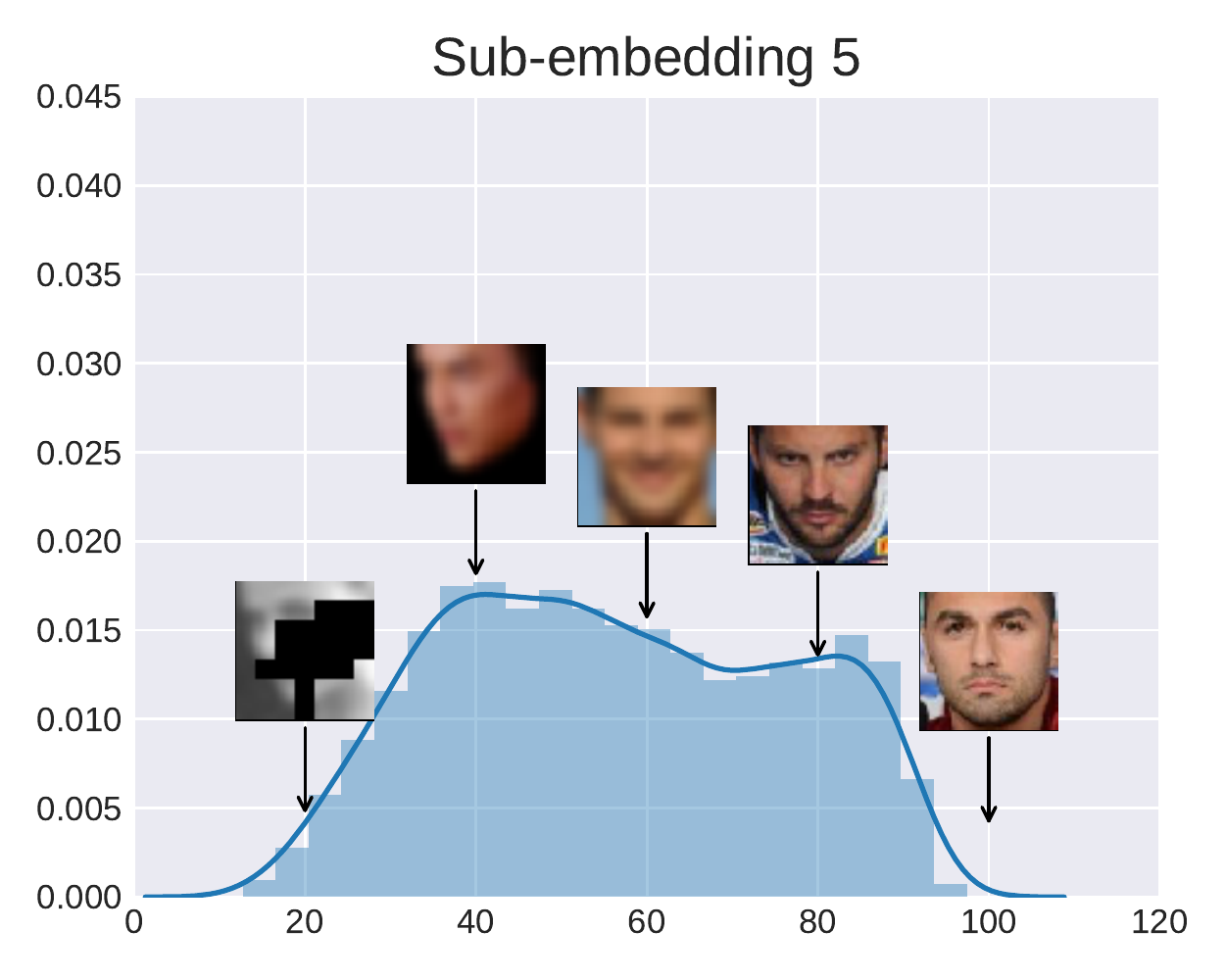}
    \includegraphics[width=0.24\linewidth]{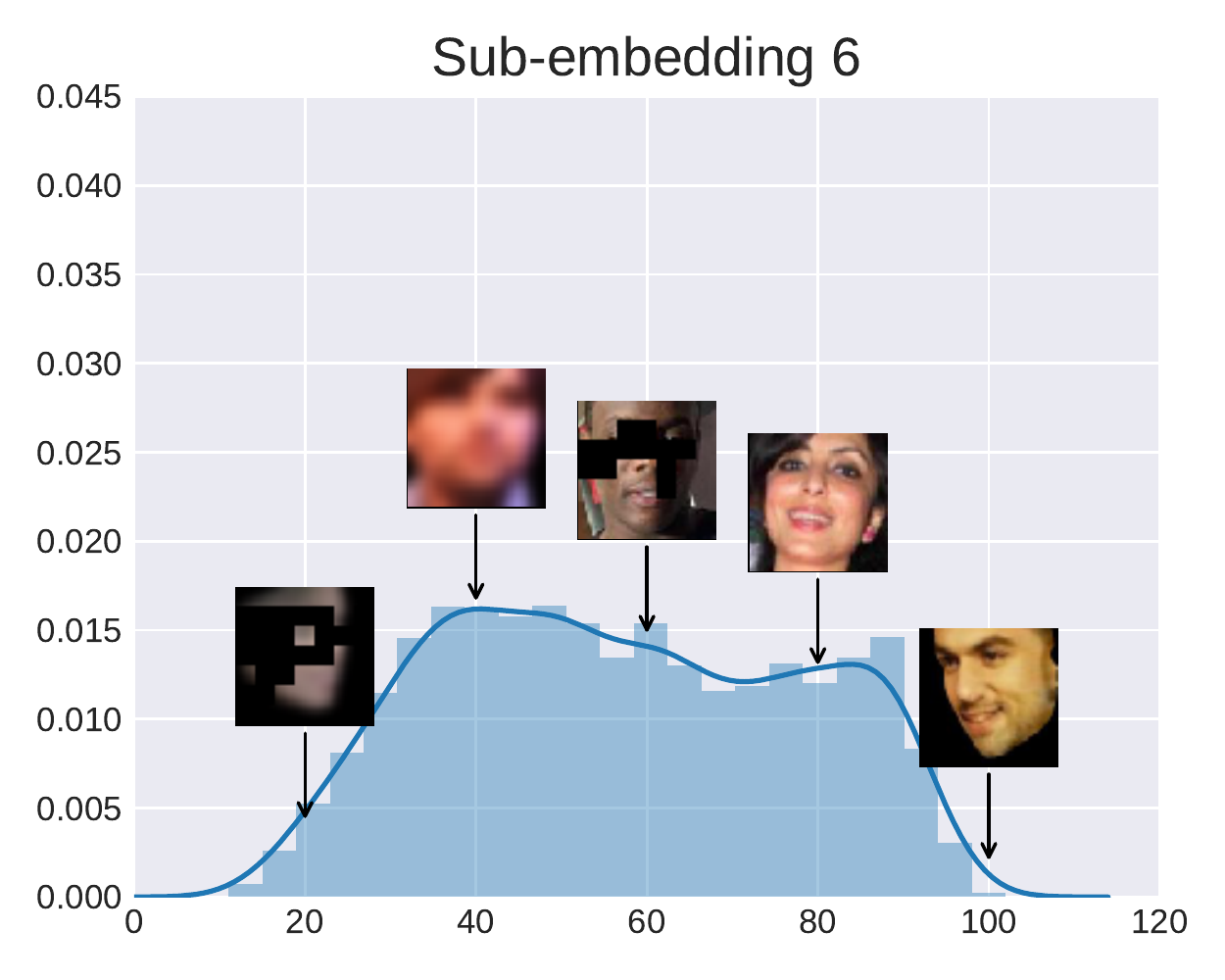}
    \includegraphics[width=0.24\linewidth]{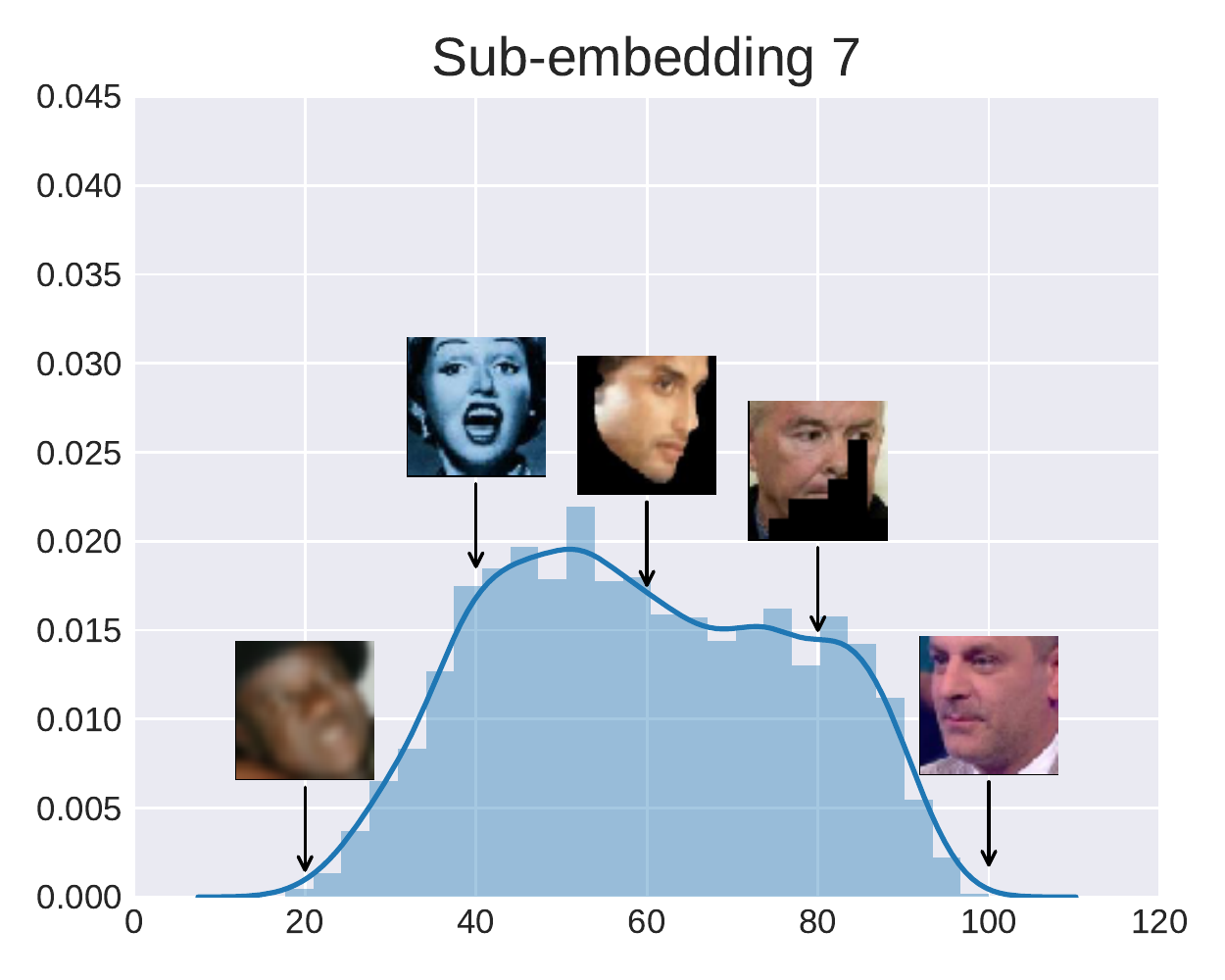}
    \includegraphics[width=0.24\linewidth]{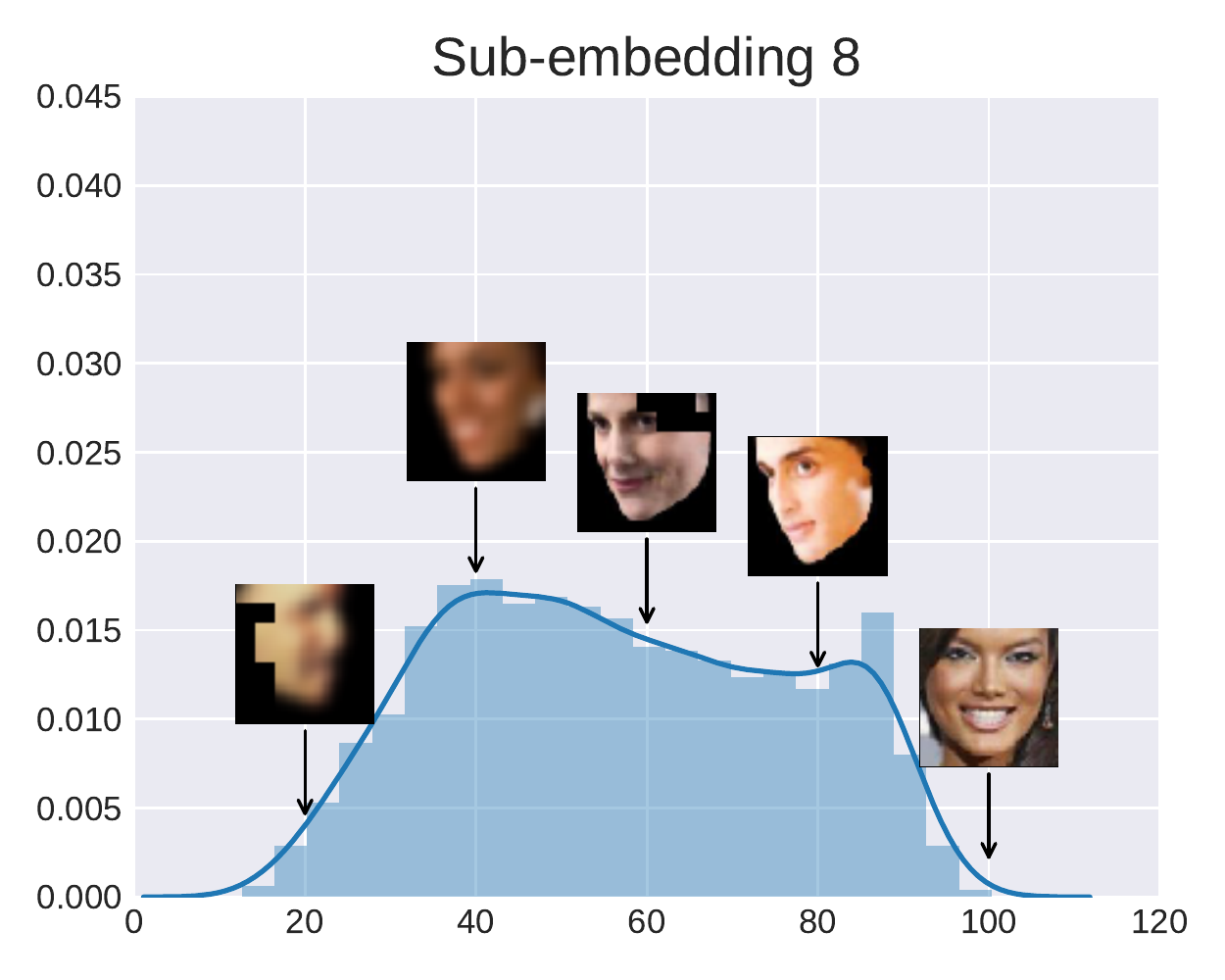}\\[1.0em]
    \includegraphics[width=0.24\linewidth]{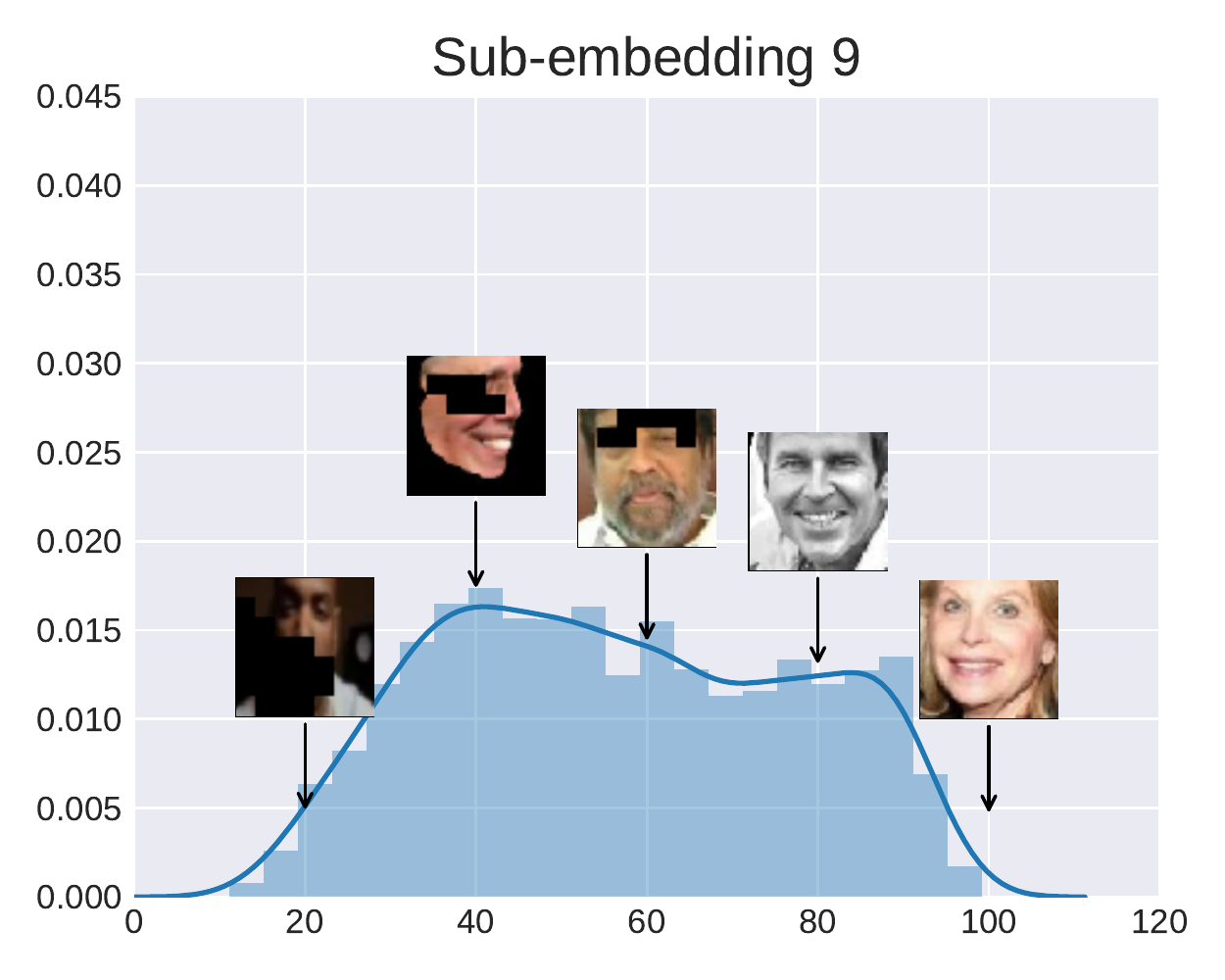}
    \includegraphics[width=0.24\linewidth]{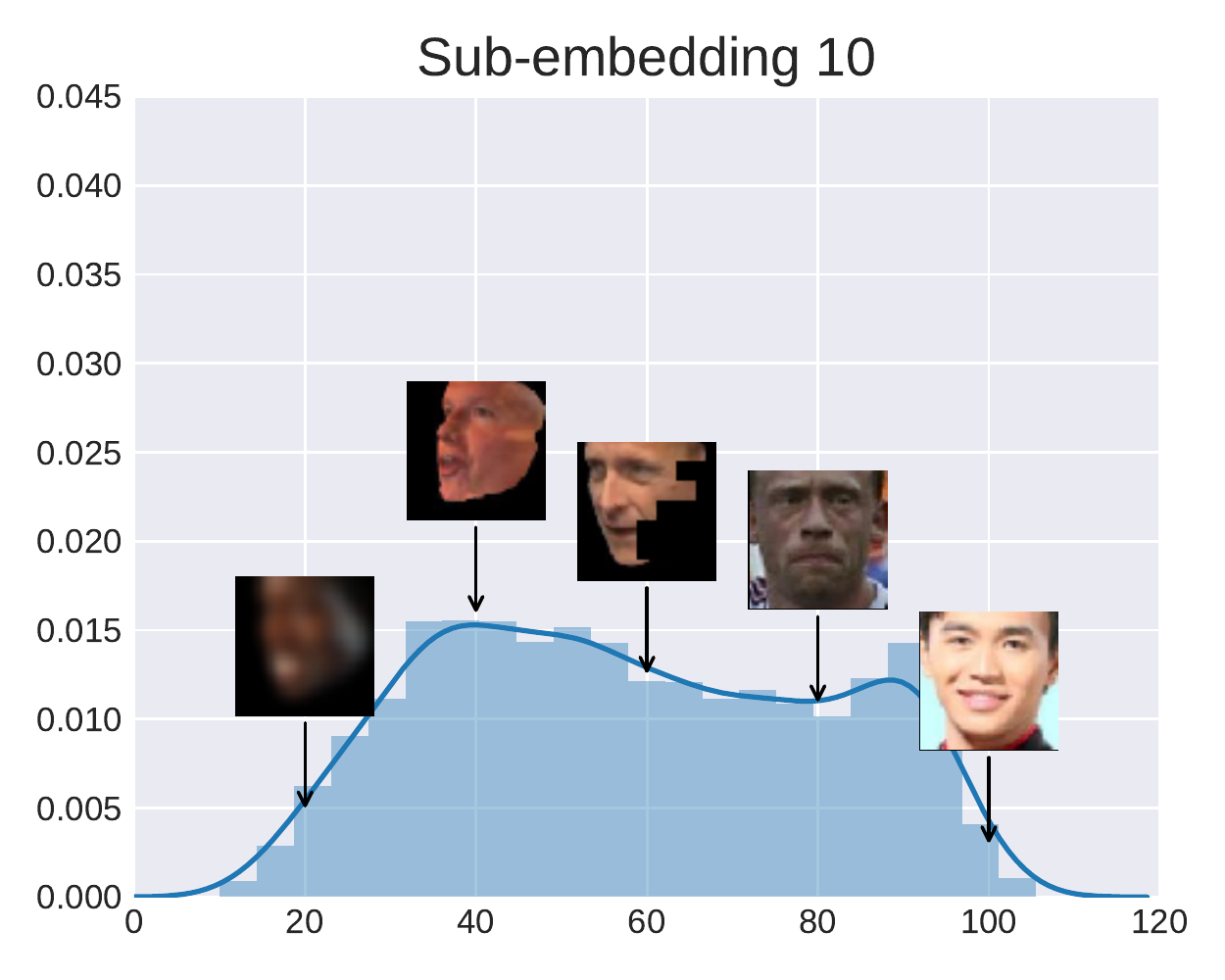}
    \includegraphics[width=0.24\linewidth]{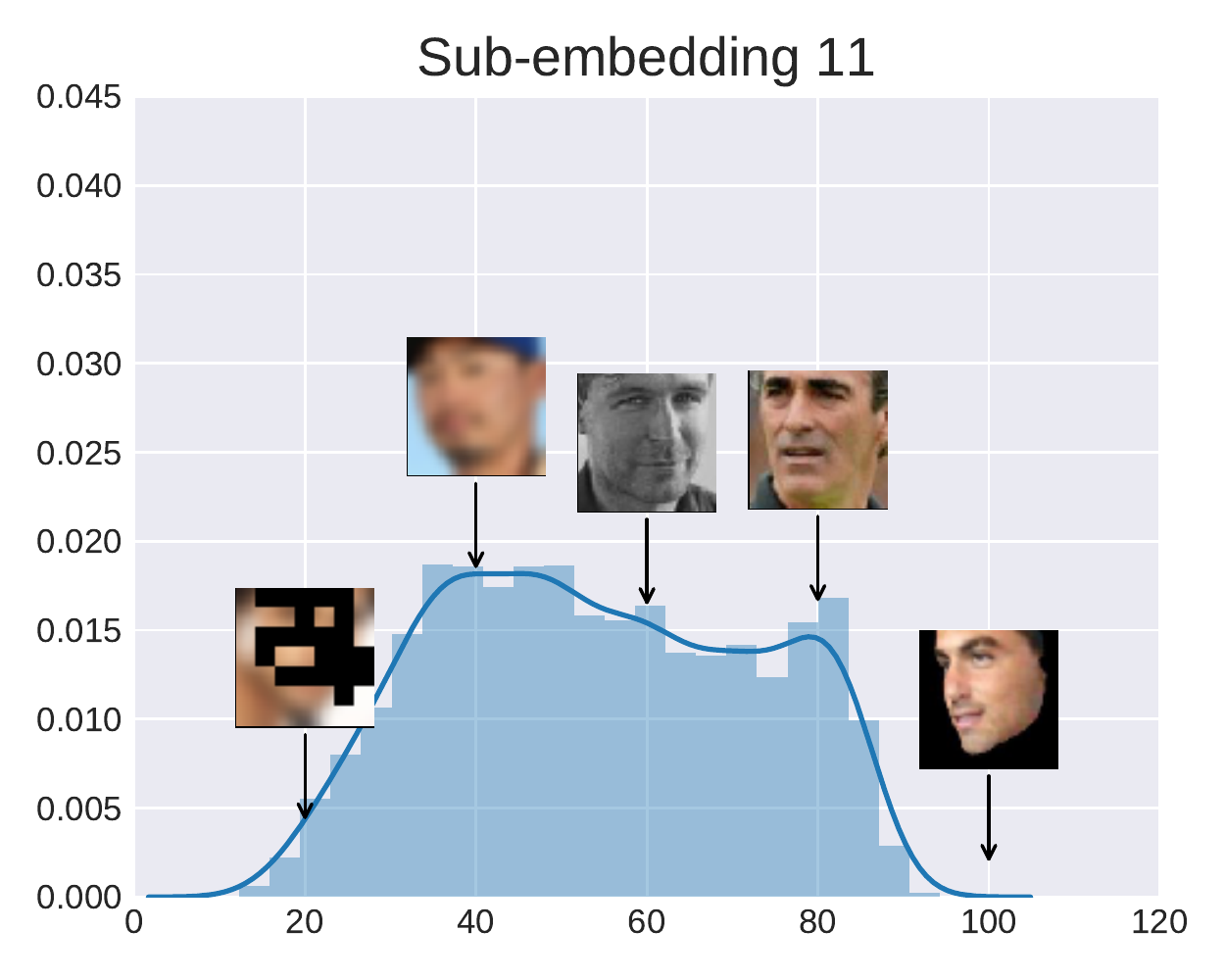}
    \includegraphics[width=0.24\linewidth]{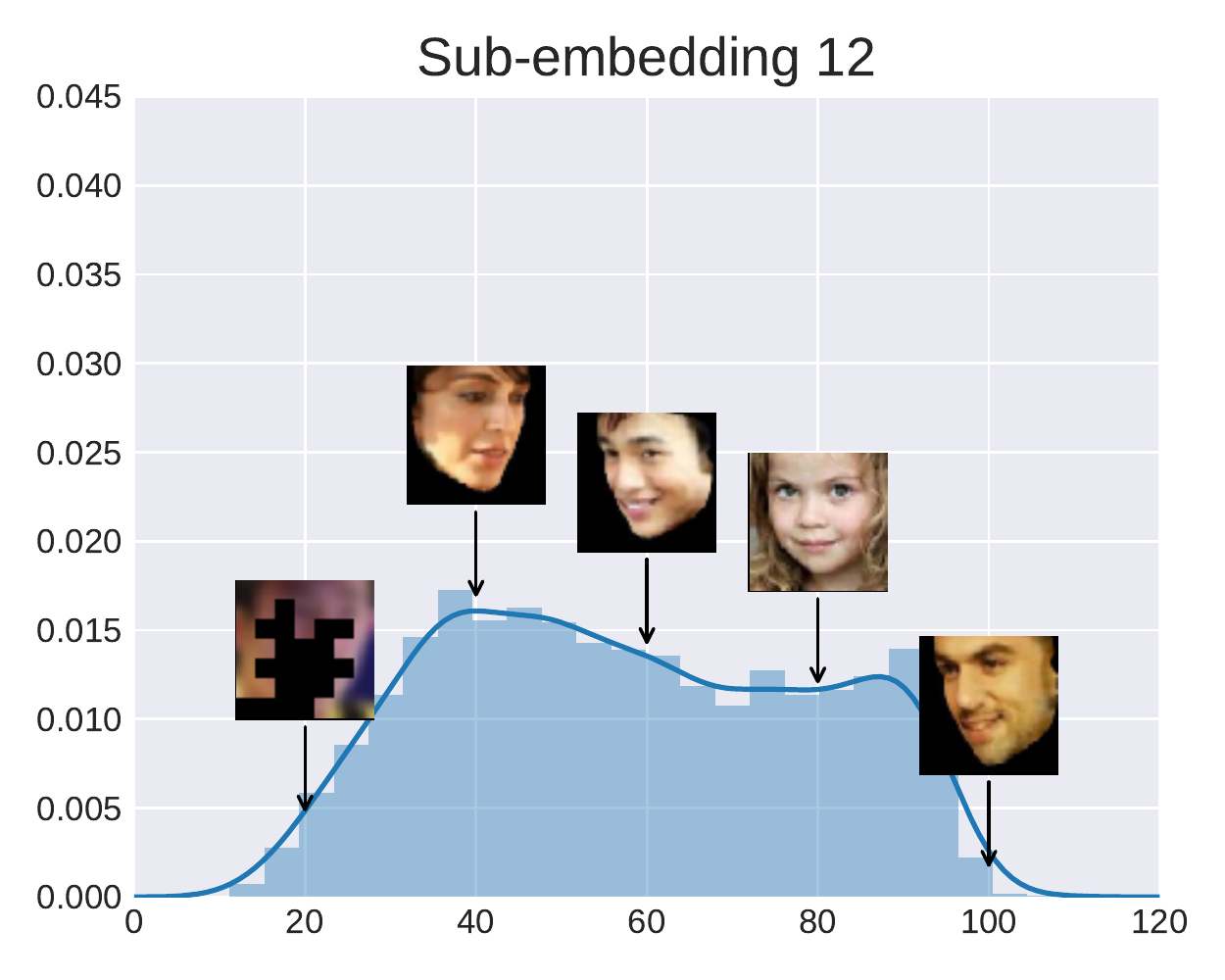}\\[1.0em]
    \includegraphics[width=0.24\linewidth]{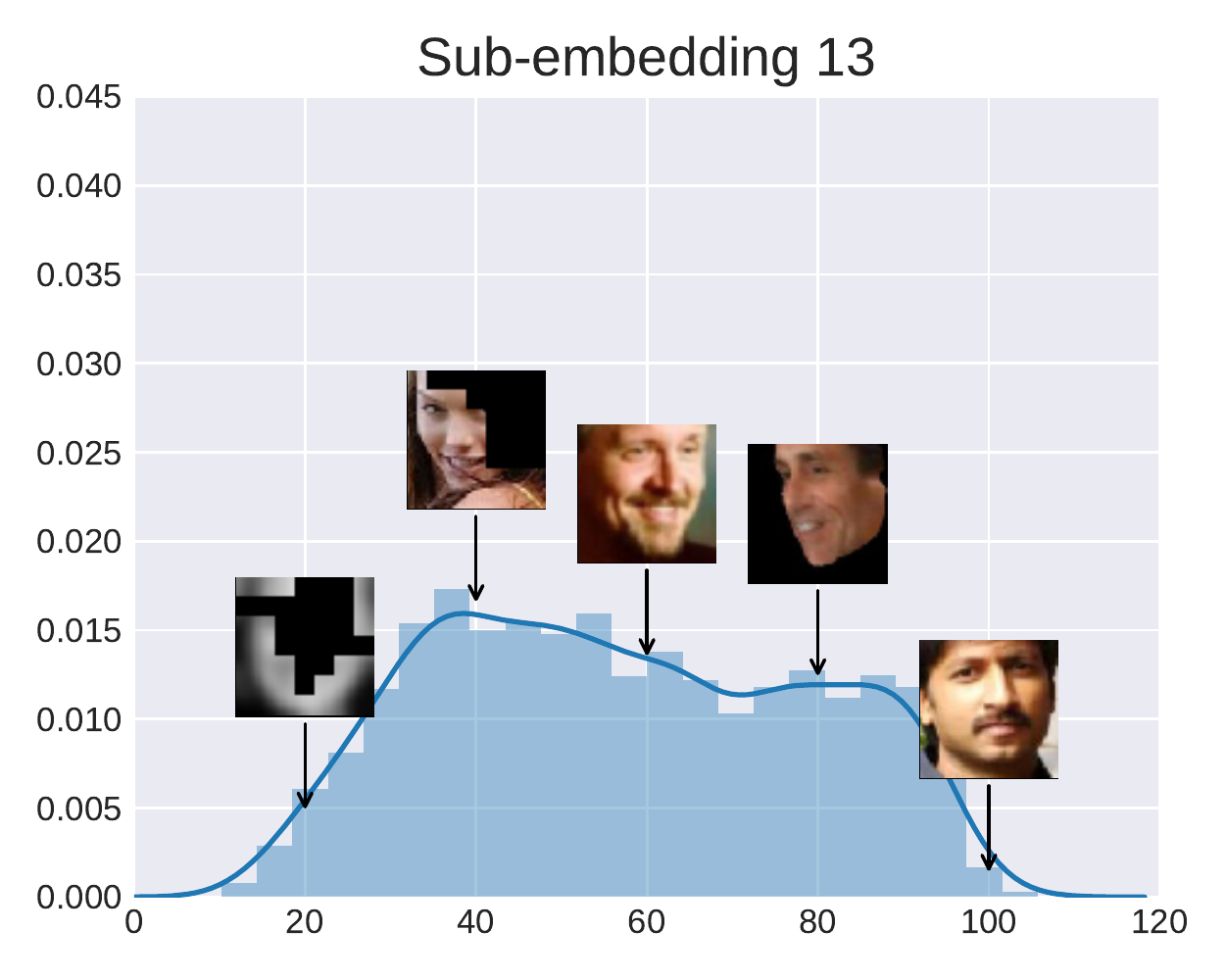}
    \includegraphics[width=0.24\linewidth]{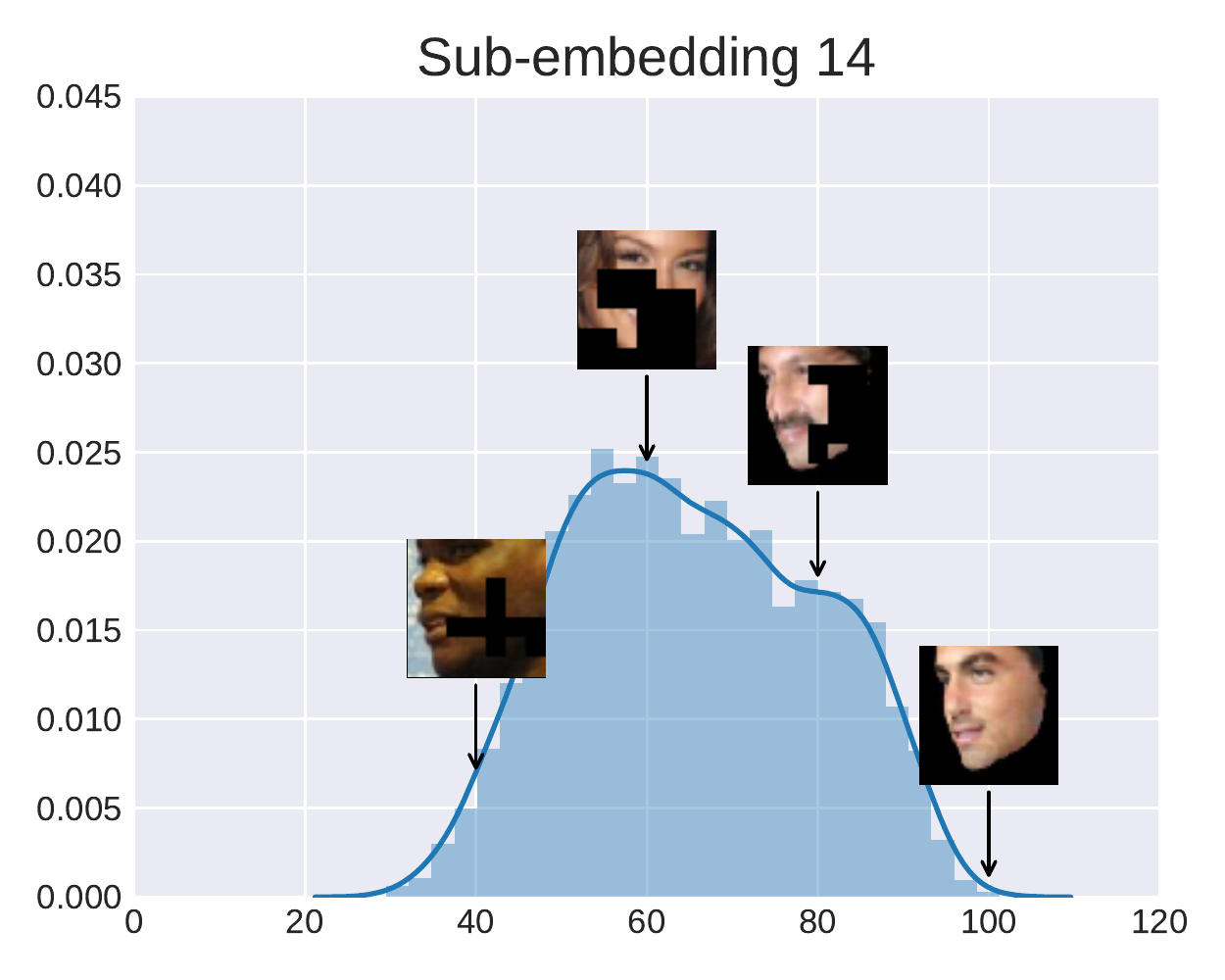}
    \includegraphics[width=0.24\linewidth]{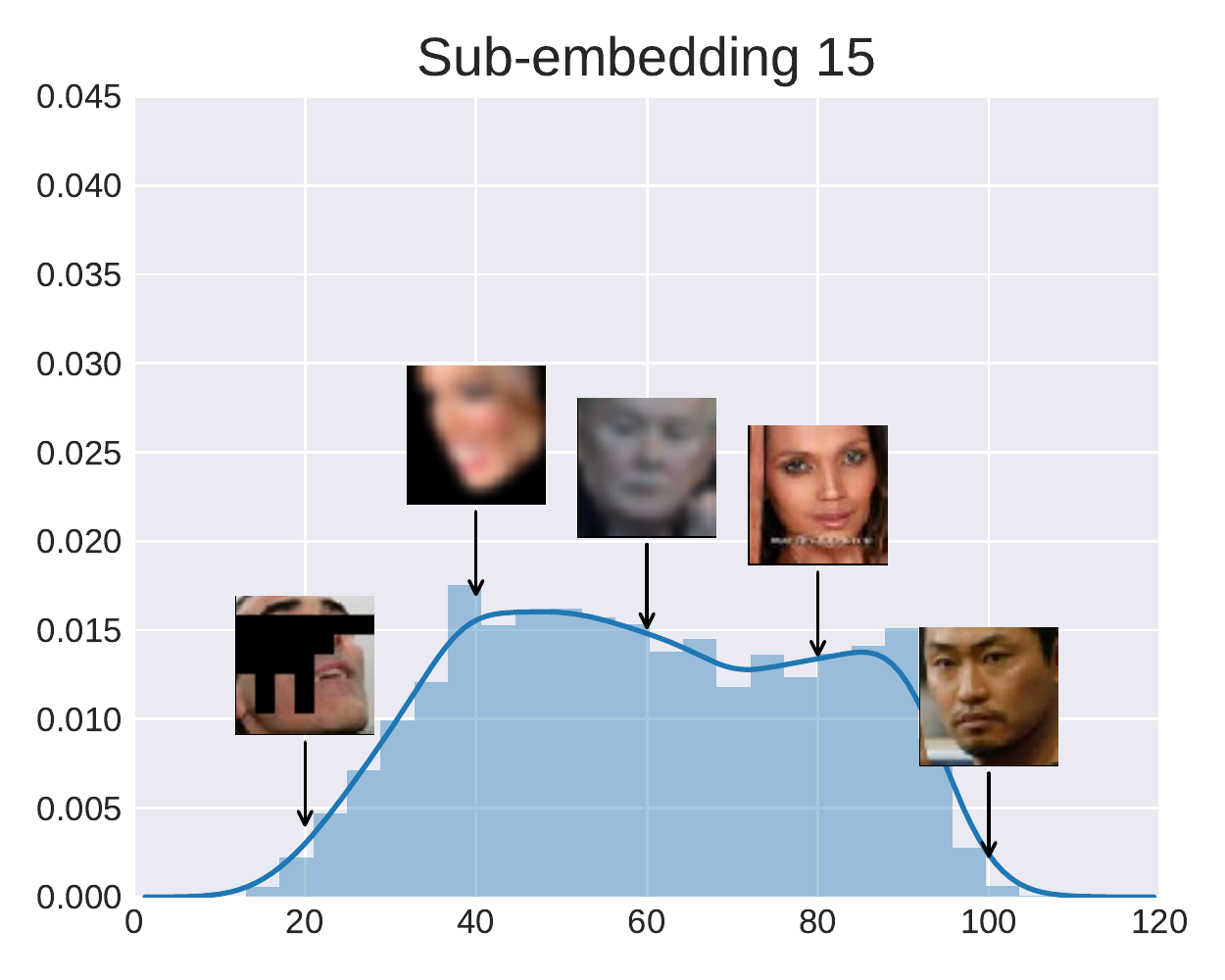}
    \includegraphics[width=0.24\linewidth]{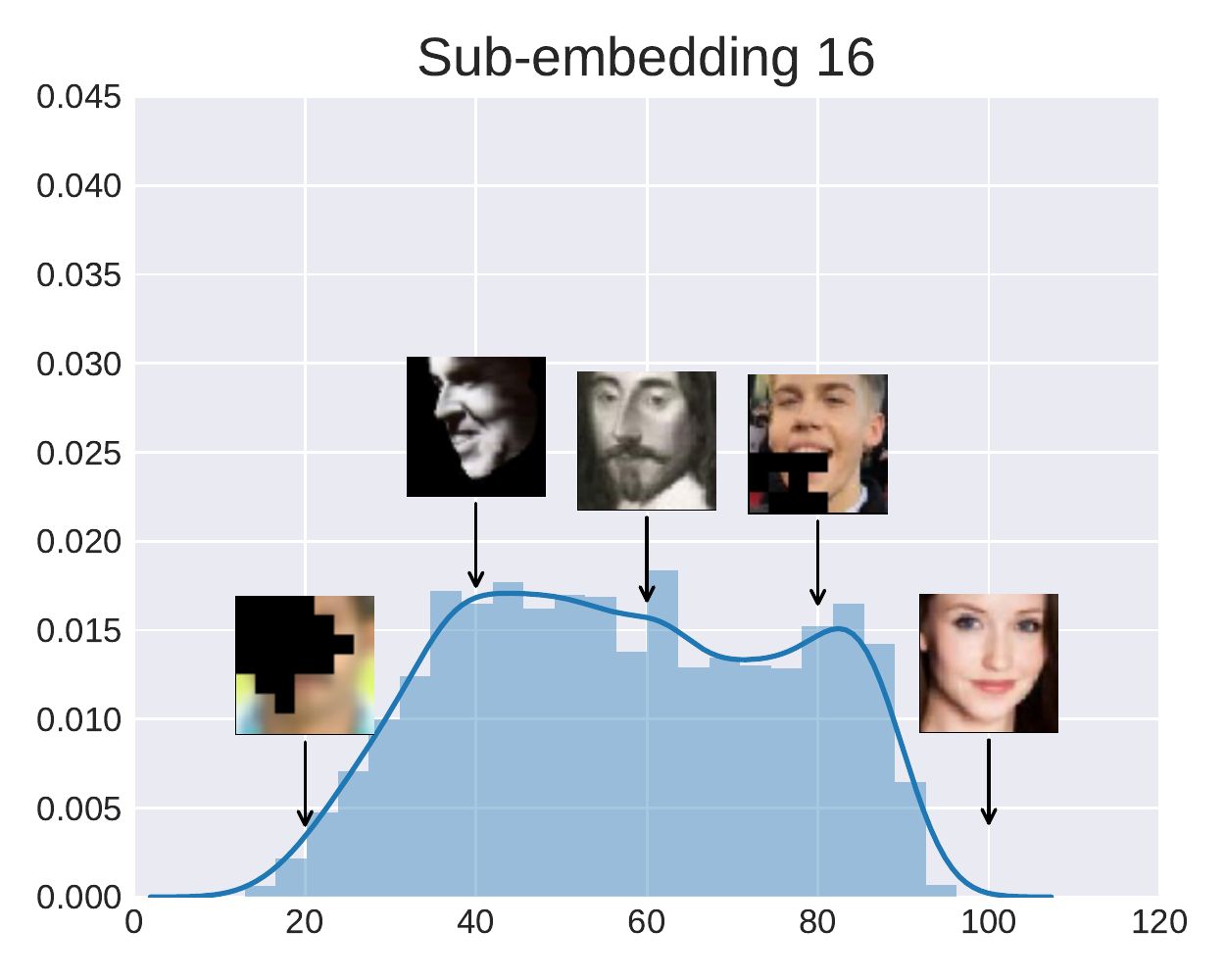}
    \caption{Visualization of sub-embedding confidence on training samples.}
    \label{fig:dist_confidence}
\end{figure*}

Figure~\ref{fig:dist_confidence} shows the distributions of confidence values during training. It can be observed that the confidences of different sub-embeddings not only have different distributions, but also vary in terms of which kind of images have high/low confidence. Since the confidence guides the training signal of the corresponding features, this reflects that the sub-embeddings learn different features complementary to each other for better identification performance.

\section{Visualization of Uncertainty}
\begin{figure*} [t]
    \centering
    \includegraphics[width=0.98\linewidth]{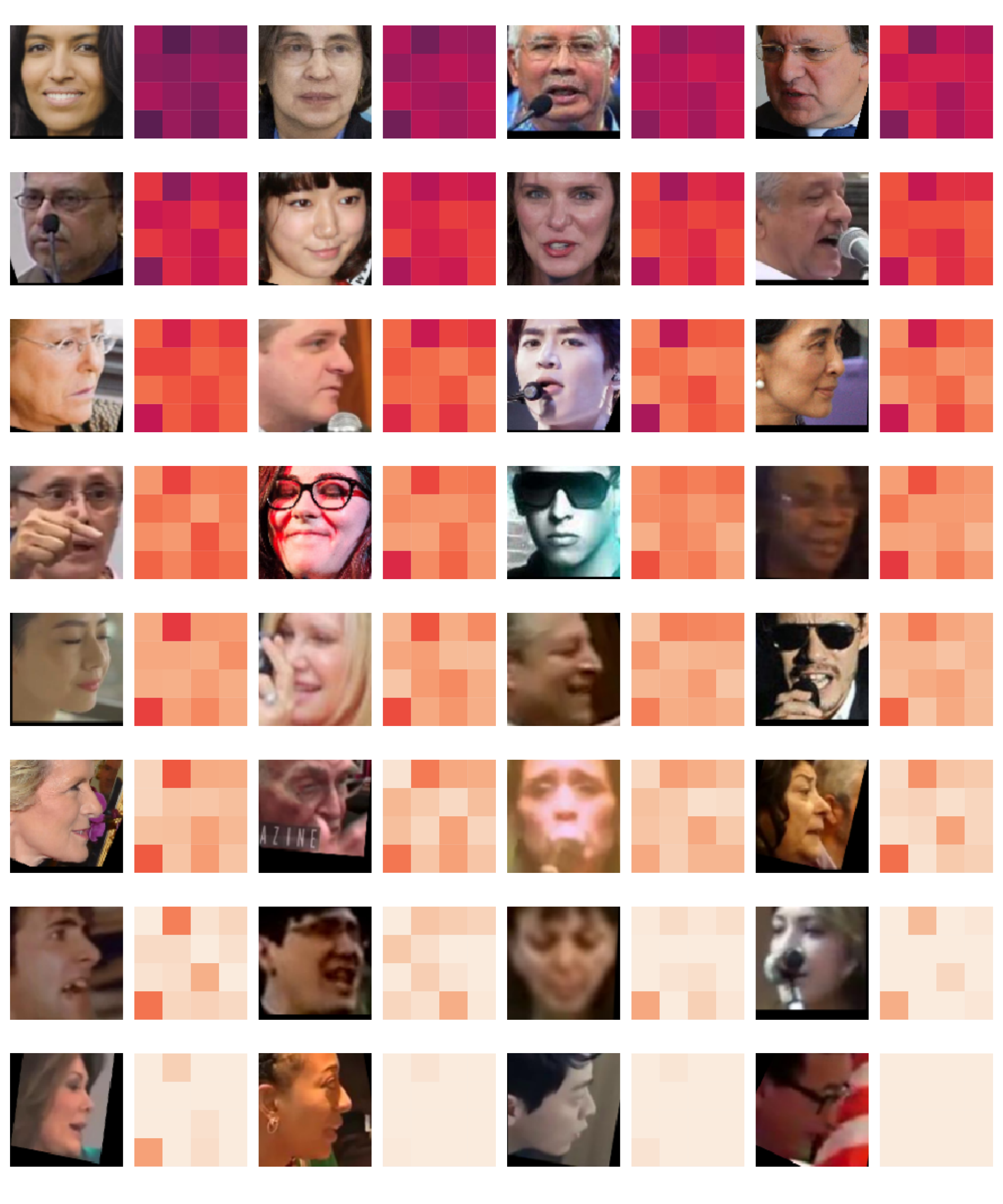}\\[-0.7em]
    \includegraphics[width=1.0\linewidth]{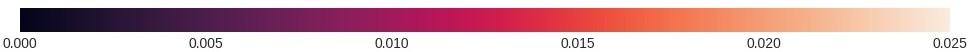}\\
    \caption{Visualization of sub-embedding uncertainty on testing images.}
    \label{fig:uncertainty_new}
\end{figure*}

\begin{figure*} [t]
    \centering
    \includegraphics[width=0.98\linewidth]{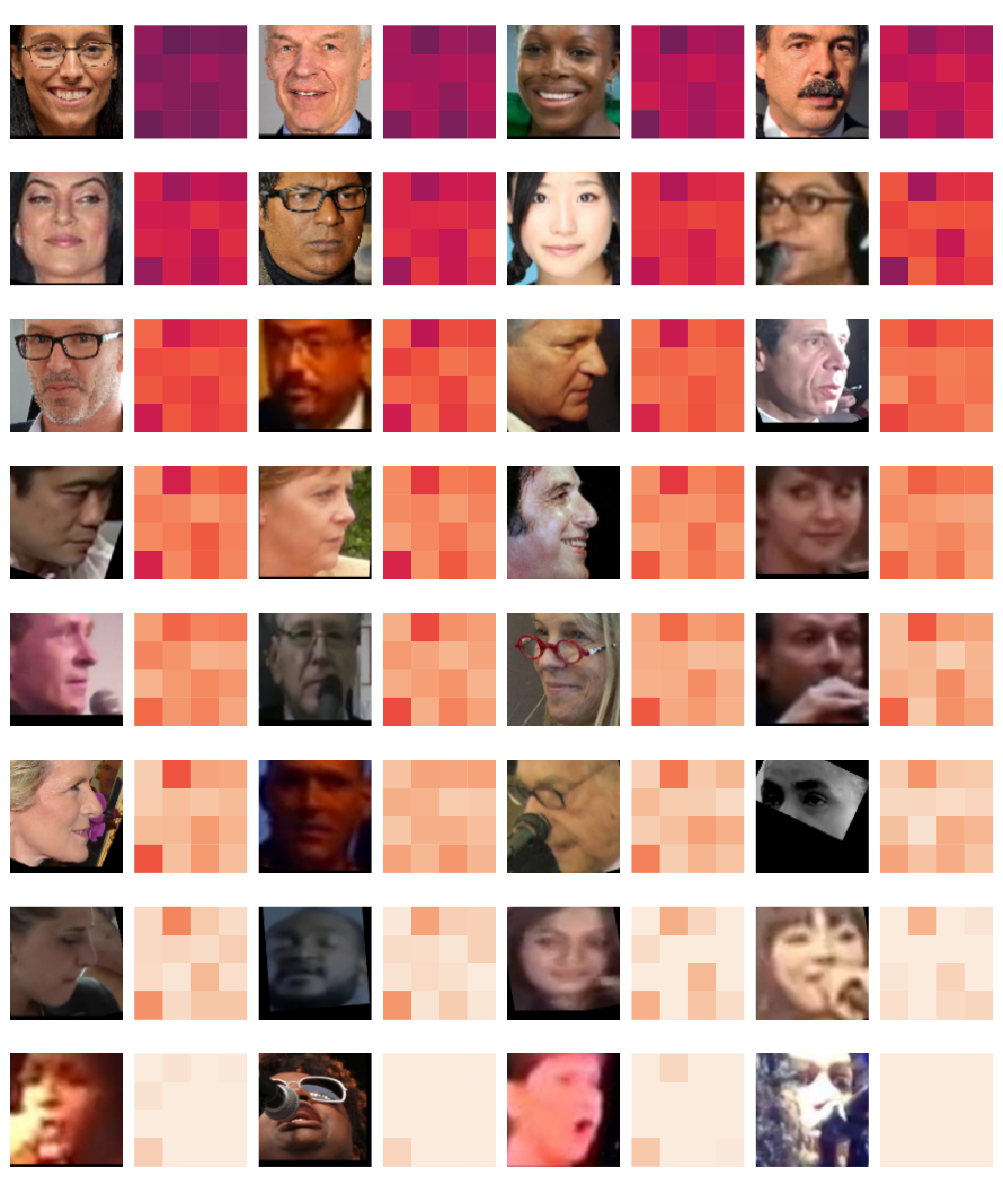}\\[-0.7em]
    \includegraphics[width=1.0\linewidth]{fig/seaborn_colorbar_new.jpg}\\
    \caption{Visualization of sub-embedding uncertainty on more testing images.}
    \label{fig:uncertainty_new_2}
\end{figure*}

In Figure~\ref{fig:uncertainty_new} and Figure~\ref{fig:uncertainty_new_2}, we show more results of uncertainty heatmaps. Overall, we can see that distinguishable face images have low uncertainty on most sub-embeddings. Faces with larger variations have some sub-embeddings with low uncertainty, depending on which kind of variation is present. For images with extremely large variations, high uncertainty is observed on all the sub-embeddings.

\section{Visualization of Face Representations}
\begin{figure*}[t]
    \captionsetup{font=small}
    \centering
    \subfloat[Baseline]{\includegraphics[width=0.49\linewidth]{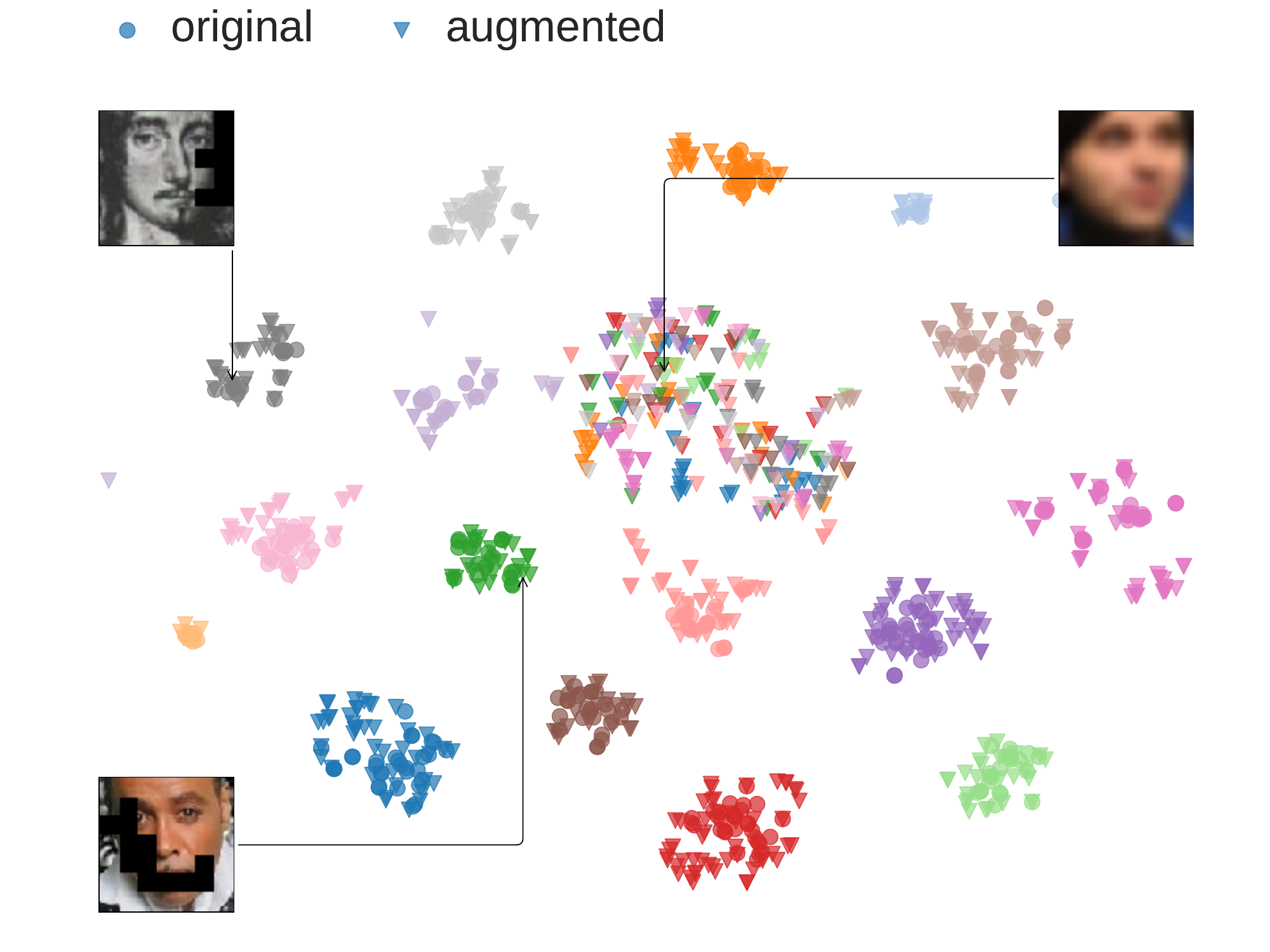}}\hfill
    \subfloat[Proposed]{\includegraphics[width=0.49\linewidth]{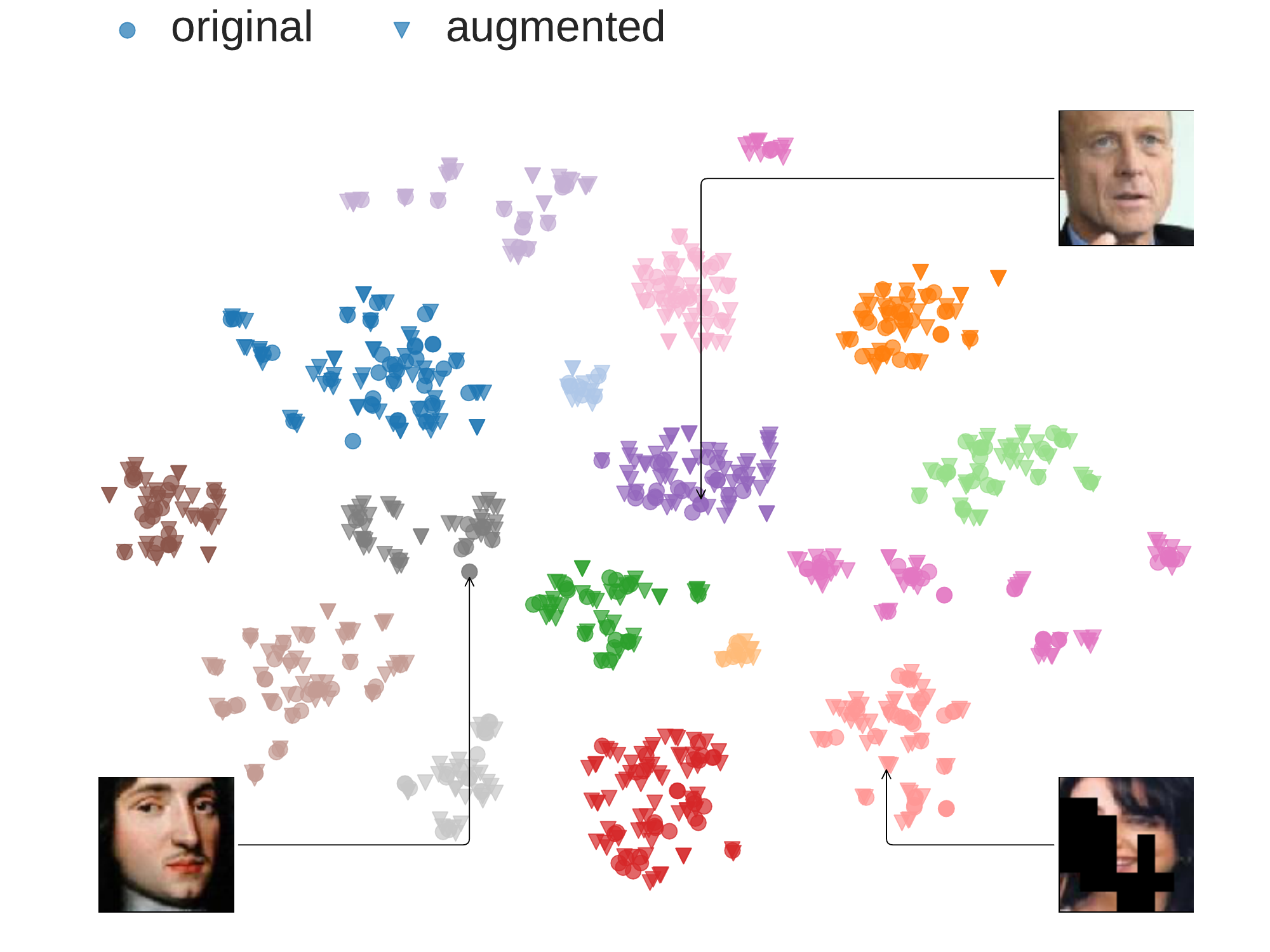}}
    \caption{t-SNE visualization of the features in a 2D space. Colors indicate the identities. Original training samples and augmented training samples are shown in circle and triangle, respectively.}
    \label{fig:tsne_mixed_appendix}
\end{figure*}

In Figure~\ref{fig:tsne_mixed_appendix}, we show the t-SNE visualization of the embeddings from the baseline (with augmentation) method as well as the proposed method. The original training samples and the augmented ones are shown in circle and triangle, respectively. Notice that some augmented samples are hard to recognize and are close to be noises. Thus, by assuming an equal confidence for all the samples, the baseline method fails to converge to a good local minimum and many augmented samples cluster together in a small area. In comparison, by focusing more on the high-quality samples, the proposed method learns a more discriminative feature space. Although noisy outliers still exist in the proposed method, they are usually close to their own identities' samples.

\section{Image Examples From the Testing Datasets}
\begin{figure*}
    \centering
    \footnotesize
    \captionsetup{font=small}
    \begin{minipage}{0.48\linewidth}
\includegraphics[width=0.166\linewidth]{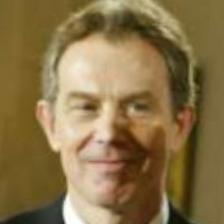}\hfill
\includegraphics[width=0.166\linewidth]{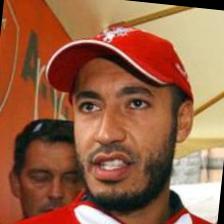}\hfill
\includegraphics[width=0.166\linewidth]{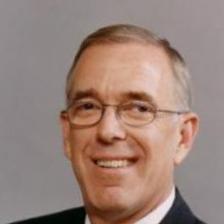}\hfill
\includegraphics[width=0.166\linewidth]{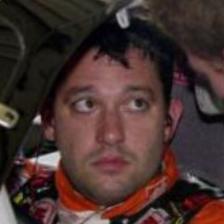}\hfill
\includegraphics[width=0.166\linewidth]{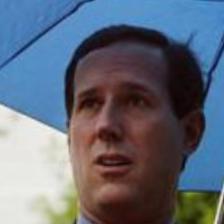}\hfill
\includegraphics[width=0.166\linewidth]{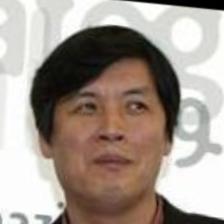}\\[-0.1em]
\includegraphics[width=0.166\linewidth]{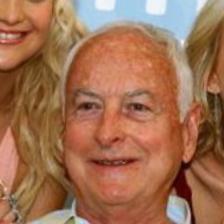}\hfill
\includegraphics[width=0.166\linewidth]{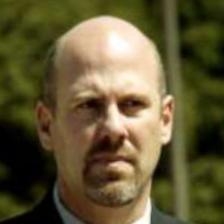}\hfill
\includegraphics[width=0.166\linewidth]{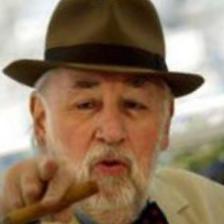}\hfill
\includegraphics[width=0.166\linewidth]{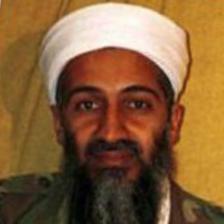}\hfill
\includegraphics[width=0.166\linewidth]{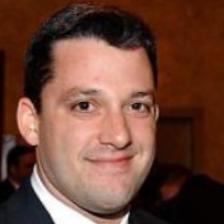}\hfill
\includegraphics[width=0.166\linewidth]{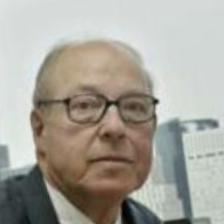}\\[-0.1em]
\includegraphics[width=0.166\linewidth]{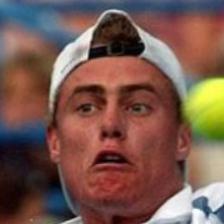}\hfill
\includegraphics[width=0.166\linewidth]{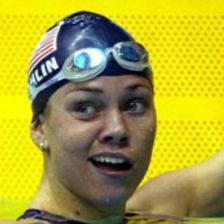}\hfill
\includegraphics[width=0.166\linewidth]{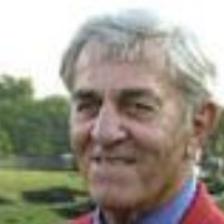}\hfill
\includegraphics[width=0.166\linewidth]{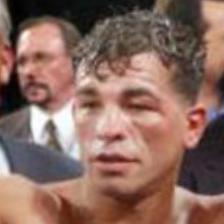}\hfill
\includegraphics[width=0.166\linewidth]{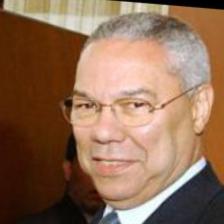}\hfill
\includegraphics[width=0.166\linewidth]{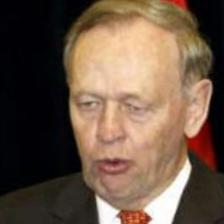}\\[-0.1em]
\includegraphics[width=0.166\linewidth]{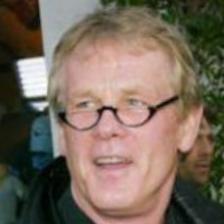}\hfill
\includegraphics[width=0.166\linewidth]{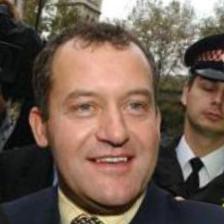}\hfill
\includegraphics[width=0.166\linewidth]{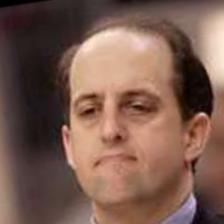}\hfill
\includegraphics[width=0.166\linewidth]{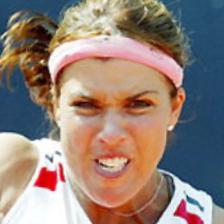}\hfill
\includegraphics[width=0.166\linewidth]{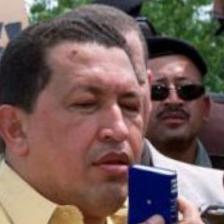}\hfill
\includegraphics[width=0.166\linewidth]{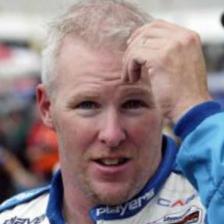}\\
    \vspace{-2.0em}\begin{center}(a) LFW\end{center}
    \end{minipage}\hfill
    \begin{minipage}{0.48\linewidth}
    \includegraphics[width=0.166\linewidth]{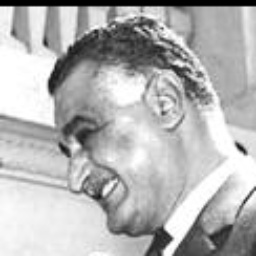}\hfill
    \includegraphics[width=0.166\linewidth]{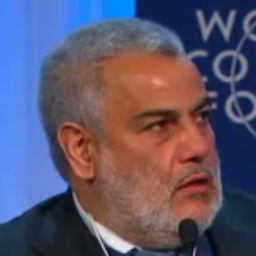}\hfill
    \includegraphics[width=0.166\linewidth]{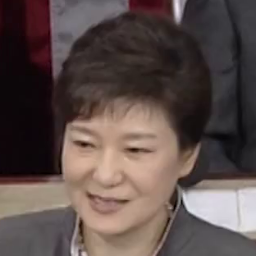}\hfill
    \includegraphics[width=0.166\linewidth]{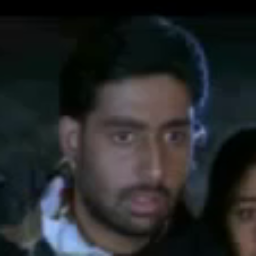}\hfill
    \includegraphics[width=0.166\linewidth]{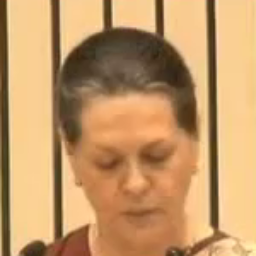}\hfill
    \includegraphics[width=0.166\linewidth]{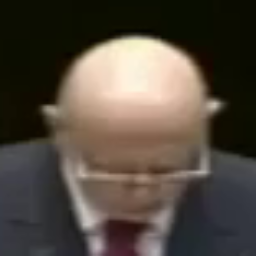}\\[-0.1em]
    \includegraphics[width=0.166\linewidth]{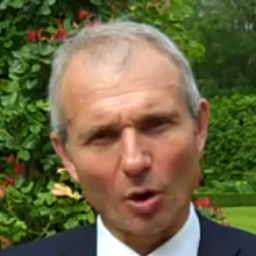}\hfill
    \includegraphics[width=0.166\linewidth]{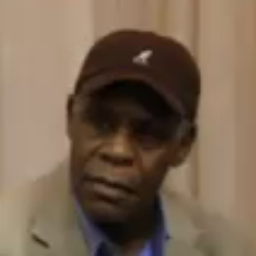}\hfill
    \includegraphics[width=0.166\linewidth]{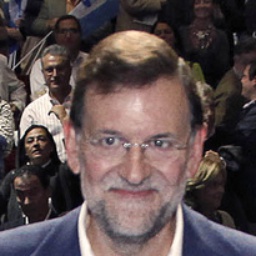}\hfill
    \includegraphics[width=0.166\linewidth]{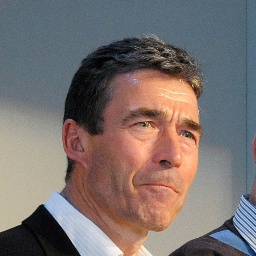}\hfill
    \includegraphics[width=0.166\linewidth]{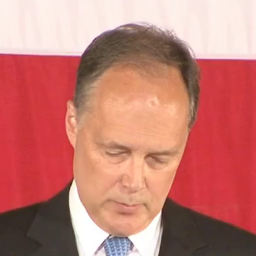}\hfill
    \includegraphics[width=0.166\linewidth]{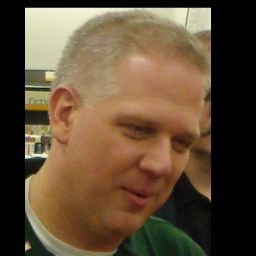}\\[-0.1em]
    \includegraphics[width=0.166\linewidth]{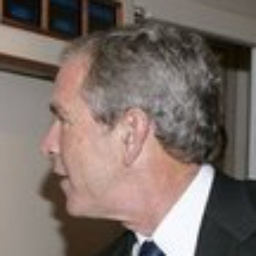}\hfill
    \includegraphics[width=0.166\linewidth]{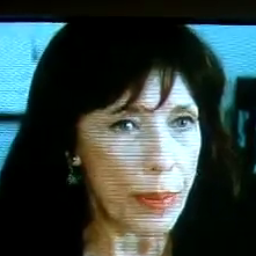}\hfill
    \includegraphics[width=0.166\linewidth]{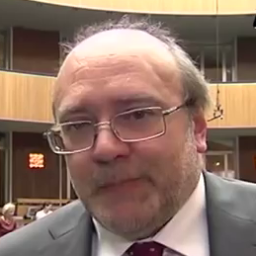}\hfill
    \includegraphics[width=0.166\linewidth]{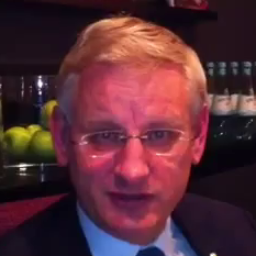}\hfill
    \includegraphics[width=0.166\linewidth]{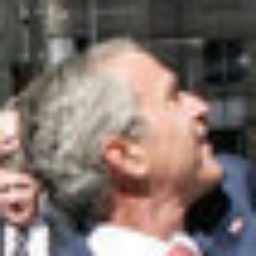}\hfill
    \includegraphics[width=0.166\linewidth]{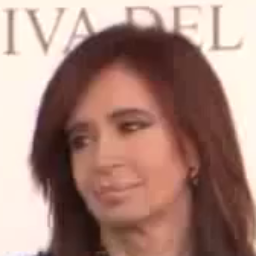}\\[-0.1em]
    \includegraphics[width=0.166\linewidth]{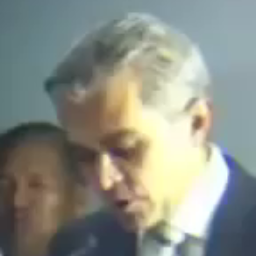}\hfill
    \includegraphics[width=0.166\linewidth]{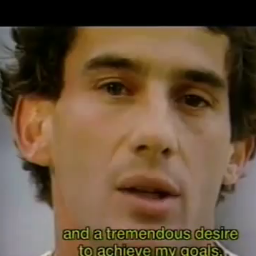}\hfill
    \includegraphics[width=0.166\linewidth]{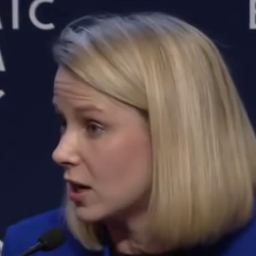}\hfill
    \includegraphics[width=0.166\linewidth]{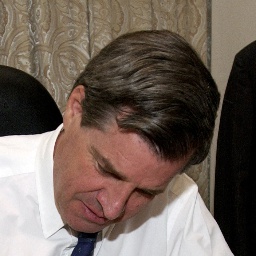}\hfill
    \includegraphics[width=0.166\linewidth]{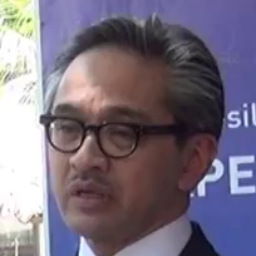}\hfill
    \includegraphics[width=0.166\linewidth]{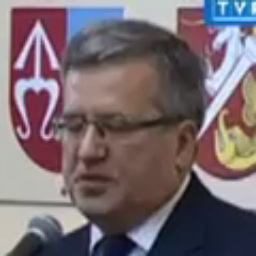}\\
    \vspace{-2.0em}\begin{center}(b) IJB-A\end{center}
    \end{minipage}
    \begin{minipage}{0.48\linewidth}
    \includegraphics[width=0.166\linewidth]{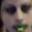}\hfill
    \includegraphics[width=0.166\linewidth]{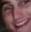}\hfill
    \includegraphics[width=0.166\linewidth]{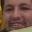}\hfill
    \includegraphics[width=0.166\linewidth]{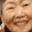}\hfill
    \includegraphics[width=0.166\linewidth]{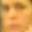}\hfill
    \includegraphics[width=0.166\linewidth]{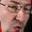}\\[-0.1em]
    \includegraphics[width=0.166\linewidth]{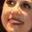}\hfill
    \includegraphics[width=0.166\linewidth]{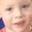}\hfill
    \includegraphics[width=0.166\linewidth]{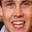}\hfill
    \includegraphics[width=0.166\linewidth]{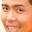}\hfill
    \includegraphics[width=0.166\linewidth]{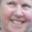}\hfill
    \includegraphics[width=0.166\linewidth]{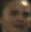}\\[-0.1em]
    \includegraphics[width=0.166\linewidth]{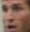}\hfill
    \includegraphics[width=0.166\linewidth]{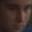}\hfill
    \includegraphics[width=0.166\linewidth]{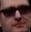}\hfill
    \includegraphics[width=0.166\linewidth]{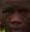}\hfill
    \includegraphics[width=0.166\linewidth]{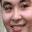}\hfill
    \includegraphics[width=0.166\linewidth]{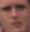}\\[-0.1em]
    \includegraphics[width=0.166\linewidth]{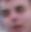}\hfill
    \includegraphics[width=0.166\linewidth]{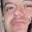}\hfill
    \includegraphics[width=0.166\linewidth]{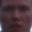}\hfill
    \includegraphics[width=0.166\linewidth]{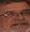}\hfill
    \includegraphics[width=0.166\linewidth]{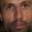}\hfill
    \includegraphics[width=0.166\linewidth]{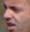}\\
    \vspace{-2.0em}\begin{center}(c) TinyFace\end{center}
    \end{minipage}\hfill
    \begin{minipage}{0.48\linewidth}
    \includegraphics[width=0.166\linewidth]{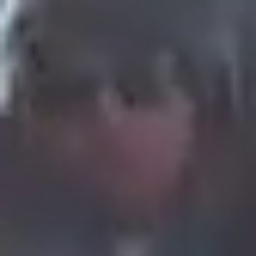}\hfill
    \includegraphics[width=0.166\linewidth]{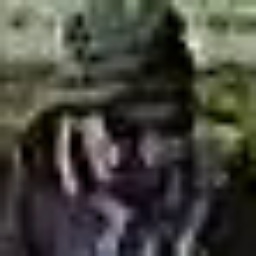}\hfill
    \includegraphics[width=0.166\linewidth]{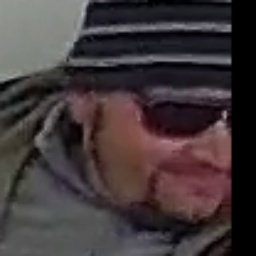}\hfill
    \includegraphics[width=0.166\linewidth]{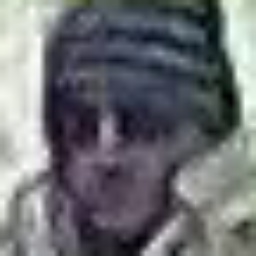}\hfill
    \includegraphics[width=0.166\linewidth]{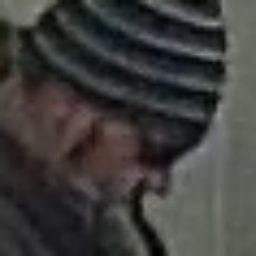}\hfill
    \includegraphics[width=0.166\linewidth]{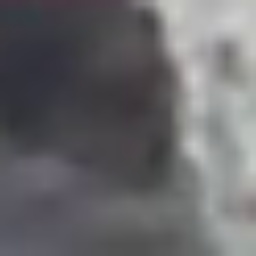}\\
    \includegraphics[width=0.166\linewidth]{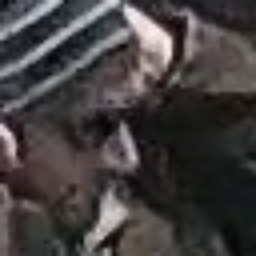}\hfill
    \includegraphics[width=0.166\linewidth]{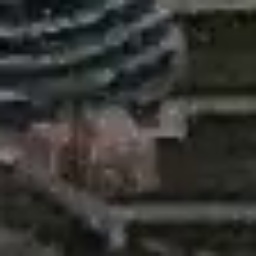}\hfill
    \includegraphics[width=0.166\linewidth]{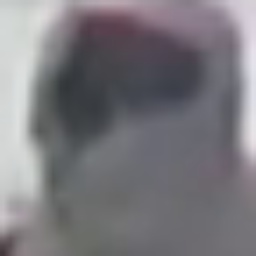}\hfill
    \includegraphics[width=0.166\linewidth]{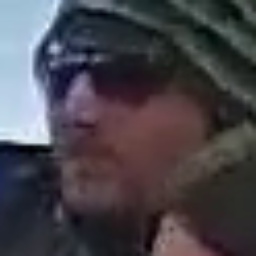}\hfill
    \includegraphics[width=0.166\linewidth]{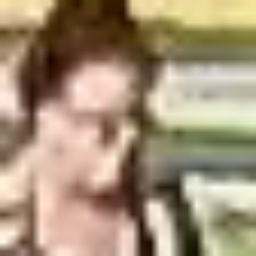}\hfill
    \includegraphics[width=0.166\linewidth]{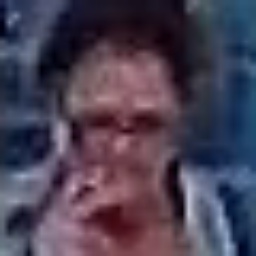}\\
    \includegraphics[width=0.166\linewidth]{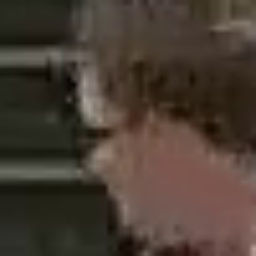}\hfill
    \includegraphics[width=0.166\linewidth]{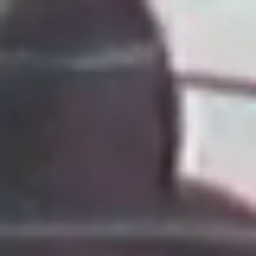}\hfill
    \includegraphics[width=0.166\linewidth]{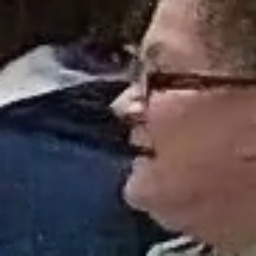}\hfill
    \includegraphics[width=0.166\linewidth]{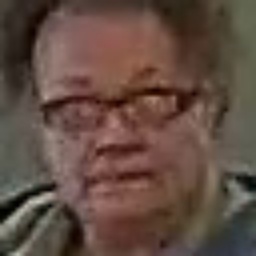}\hfill
    \includegraphics[width=0.166\linewidth]{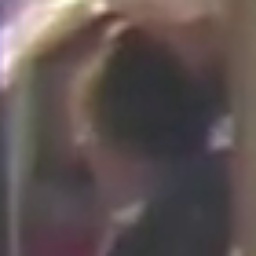}\hfill
    \includegraphics[width=0.166\linewidth]{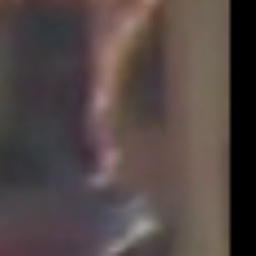}\\
    \includegraphics[width=0.166\linewidth]{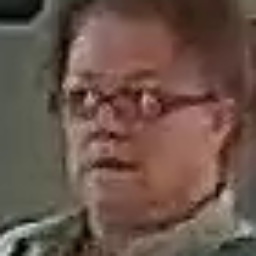}\hfill
    \includegraphics[width=0.166\linewidth]{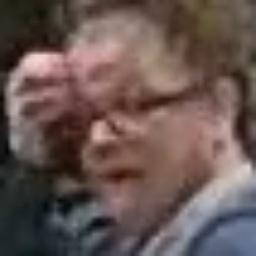}\hfill
    \includegraphics[width=0.166\linewidth]{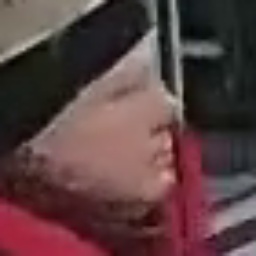}\hfill
    \includegraphics[width=0.166\linewidth]{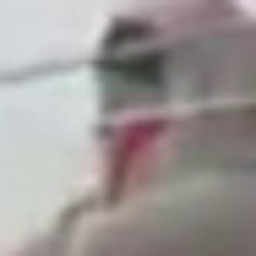}\hfill
    \includegraphics[width=0.166\linewidth]{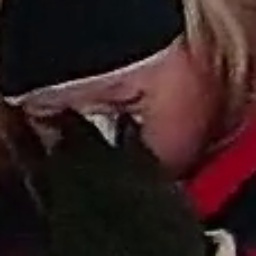}\hfill
    \includegraphics[width=0.166\linewidth]{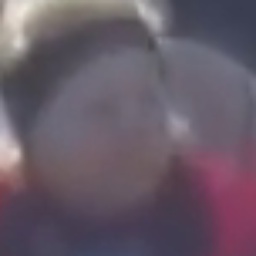}\\
    \vspace{-2.0em}\begin{center}(d) IJB-S\end{center}
    \end{minipage}

    \vspace{-0.9em}\caption{Examples images from the testing datasets.}
    \label{fig:exp_dataset_appendix}
\end{figure*}

Figure~\ref{fig:exp_dataset_appendix} shows more image examples from different types of the dataset. The images in the LFW (Type I) dataset are mostly high quality face images with limited variations. Therefore, different models in our experiment all achieve similar performance on this dataset. The images in the IJB-A (Type II) show more variations, some of which are extremely challenging. This requires the representation model to be able to perform a cross-domain matching between images of high quality and low quality. Further, the TinyFace and IJB-S (Type III) datasets are mostly composed of low-quality faces. This requires the face representation to be invariant to large variations that can hardly be found in the public training datasets.

\begin{table}[t]
\captionsetup{font=footnotesize}
\newcommand{\mr}[1]{\multirow{2}{*}{#1}}
\footnotesize
\setlength{\tabcolsep}{6pt}
\begin{center}
\begin{tabularx}{1.00\linewidth}{c|c| c|c| c|c}
\Xhline{2\arrayrulewidth}
\multicolumn{2}{c|}{Method} & \multicolumn{2}{c|}{TinyFace} & \multicolumn{2}{c}{IJB-S} \\\hline
Augmentable & Additional & \mr{Rank1} & \mr{Rank5} & \mr{Rank1} & \mr{Rank5} \\
Variations  & Variations & & & & \\
\Xhline{2\arrayrulewidth}
% 0 & 0               & 57.30 & 63.73 & 59.66 & 66.30 \\
0 & 0               & 55.04 & 60.97 & 59.71 & 66.32\\\hline
% 3 & 0               & 57.22 & 62.37 & 60.13 & 66.14 \\
3 & 0               & 54.99 & 61.32 & 62.22 & 67.03\\\hline
% 3 & 1               & 59.82 & 64.59 & 60.69 & 66.77 \\
3 & 1               & 61.80 & 67.94 & 62.30 & 67.51 \\\hline
% 3 & 3               & 61.32 & 66.34 & 60.74 & 66.59\\
3 & 3               & 63.89 & 68.67 & 61.98 & 67.12\\
\Xhline{2\arrayrulewidth}
\end{tabularx}
\vspace{-0.7em}\caption{Gradually adding more variations into variation decorrelation loss. All of the models use all the other modules proposed in the paper.}
\label{tab:ablation_variation}
\end{center}
\end{table}

\begin{table*}[h]
\newcommand{\hly}{\cellcolor{Y}}
\newcommand{\hlg}{\cellcolor{G}}
\captionsetup{font=footnotesize}
\begin{threeparttable}
    \centering
    \footnotesize
    \setlength{\tabcolsep}{5.5pt}
		\centering
        \scalebox{1.0}{
		\begin{tabularx}{\linewidth}{X|c || c|c|c|c|c || c|c|c|c|c}
\Xhline{2\arrayrulewidth}
		\multirow{2}{*}{Method} & \multirow{2}{*}{Training Data} & \multicolumn{5}{c||}{Surveillance-to-Single} & \multicolumn{5}{c}{Surveillance-to-Booking} \\
		\cline{3-7} \cline{8-12}
		&               & Rank-1 & Rank-5 & Rank-10 & 1\%  & 10\%   
		                & Rank-1 & Rank-5 & Rank-10 & 1\%  & 10\%  \\
\Xhline{2\arrayrulewidth}
% 		C-FAN~\cite{gong2019video}    & 5.0M   
% 		                & 50.82 & 61.16 & 64.95 & 16.44 & 24.19 
% 		                & 53.04 & 62.67 & 66.35 & 27.40 & 29.70 
% 		                & 10.05 & 17.55 & 21.06 & 0.11 & 0.68 \\
	     PFE~\cite{shi2019probabilistic} & 4.4M  
	            & 50.16 & 58.33 & 62.28 & 31.88 & 35.33 
		        & 53.60 & 61.75 & 64.97 & 35.99 & 39.82 \\
		 ArcFace~\cite{deng2018arcface}\tnote{+} & 5.8M  
		        & 57.35 & 64.42 & 68.36 & 41.85 & 50.12 
		        & 57.36 & 64.95 & 68.57 & 41.23 & 49.18\\\hline
		 Ours (Basline) & 4.8M  
		        & 47.94 & 55.40 & 59.37 & 25.60 & 36.03 
		        & 37.14 & 46.75 & 51.59 & 24.75 & 31.10\\
		 Ours (Baseline + VA) & 4.8M    
		        & 60.61 & 66.53 & 68.57 & 31.97 & 44.25 
            	& 51.27 & 58.94 & 63.25 & 31.19 & 44.22\\
		 Ours (all) & 4.8M      
		        & 58.94 & 65.48 & 68.31 & 37.57 & 50.17 
		        & 60.74 & 66.59 & 68.92 & 37.11 & 51.00\\
		 Ours (all) + PA & 4.8M  
		        & 59.79 & 65.78 & 68.20 & 41.06 & 53.23 
		        & 61.98 & 67.12 & 69.10 & 42.73 & 53.48\\
\Xhline{2\arrayrulewidth}
		\end{tabularx}}
    \vspace{-0.7em}\caption{Performance comparison on the IJB-S dataset. The performance is reported in terms of rank retrieval (closed-set) and TPIR@FPIR (open-set) instead of the media-normalized version~\cite{IJBS}. The numbers ``1\%'' and ``10\%'' in the second row refer to the FPIR. ``+'' indicates the testing performance by using the released models from corresponding authors.}\vspace{-0.5em}
    \label{tab:ijbs}
\end{threeparttable}
\end{table*}

% \begin{figure*} [t]
%     \centering
%     \includegraphics[width=0.98\linewidth]{fig/stem_attention_sidmoid_new_3.pdf}\\[-0.7em]
%     \includegraphics[width=1.0\linewidth]{fig/seaborn_colorbar_new.jpg}\\
%     \caption{Visualization of sub-embedding uncertainty on more testing images.}
%     \label{fig:uncertainty_new_3}
% \end{figure*}

\end{document}